\documentclass[oneside,onecolumn,11pt,a4paper]{book}
\usepackage[utf8]{inputenc}
\usepackage[english]{babel}

\usepackage{graphicx}
\usepackage{amsmath}
\usepackage{amssymb}
\usepackage{mathtools}
\usepackage{color}
\usepackage{siunitx}
\usepackage{array}
\usepackage{float}
\usepackage{caption}
\usepackage{subcaption}
\usepackage{colortbl}
\usepackage{tabularx}
\usepackage{tabu}
\usepackage{textcomp}
\usepackage{minitoc}
\usepackage{fancyhdr}
\usepackage{algorithm}
\usepackage{algpseudocode}
\usepackage[breakable, theorems, skins]{tcolorbox}
\usepackage{frontespizio}
\usepackage{setspace}
\usepackage{multirow}
\usepackage{booktabs}
\usepackage{rotating}
\usepackage{hyperref}
\usepackage{breakcites}
\newcommand{\etal}{\textit{et al}. }
\usepackage{soul}
\usepackage[export]{adjustbox}

\usepackage{algorithmicx}

\usepackage{xcolor,pifont}
\newcommand*\colourcheck[1]{%
  \expandafter\newcommand\csname #1check\endcsname{\textcolor{#1}{\ding{52}}}%
}
\colourcheck{blue}
\colourcheck{green}
\colourcheck{red}

\usepackage{arydshln}
\usepackage{booktabs}

\newcommand{\keypoint}[1]{\vspace{0.0cm}\noindent\textbf{#1}\quad}
\newcommand{\cmark}{\ding{51}}%
\newcommand{\xmark}{\ding{55}}%

\newcommand{\aneeshan}[1]{\textcolor{black}{#1}}
\newcommand{\cut}[1]{}
\def\red#1{\textcolor[rgb]{1,0,0}{#1}}
\def\blue#1{\textcolor[rgb]{0,0,1}{#1}}

\begin{document}	
\frontmatter
\pagestyle{fancy}
\renewcommand{\headrulewidth}{0pt}
\fancyhf{}
\setcounter{secnumdepth}{3}
\setcounter{tocdepth}{2}
\dominitoc
\mtcsettitle{minitoc}{Content}

\begin{titlepage}

\centering 

{\bfseries {\LARGE{Towards Practicality of \\[6pt] Sketch-Based Visual Understanding}}}

\vspace{1.7cm}

{\LARGE Ayan Kumar Bhunia}
\vspace{1.7cm}

{\LARGE Centre for Vision, Speech and Signal Processing}
\vspace{0.5cm}

{\LARGE Faculty of Engineering and Physical Sciences}
\vspace{0.5cm}

{\LARGE University of Surrey, United Kingdom}
\vspace{1.2cm}

\includegraphics[width=0.4\textwidth]{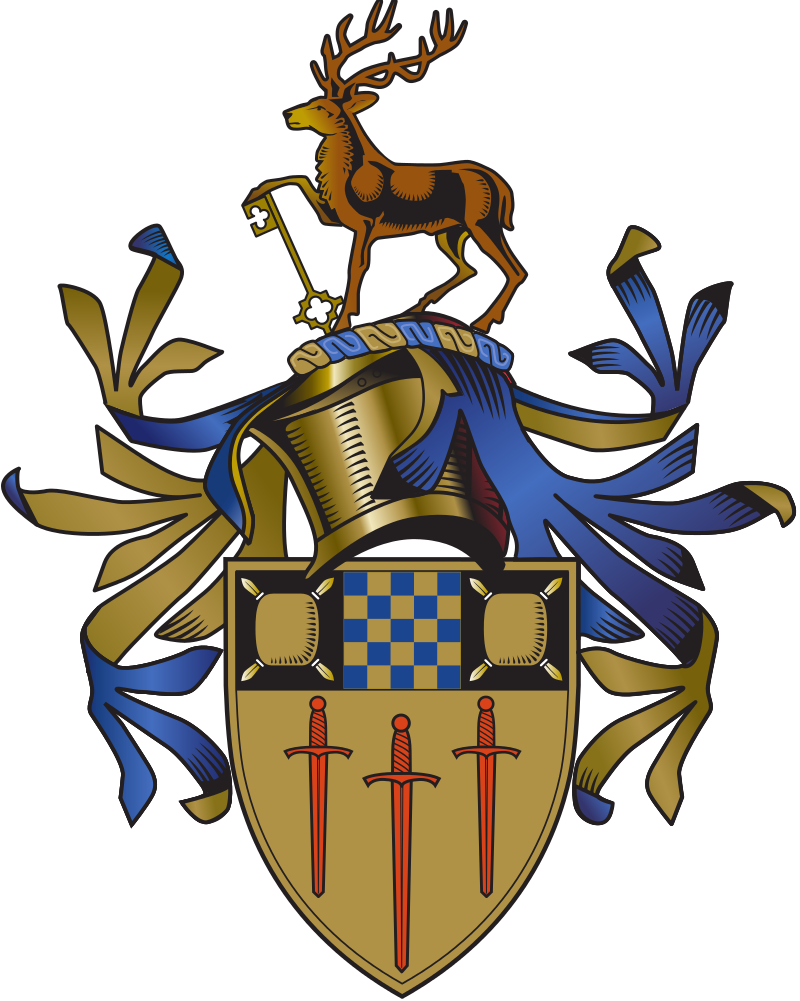}
\vspace{1.2cm}

{\LARGE This dissertation is submitted for the degree of}
\vspace{0.5cm}

{\LARGE \textit{Doctor of Philosophy}}
\vspace{1.2cm}

{\LARGE Aug 2022}

\end{titlepage}
\newpage\null\thispagestyle{empty}\newpage

\pagenumbering{roman}

\chapter*{}
    \textit{This thesis is dedicated to my parents, Mrs. Jyotsna Bhunia and Mr. Anjan Kumar Bhunia, for their endless support and encouragement.}

\newpage\null\thispagestyle{empty}\newpage

\chapter*{Declaration}
   \vspace{-2em}
I hereby declare that this thesis is the outcome of my own efforts. It has not been submitted to this or any other university for a degree or professional qualification, in whole or in part. Any ideas, data or any source of information are acknowledged carefully in the text, bibliography or footnotes. Some parts of this thesis have been published as follows:
   \vspace{1em}
   
   \begin{enumerate}

    \item[Chapter \ref{chapter:OTF}]  
     \textbf{Ayan Kumar Bhunia}, Yongxin Yang, Timothy M. Hospedales, Tao Xiang, Yi-Zhe Song.``Sketch Less for More: On-the-Fly Fine-Grained Sketch Based Image Retrieval".  IEEE/CVF Conference on Computer Vision and Pattern Recognition ({CVPR}) 2020.

   \item[Chapter \ref{chapter:SSL_FGSBIR}] \noindent \textbf{Ayan Kumar Bhunia}, Pinaki Nath Chowdhury, Aneeshan Sain, Yongxin Yang, Tao Xiang, Yi-Zhe Song. ``More Photos are All You Need: Semi-Supervised Learning for Fine-Grained Sketch Based Image Retrieval".  
    IEEE/CVF Conference on Computer Vision and Pattern Recognition ({CVPR}) 2021.

   \item[Chapter \ref{chapter:NT_FGSBIR}] \noindent \textbf{Ayan Kumar Bhunia}, Subhadeep Koley, Abdullah Faiz Ur Rahman Khilji, Aneeshan Sain, Pinaki Nath Chowdhury, Tao Xiang, Yi-Zhe Song. ``Sketching without Worrying: Noise-Tolerant Sketch-Based Image Retrieval". 
    IEEE/CVF Conference on Computer Vision and Pattern Recognition ({CVPR}) 2022.

    \item[Chapter \ref{chapter:Self}] \noindent \textbf{Ayan Kumar Bhunia}, Pinaki Nath Chowdhury, Yongxin Yang, Timothy Hospedales, Tao Xiang, Yi-Zhe Song. ``Vectorization and Rasterization: Self-Supervised Learning for Sketch and Handwriting".  IEEE/CVF Conference on Computer Vision and Pattern Recognition ({CVPR}) 2021.  
 
   \item[Chapter \ref{chapter:SBIL}] \noindent \textbf{Ayan Kumar Bhunia}, Viswanatha Reddy Gajjala, Subhadeep Koley, Rohit Kundu, Aneeshan Sain, Tao Xiang, Yi-Zhe Song. ``Doodle It Yourself: Class Incremental Learning by Drawing a Few Sketches". 
    IEEE/CVF Conference on Computer Vision and Pattern Recognition ({CVPR}) 2022.

   \end{enumerate}

\newpage\null\thispagestyle{empty}\newpage   

\chapter{List of Publications}

\begin{enumerate}
   \item  \noindent \textbf{Ayan Kumar Bhunia}, Yongxin Yang, Timothy M. Hospedales, Tao Xiang, Yi-Zhe Song.``Sketch Less for More: On-the-Fly Fine-Grained Sketch Based Image Retrieval". IEEE/CVF Conference on Computer Vision and Pattern Recognition (\textbf{CVPR}) 2020.

   \item \noindent \textbf{Ayan Kumar Bhunia}, Pinaki Nath Chowdhury, Yongxin Yang, Timothy Hospedales, Tao Xiang, Yi-Zhe Song. ``Vectorization and Rasterization: Self-Supervised Learning for Sketch and Handwriting". 
    IEEE/CVF Conference on Computer Vision and Pattern Recognition (\textbf{CVPR}) 2021.

    \item \noindent \textbf{Ayan Kumar Bhunia}, Pinaki Nath Chowdhury, Aneeshan Sain, Yongxin Yang, Tao Xiang, Yi-Zhe Song. ``More Photos are All You Need: Semi-Supervised Learning for Fine-Grained Sketch Based Image Retrieval".  
    IEEE/CVF Conference on Computer Vision and Pattern Recognition (\textbf{CVPR}) 2021.

   \item \noindent \textbf{Ayan Kumar Bhunia}, Viswanatha Reddy Gajjala, Subhadeep Koley, Rohit Kundu, Aneeshan Sain, Tao Xiang, Yi-Zhe Song. ``Doodle It Yourself: Class Incremental Learning by Drawing a Few Sketches".  
    IEEE/CVF Conference on Computer Vision and Pattern Recognition (\textbf{CVPR}) 2022.

   \item \noindent \textbf{Ayan Kumar Bhunia}, Subhadeep Koley, Abdullah Faiz Ur Rahman Khilji, Aneeshan Sain, Pinaki Nath Chowdhury, Tao Xiang, Yi-Zhe Song. ``Sketching without Worrying: Noise-Tolerant Sketch-Based Image Retrieval".  
    IEEE/CVF Conference on Computer Vision and Pattern Recognition (\textbf{CVPR}) 2022.

\item \noindent \textbf{Ayan Kumar Bhunia*}, Ayan Das*, Umar Riaz Muhammad*, Yongxin Yang, Timothy M. Hospedales, Tao Xiang, Yulia Gryaditskaya, Yi-Zhe Song. ``Pixelor: A Competitive Sketching AI Agent. So you think you can beat me?".  
    \textbf{SIGGRAPH Asia} 2020. (*Equal Contribution)

\item \noindent Aneeshan Sain, \textbf{Ayan Kumar Bhunia}, Yongxin Yang, Tao Xiang, Yi-Zhe Song. ``Cross-Modal Hierarchical Modelling for Fine-Grained Sketch Based Image Retrieval".  
    British Machine Vision Conference (\textbf{BMVC}) 2020.

\item \noindent Ruoyi Du, Dongliang Chang, \textbf{Ayan Kumar Bhunia}, Jiyang Xie, Zhanyu Ma, Yi-Zhe Song, Jun Guo. ``Fine-grained visual classification via progressive multi-granularity training of jigsaw patches".  
    European Conference on Computer Vision (\textbf{ECCV}) 2020.
    
\item \noindent Aneeshan Sain, \textbf{Ayan Kumar Bhunia}, Yongxin Yang and , Tao Xiang, Yi-Zhe Song. ``StyleMeUp: Towards Style-Agnostic Sketch-Based Image Retrieval".  
    IEEE/CVF Conference on Computer Vision and Pattern Recognition (\textbf{CVPR}) 2021.

\item \noindent \textbf{Ayan Kumar Bhunia}, Shuvozit Ghose, Amandeep Kumar, Pinaki Nath Chowdhury, Aneeshan Sain, Yi-Zhe Song. ``MetaHTR: Towards Writer-Adaptive Handwritten Text Recognition".  
    IEEE/CVF Conference on Computer Vision and Pattern Recognition (\textbf{CVPR}) 2021.
    
    \item \noindent \textbf{Ayan Kumar Bhunia}, Aneeshan Sain, Pinaki Nath Chowdhury, Yi-Zhe Song. ``Text is Text, No Matter What: Unifying Text Recognition using Knowledge Distillation".  
    IEEE International Conference on Computer Vision (\textbf{ICCV}) 2021.

\item \noindent \textbf{Ayan Kumar Bhunia}, Pinaki Nath Chowdhury, Aneeshan Sain, Yi-Zhe Song. ``Towards the Unseen: Iterative Text Recognition by Distilling from Errors".  
    IEEE International Conference on Computer Vision (\textbf{ICCV}) 2021.

\item \noindent \textbf{Ayan Kumar Bhunia}, Aneeshan Sain, Amandeep Kumar, Shuvozit Ghose, Pinaki Nath Chowdhury, Yi-Zhe Song. ``Joint Visual Semantic Reasoning: Multi-Stage Decoder for Text Recognition". IEEE International Conference on Computer Vision (\textbf{ICCV}) 2021.

\item \noindent Pinaki Nath Chowdhury, \textbf{Ayan Kumar Bhunia}, Viswanatha Reddy Gajjala, Aneeshan Sain, Tao Xiang, Yi-Zhe Song. ``Partially Does It: Towards Scene-Level FG-SBIR with Partial Input". 
	IEEE/CVF Conference on Computer Vision and Pattern Recognition (\textbf{CVPR}) 2022. 

 \item \noindent Aneeshan Sain, \textbf{Ayan Kumar Bhunia}, Vaishnav Potlapalli , Pinaki Nath Chowdhury, Tao Xiang, Yi-Zhe Song. ``Sketch3T: Test-time Training for Zero-Shot SBIR".  
    IEEE/CVF Conference on Computer Vision and Pattern Recognition (\textbf{CVPR}) 2022.

\item \noindent \textbf{Ayan Kumar Bhunia}, Aneeshan Sain, Parth Hiren Shah, Animesh Gupta, Pinaki Nath Chowdhury, Tao Xiang , Yi-Zhe Song. ``Adaptive Fine-Grained Sketch-Based Image Retrieval".  
    European Conference on Computer Vision (\textbf{ECCV}) 2022.  

\item \noindent Pinaki Nath Chowdhury, Aneeshan Sain, \textbf{Ayan Kumar Bhunia},  Tao Xiang, Yulia Gryaditskaya,  Yi-Zhe Song . ``FS-COCO: Towards Understanding of Freehand Sketches of Common Objects in Context".  
    European Conference on Computer Vision (\textbf{ECCV}) 2022.  

\end{enumerate}  

\newpage\null\thispagestyle{empty}\newpage

\chapter{Acknowledgements}
    I would like to express my sincere gratitude firstly to my supervisor Prof Yi-Zhe Song for allowing me to pursue my PhD at SketchX lab with a fully-funded PhD scholarship. Besides his meticulous research guidance, SonG has patiently taught me the professional skills for becoming a proper researcher over time. Having him as my PhD supervisor is one of the best decisions of my life so far. I shall always be grateful for his mentorship, which has helped me prosper in research, career and life. Also, a huge thanks to Prof. Tao Xiang and Prof. Timothy M. Hospedales for their insightful suggestions on my research works. 
    
    Words would fall short to describe the support and favours I received from my SketchX lab mates -- Aneeshan Sain, Pinaki Nath Chowdhury, and Subhadeep Koley. Despite being less social, I never felt isolated in this hectic journey of my PhD because of their constant presence -- both academic and emotional. Life and research would not have been easy without their love and friendship. In particular, I am grateful to Aneeshan for sticking with me through highs and lows, and helping me proofread my thesis.
    
    I would also like to thank Dr. Kaiyue Pang, Dr. Conghui Hu, Dr. Anran Qi, Ayan Das, Ling Luo, Dr. Umar Riaz Muhammad, Yue Zhong, Lan Yang for being not only amazing colleagues but also great friends.
    
    I shall always be grateful to Prof Partha Pratim Roy and Prof Umapada Pal, who provided me with the initial opportunity to pursue my undergraduate research work with them. Many of these research outputs helped me secure a fully-funded PhD position in the UK, without which coming this far would have been very difficult if not impossible. 
    
    Above all, I thank my parents for their endless love, support, and encouragement. I am humbled by all the sacrifices they had made throughout their lives towards my better future. Also a special thanks to my talented brother, Ankan.

 \newpage\null\thispagestyle{empty}\newpage

\chapter{Abstract}
Sketches have been used to conceptualise and depict visual objects from pre-historic times. Sketch research has flourished in the past decade, particularly with the proliferation of touchscreen devices. Much of the utilisation of sketch has been anchored around the fact that it can be used to delineate visual concepts universally irrespective of age, race, language, or demography. The fine-grained interactive nature of sketches facilitates the application of sketches to various visual understanding tasks, like image retrieval, image-generation or editing, segmentation, 3D-shape modelling etc. However, sketches are highly abstract and subjective based on the perception of individuals. Although most agree that sketches provide fine-grained control to the user to depict a visual object, many consider sketching a tedious process due to their limited sketching skills compared to other query/support modalities like text/tags. Furthermore, collecting fine-grained sketch-photo association is a significant bottleneck to commercialising sketch applications. Therefore, this thesis aims to progress sketch-based visual understanding towards more practicality. 





Being able to model fine-grained details of a visual concept easily, {sketch} understandably holds immense potential as a medium of query -- even better than texts that are at times insufficient to pin down fine-grained visual details. Out of all sketch-related applications therefore, fine-grained sketch-based image retrieval (FG-SBIR) has received the most attention, due to its significant commercial potential in the retail industry. FG-SBIR aims at retrieving a particular photo instance given a user’s query sketch, out of a gallery of photos having a particular category. Given the dominant prevalence of touchscreen devices now, the world is already primed for using sketch as a practical query modality for fine-grained retrieval. However, talking of applicability at an industrial scale, practically using sketch as a query-medium for retrieval is yet to gain traction due to a few significant barriers. 
{Breaking these barriers, this thesis addresses the practicality of FG-SBIR via two themes putting forth five major contributions. The first theme comprises three contributions which focus particularly on the \emph{practical deployment} of FG-SBIR which is one of the major forefront of sketch research.
The second theme caters to the \emph{widespread applicability} of sketches for real-world applications, consisting of two more contributions.}
{In the first theme, our \cut{leading}first chapter starts by figuring} out that the widespread applicability of FG-SBIR is hindered as drawing a sketch takes time, and most people struggle to draw a complete and faithful sketch. We thus reformulate the conventional FG-SBIR framework to tackle these challenges, with the ultimate goal of retrieving the target photo with the least number of strokes possible. We further propose an on-the-fly design that starts retrieving as soon as the user starts drawing. Accordingly, we devise a reinforcement learning-based cross-modal retrieval framework that optimises the ground-truth photo's rank over a complete sketch drawing episode.  

\aneeshan{The second chapter discovers} that the lack of sketch-photo pairs largely bottlenecks FG-SBIR performance. We therefore introduce a novel semi-supervised framework for instance-level cross-modal retrieval that can leverage large-scale unlabelled photos to account for data scarcity. The core of our semi-supervision design is a sequential photo-to-sketch generation model that aims to generate paired sketches for unlabelled photos. We further introduce a discriminator guided mechanism to guide against unfaithful generation, together with a distillation loss-based regulariser to provide tolerance against noisy training samples. 

Thirdly, we notice that the fear-to-sketch problem (i.e., “I can’t sketch”) has proven to be fatal for the widespread adoption of fine-grained SBIR. A pilot study revealed that the secret lies in the existence of noisy strokes, but not so much of the “I can’t sketch”. We thus design a stroke subset selector that detects noisy strokes, leaving only those that positively contribute to a successful retrieval. Our Reinforcement Learning based formulation quantifies the importance of each stroke present in a given subset, based on the extent to which that stroke contributes to retrieval.

{Moving on to our second theme, in the fourth chapter we focus on learning powerful representations via self-supervised learning from unlabelled data, thus increasing \aneeshan{the scope of sketches}. Towards this, we advocate \aneeshan{for the dual-modality nature of sketches}\cut{that the existence of hand-drawn sketches in two modality} -- rasterized images and vector coordinate sequences, which is pivotal in designing a self-supervised pre-text task for our goal. We address this dual representation by proposing two novel cross-modal translation pre-text tasks for self-supervised feature learning: Vectorization and Rasterization. Vectorization learns to map image space to vector coordinates, and rasterization the \aneeshan{opposite}. We show that our learned encoder modules benefit both raster-based and vector-based downstream \aneeshan{tasks} to analysing hand-drawn data.}


{ 
{The final chapter further contributes to the second theme by opening } up a new avenue of sketch research with a novel sketch-based visual understanding task. Here we dictate the potential of sketch as a support modality for few-shot class incremental learning (FSCIL) --  a setup that explicitly highlights the relevance of sketch as an information-modality. Pushing FSCIL further, we address two key questions that bottleneck its ubiquitous application (i) can the model learn from diverse modalities other than just photo (as humans do), and (ii) what if photos are not readily accessible (due to ethical and privacy constraints). 
The product is a “Doodle It Yourself” (DIY) FSCIL framework where the users can freely sketch a few examples of a novel class for the model to learn recognising photos of that class.
}

\thispagestyle{empty}

\renewcommand*\contentsname{Content}
\tableofcontents
\listoffigures 
\listoftables

\mainmatter 

\pagenumbering{arabic}
\minitoc

\pagestyle{fancy}				
\renewcommand{\chaptermark}[1]{\markboth{#1}{}}
\renewcommand{\sectionmark}[1]{\markright{\thesection\ #1}}

\fancyhf{} 
\fancyhead[LE,RO]{\bfseries\thepage} 
\fancyhead[LO]{\bfseries\rightmark}
\fancyhead[RE]{\bfseries\leftmark} 
\setlength{\headheight}{13.60pt}
\renewcommand{\headrulewidth}{0.5pt}
\renewcommand{\footrulewidth}{0pt}

 \chapter{Introduction}
    \label{chapter:introduction}
    \minitoc
    \newpage
    \section{Background}
    
\begin{figure}[!hbt]
\centering
\includegraphics[width=0.7\linewidth]{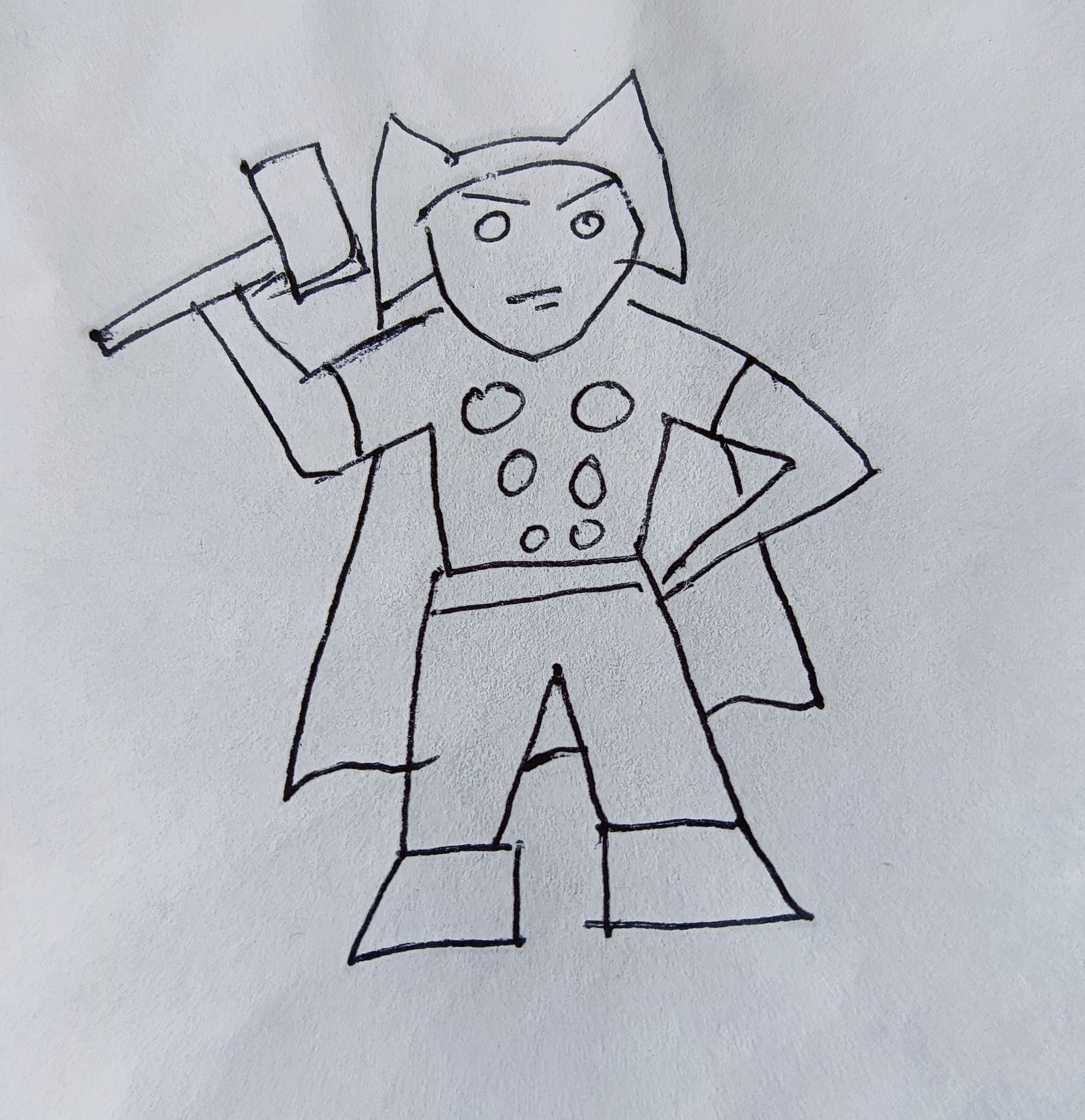}
\vspace{-0.25cm}
\caption{\aneeshan{The very sketch of Thor by my 5-years-old niece.}}
\label{fig:thor}
\end{figure}

Looking at a stuffed toy last night, my niece scoffed, ``This isn't Thor! Let me show you..". Swiftly drawing a rough sketch of a man holding a hammer (Figure~\ref{fig:thor}), she proudly upheld a piece of paper. I was mesmerised by the trivial yet most fundamental utility of sketching -- clearly the most representative  mode of communication amongst humans. An amateur sketch is such a quick and easy way of expressing one's imagination, that one may claim \textit{To imagine is to sketch}. Sketching can be traced back to dark caves, pre-dating the development of human rationale, where simple concepts like danger were communicated via drawn figures among cavemen. It is but natural that modern research explore and utilise its potential to the fullest -- more so, in this day and age of touchscreen devices, where finger-traces dominate the daily life. Pursuing deeper we realise, that \textit{sketching} a concept is often more illustrative than jotting down some textual description. Even in its imperfection, a sketch of a dog would easily put details like pose, body structure, and even its breed or state (e.g hurt or happy) etc. into perspective, much better than a long sentence describing the same. Computer vision has thus progressed in researching this very medium of data that can easily abstract the concept from an image, to fork out applications useful in real world.

\vspace{0.2cm}

Among the fairly researched directions on amateur sketches in the vision community now, one of the most dominant is that of sketch-based image retrieval \cite{collomosse2019livesketch, yelamarthi2018zero, dey2019doodle}. This falls under the umbrella of content based image retrieval problems, like text-based image retrieval \cite{wang2019camp}. In particular, it refers to the task of retrieving an image based on a sketch given as a query. One may wonder, why use sketch when we already have text as a predominant medium for querying..? Well, unlike texts, a sketch can convey meaningful visual descriptions that sometimes are quite ambiguous or too lengthy to describe in words. Furthermore, with the recent upsurge of user-friendly touchscreen devices and accessories, collecting sketches on a large scale, is no longer a big hurdle. This has further encouraged  the vision community to explore how human-drawn freehand sketches can be employed for various visual understanding tasks. For instance, sketches have been used in image retrieval \cite{collomosse2019livesketch, yelamarthi2018zero, dey2019doodle}, image generation \cite{ghosh2019interactive, chen2020deepfacedrawing}, image editing \cite{jo2019sc, yang2020deep}, segmentation \cite{hu2020sketch}, image-inpainting \cite{xie2021exploiting}, video synthesis \cite{li2021deep}, representation learning \cite{wang2021sketchembednet}, 3D shape modelling \cite{zhang2021sketch2model}, augmented reality \cite{yan2020interactive}, among others \cite{xu2020deep}.  


Sketch-based image retrieval (SBIR) \cite{collomosse2019livesketch, dey2019doodle, dutta2019semantically} was initially posed as a category-level retrieval \cite{collomosse2019livesketch} problem, where given a sketch of any category (e.g. cat, dog, etc.), the model was expected to retrieve an image of the same category. However, it became quite evident that the key advantage of a sketch over text/tag-based image retrieval lies in its ability to convey \emph{fine-grained} details \cite{engilberge2018finding}. Despite starting as a category-level retrieval problem \cite{collomosse2019livesketch, dey2019doodle, sketch2vec, SBIR_imbalance}, the fine-grained nature of sketches therefore directed current research focus more towards \textit{fine-grained} SBIR (FG-SBIR) \cite{yu2016sketch, pang2020solving, pang2019generalising}.  FG-SBIR, unlike its predecessor (categorical SBIR) aims to retrieve a \emph{particular} photo based on a query sketch at an intra-category basis. In other words, fine-grained SBIR fetches a \emph{particular} photo from a gallery containing photos of the same category -- e.g retrieving a particular shoe from a shoe-gallery.

There are quite a few challenges in this arena however, which are revealed when we take a closer look at making the representative latent space discriminative based on these fine-grained details. Over time, works have addressed some of these challenges like its highly abstract nature \cite{sain2020cross}, or the huge diversity in its style of drawing \cite{sain2021stylemeup}, scarcity of training sketch-data \cite{pang2019generalising, pang2020solving}, and the wide domain gap sketches have from images \cite{dey2019doodle}, etc. Delving deeper we discover that although significant, these works have barely scratched the surface of the huge potential this topic holds. To put sketch up as a commercially viable method of retrieving images, and establish sketch's significance as a modality in real-world applications, there are a few more challenges from the practicality viewpoint, that need to be addressed in context of visual understanding of sketches. For instance, how can we attune to users who are reluctant to sketch as they are not quite good at it; or say how can we use sketch as a modality to alleviate scenarios where photos are unavailable due to privacy concerns, etc. Overcoming such challenges is imperative for moving forward in this direction.

The goal of this thesis therefore, is to investigate further into the sketch-based visual understanding problem from the angle of practicality.
\cut{ In other words, just as practicality refers to the actual action or experience of something, instead of just introspecting theories or ideas on it, this thesis primarily focuses on exploring and overcoming the hurdles that rise when FG-SBIR paradigm is actually deployed in the real-world on a global scale.}
\cut{ instead of theoretical advancements only}
From a high-level motivational perspective, the \emph{five contributing chapters} of this thesis follows \emph{two major themes}. \aneeshan{First}, we push the boundaries of fine-grained SBIR literature towards a more practical, convenient, and cost-effective deployment. The second theme caters to a more general area focusing on adapting sketch-based visual understanding to a wide range of tasks like unsupervised sketch representation learning and few-shot incremental learning. 

\section{Problem statements and Solutions}
Towards making sketch-based visual understanding more practical, we have identified five challenges based on the current literature, and have proposed solutions for the same. Out of them, the first three are centred around the popular problem of fine-grained sketch-based image retrieval. As our fourth problem,  we delve into designing a self-supervised task for sketches to reduce the bottleneck of limited dataset for sketch-based visual understanding tasks. Finally, we introduce a novel sketch-based visual understanding task as sketch-based few-shot class incremental learning as our fifth problem statement. In this subsection, we briefly describe the five problem statements and their respective solutions respectively.  

\subsection{On-the-fly Fine-Grained SBIR}
\noindent \textbf{Problem Statement:}  
The widespread applicability of fine-grained sketch-based image retrieval (FG-SBIR) is  hindered by two factors -- drawing a sketch takes time, and most people struggle to draw a complete and faithful sketch.
Firstly, while sketch can convey fine-grained appearance details more easily than text \cite{gu2018look, wang2019camp}, drawing a complete sketch is slow compared to clicking a tag \cite{gattupalli2019weakly} or typing a search keyword. Secondly, although state-of-the-art vision systems are good at recognising badly drawn sketches \cite{sangkloy2016sketchy,yu2016sketchAnet}, users who perceive themselves as someone who “can’t sketch” worry about getting details wrong and receiving inaccurate results.
\vspace{0.2cm}  

\noindent \textbf{Solution:} 
We reformulate the conventional FG-SBIR framework to tackle these challenges, with the ultimate goal of retrieving the target photo with the least number of strokes possible. We further propose an on-the-fly design that starts retrieving as soon as the user starts drawing. To accomplish this, we devise a reinforcement learning based cross-modal retrieval framework that directly optimises rank of the ground-truth photo over a complete sketch drawing episode. Additionally, we introduce a novel reward scheme that circumvents the problems related to irrelevant sketch strokes, and thus provides us with a more consistent rank list during the retrieval. We achieve superior early-retrieval efficiency over state-of-the-art methods and alternative baselines on two publicly available fine-grained sketch retrieval datasets.

\subsection{Semi-Supervised Fine-Grained SBIR}
\noindent \textbf{Problem Statement}  
A fundamental challenge faced by existing Fine-Grained Sketch-Based Image Retrieval (FG-SBIR) models is that of \textit{data scarcity} – model performances are largely bottlenecked by the lack of sketch-photo pairs. Whilst the number of photos can be easily scaled, each corresponding sketch still needs to be individually produced. Recent FG-SBIR works \cite{yu2016sketch, pang2019generalising, bhunia2020sketch, pang2020solving} predominately rely on \textit{fully-supervised} triplet-loss based deep networks to yield retrieval performances of practical value. The underlying assumption is largely inline with the progression of supervised photo-only models -- that one can always (relatively easily) obtain additional labelled training data to sustain desired performance gains. This assumption however does not hold for FG-SBIR -- sketch-photo pairs can not be easily scaled as per their photo-only counterparts. That is, instead of crawling and then labelling photos, the corresponding sketch for any given photo will need to be separately drawn by hand. As a result, current FG-SBIR datasets still remain in their thousands (6.7K for QMUL-ShoeV2 \cite{yu2016sketch, song2017deep}, and 2K for QMUL-ChairV2 \cite{song2017deep}), while photo-datasets \cite{russakovsky2015imagenet} are available in millions. This data scarcity problem has consequently resulted in very recent attempts that aim at designing generalisable and zero-shot models \cite{pang2019generalising}, yet performances of these models remain far away from fully-supervised alternatives. Therefore, we aim to mitigate such an upper-bound on sketch data, and study whether unlabelled photos alone (of which they are many) can be cultivated for performance gain.
\vspace{0.2cm} 

\noindent \textbf{Solution}
We introduce a novel semi-supervised framework for cross-modal retrieval that can additionally leverage large-scale unlabelled photos to account for data scarcity. At the center of our semi-supervision design is a sequential photo-to-sketch generation model that aims to generate paired sketches for unlabelled photos. We further introduce a discriminator-guided mechanism to guide against unfaithful generation, together with a distillation loss-based regularizer to provide tolerance against noisy training samples. Last but not least, we treat generation and retrieval as two conjugate problems, where a joint learning procedure is devised for each module to mutually benefit from each other. Extensive experiments show that our semi-supervised model yields a significant performance boost over the state-of-the-art supervised alternatives, as well as existing methods that
can exploit unlabelled photos for FG-SBIR.    

\subsection{Noise-Tolerant Fine-Grained SBIR}
    \noindent \textbf{Problem Statement}      Sketching enables many exciting applications, notably, image retrieval. 
    Despite great strides made on fine-grained SBIR \cite{bhunia2021more, pang2019generalising, PartialSBIR}, the \textit{fear-to-sketch} has proven to be fatal for its omnipresence -- a \emph{``I can't sketch"} reply is often the end of it. This \emph{``fear"} is predominant for fine-grained SBIR (FG-SBIR), where the system dictates users to produce even more faithful and diligent sketches than that required for category-level retrieval \cite{collomosse2019livesketch}. Through some pilot study it has been found that, in most cases it is not about how bad a sketch is -- most \textit{can} sketch (even a rough outline) -- the devil lies in the fact that users typically draw irrelevant (noisy) strokes that are detrimental to the overall retrieval performance. This observation has largely inspired us to alleviate the \emph{``can't sketch"} problem by \emph{eliminating} the noisy strokes through selecting an optimal subset that \textit{can} lead to effective retrieval. 
\vspace{0.2cm}    

\noindent \textbf{Solution} In this chapter, we tackle this \emph{``fear"} head-on and propose for the first time a \emph{pre-processing} module for FG-SBIR that essentially let the users sketch without the worry of \emph{``I can't''}. We consequently design a stroke subset selector that {detects noisy strokes, leaving only those} which make a positive contribution towards successful retrieval. Our Reinforcement Learning based formulation quantifies the importance of each stroke present in a given subset, based on the extent to which that stroke contributes to retrieval. When combined with pre-trained retrieval models as a pre-processing module, we achieve a significant gain of 8\%-10\% over standard baselines and in turn report new state-of-the-art performance. Last but not least, we demonstrate that the selector once trained, can also be used in a plug-and-play manner to empower various sketch applications in ways that were not previously possible.


\subsection{Self-Supervised Learning for Sketches}
    \noindent \textbf{Problem Statement}   In order to alleviate the data annotation bottleneck, many unsupervised methods \cite{noroozi2016unsupervised, kingma2013auto, doersch2015unsupervised, caron2018deep, he2020momentum} propose to pre-train a good feature representation from large scale unlabelled data. A common approach is to define a \emph{pre-text task} whose labels can be obtained free-of-cost, e.g. colorization \cite{zhang2016colorful}, jigsaw solving \cite{noroozi2016unsupervised}, image rotation prediction \cite{gidaris2018unsupervised}, etc. Self-supervised learning has gained prominence due to its efficacy at learning powerful representations from unlabelled data that achieve excellent performance on many challenging downstream tasks. The motivation is that a network trained to solve such a pre-text task should encode high-level semantic understanding of the data that can be used to solve other downstream tasks like classification, retrieval, etc. Apart from traditional object classification, detection or semantic segmentation,  self-supervision has been extended to sub-domains like human pose-estimation \cite{kocabas2019self}, co-part segmentation \cite{hung2019scops}, and depth estimation \cite{godard2019digging}.   Supervision-free pre-text tasks are challenging to design and usually modality specific. Although there is a rich literature of self-supervised methods for either spatial (such as images) or temporal data (sound or text) modalities, a common pre-text task that benefits both modalities is largely missing. Therefore, we seek to introduce a self-supervised method for a class of visual data that is distinctively different than photos: sketches \cite{yu2016sketchAnet, sangkloy2016sketchy} and handwriting \cite {poznanski2016cnn} images.  

      \vspace{0.2cm}          
\noindent \textbf{Solution} In this chapter, we are interested in defining a self-supervised
pre-text task for sketches and handwriting data. This data is uniquely characterised by its existence in dual modalities of rasterized images and vector coordinate sequences. We address and exploit this dual representation by proposing two novel cross-modal translation pre-text tasks for self-supervised feature learning: Vectorization and Rasterization. Vectorization learns to map image space to vector coordinates and rasterization maps vector coordinates to image space. We show that our learned encoder modules benefit both raster-based and vector-based downstream approaches to analysing hand-drawn data. Empirical evidence shows that our novel pre-text tasks surpass existing single and multi-modal self-supervision methods. We motivate our pre-text task in the the context of the handwriting recognition literature of late ’90s. There are numerous works  \cite{kato2000recovery} in top venues that aim to retrieve the online trajectory information from offline handwritten images. The idea is that retrieving the time information and combining it with spatial images could improve recognition accuracy. We rejuvenate this idea in context of representation learning to define a pre-text task for both sketch and handwriting to distil underlying properties of the data from its dual representation. While handwritten trajectory retrieval used to be handled via classical methods (graph, HMM, etc), we develop a neural network solution for sketch-image to sketch-vector translation that is end-to-end trainable

\subsection{Sketch-Based Incremental Learning}
\noindent \textbf{Problem Statement}
Fully supervised learning has served us  with great performances on ImageNet already surpassing human-level \cite{he2015delving}. In reality, however, such progress is primarily limited to a small number of object classes where labels were explicitly curated (1000 in ImageNet vs. possibly millions out there).  The human visual system is remarkable in learning new visual concepts from just a few examples. This is precisely the goal behind few-shot class incremental learning (FSCIL), where the emphasis is additionally placed on ensuring that the model does not suffer from “forgetting”. Class Incremental Learning \cite{li2017learning, hsu2018re, kirkpatrick2017overcoming} is one of the popular fronts that attempt to extend model perception to novel classes while not ``forgetting" about classes learned already. Amongst its many variants, the very recent Few-Shot Class Incremental Learning (FSCIL) \cite{tao2020few} is the most realistic where it also dictates the model to learn new classes with {{very few}} examples, the same as humans do.
     
     As easy as providing a few samples might sound, questions start to emerge in practice as to (i) what data modality should the samples take? and (ii) how could these samples be obtained in practice. These questions, we argue, are key to the potentially ubiquitous application of FSCIL as (i) humans also learn from a broad range of data modalities that are not limited to just photo, and (ii) there are scenarios where photos are not necessarily always readily available due to privacy and ethical constraints (e.g., copyright). 
     
     \vspace{0.2cm} 
     
    \noindent \textbf{Solution}
We push the boundary further for FSCIL by addressing two
key questions that bottleneck its ubiquitous application (i)
can the model learn from diverse modalities other than just
photo (as humans do), and (ii) what if photos are not readily
accessible (due to ethical and privacy constraints). Our
key innovation lies in advocating the use of sketches as a
new modality for class support. The product is a “Doodle
It Yourself” (DIY) FSCIL framework where the users can
freely sketch a few examples of a novel class for the model to
learn to recognise photos of that class. For that, we present
a framework that infuses (i) gradient consensus for domain
invariant learning, (ii) knowledge distillation for preserving
old class information, and (iii) graph attention networks for
message passing between old and novel classes. We experimentally
show that sketches are better class support than
text in the context of FSCIL, echoing findings elsewhere in
the sketching literature. 
    
    \section{Thesis outline}
    The thesis is organized as follows:    

    \begin{description}
        \item[Chapter \ref{chapter:literaturereview}] presents the existing literature on sketch-based visual understanding, sketch-based image retrieval, and sketch-representation learning. Furthermore, we briefly summarise some generic computer vision literature like semi-supervised learning, self-supervised learning, incremental learning, reinforcement learning for vision tasks etc., that intersects with our sketch-based visual understanding research at large.
        
         \item[Chapter \ref{chapter:OTF}] reformulates the conventional FG-SBIR framework via an \textit{on-the-fly} design that starts retrieving as soon as the user starts drawing. To do so a reinforcement learning based cross-modal retrieval framework is designed that directly optimises rank of the ground-truth photo over a complete sketch drawing episode.

         \item[Chapter \ref{chapter:SSL_FGSBIR}] introduces a novel semi-supervised framework for cross-modal retrieval that can additionally leverage large-scale unlabelled photos to account for data scarcity issue for fine-grained sketch-based image retrieval. While it includes a a sequential photo-to-sketch generation model, that aims to generate paired sketches for unlabelled photos, a discriminator-guided mechanism is designed to guide against unfaithful generation, together with a distillation loss-based regularizer to provide tolerance against noisy synthetic training samples.
            
        \item[Chapter \ref{chapter:NT_FGSBIR}]  starts with a pilot study that reveals that the noisy/irrelevant strokes drawn by the users are a bottleneck for fine-grained SBIR. Therefore, a reinforcement learning based stroke subset selector is designed that detects noisy strokes, leaving only those which make a positive contribution towards successful retrieval. Furthermore, this provides a way to quantify the importance of each stroke present in a given subset, based on the extent to which that stroke contributes to retrieval.

         \item[Chapter \ref{chapter:Self}] defines a self-supervised pre-text task for sketches and handwriting data that  are uniquely characterised by their existence in dual modalities of rasterized images and vector coordinate sequences. This dual representation is exploited  through two novel cross-modal translation pre-text tasks for self-supervised feature learning: Vectorization and Rasterization.
         
         \item[Chapter \ref{chapter:SBIL}] introduces a novel sketch-based few-shot incremental learning problem. This pushes the boundary further for FSCIL by addressing two key questions that bottleneck its ubiquitous application (i) can the model learn from diverse modalities other than just photo (as humans do), and (ii) what if photos are \textit{not readily accessible} (due to ethical and privacy constraints). This is achieved by introducing sketches as class support for FSCIL, allowing the system to learn from modalities other than just photos and addressing issues around ethics and privacy while allowing user creativity.

        \item[Chapter \ref{chapter:conclusionAndFutureWork}] provides a summary of the research outputs through a conclusion and discusses various potential future research directions that could be carried out as further works. 

        \end{description}

\chapter{Literature Review}
    \label{chapter:literaturereview}
    \minitoc
    \newpage

        {This chapter starts by discussing the overview of sketch literature and briefly introduces the influence of sketch in computer vision literature for different visual understanding tasks. It then talks about specific sub-topics of sketch literature alongside various research topics from computer vision literature that overlaps with the problems that we are dealing with in this thesis. Following this overview of sketch literature, we delve into the developments in sketch representation learning in the recent deep-learning era.
        Three out of five problem statements in this thesis deal with how Sketch-Based Image Retrieval (SBIR) could be made more practical -- both in terms of ease of training, and convenience of the user, from an application point of view. We therefore categorise the topic of Sketch-Based Image Retrieval (SBIR), into two subsections for category-level SBIR and fine-grained SBIR, respectively. Being one of the crucial concepts for on-the-fly fine-grained SBIR and Noise-Tolerant SBIR, we next discuss the relevant literature dealing with Partial Sketches. Subsequently, we outline the earlier attempts to deal with the data-scarcity issue of fine-grained SBIR and how photo-to-sketch generation could be a straightforward way to augment training data for unlabelled photos. Finally, we summarise the relevant literature that intersects with general computer vision literature. For instance, as Reinforcement Learning (RL) based training is involved in three of our problems, we discuss how RL has been adopted in various computer vision tasks. Furthermore, the generic concept of Semi-supervised Learning, Self-supervised Learning, Incremental Learning and the idea of Learning from Noisy-Labels are directly relevant to our problem of Semi-supervised fine-grained SBIR, Self-supervised learning on sketches, Sketch-Based Incremental Learning and Noise-Tolerant SBIR, respectively whose literature has been discussed accordingly.}

    
\section{Sketch for Visual Understanding}

Hand-drawn sketches, by nature, are enriched with the various human visual system like visual understanding abilities and are very close to the cognitive-subconscious of human intelligence \cite{hertzmann2020line}. Consequently, it has facilitated various visual understanding tasks in the past. Apart from the widely studied Sketch-Based Image Retrieval \cite{collomosse2019livesketch, yelamarthi2018zero, dey2019doodle, sampaio2020sketchformer, sain2021stylemeup, pang2019generalising, song2017deep, bhunia2020sketch, bhunia2021more, bhunia2022adaptive, chowdhury2022fs}, sketch has also been employed in a variety of vision understanding tasks, including realistic image generation \cite{ghosh2019interactive}, image-editing \cite{yang2020deep}, segmentation \cite{hu2020sketch}, video synthesis \cite{li2021deep}, representation learning \cite{wang2021sketchembednet}, object localisation \cite{tripathi2020sketch}, image-inpainting \cite{xie2021exploiting}, 3D shape retrieval \cite{luo2020towards}, 3D shape modelling \cite{zhang2021sketch2model}, among others \cite{xu2020deep}. Due to its commercial importance, sketch-based image retrieval (SBIR) \cite{dutta2019semantically, sain2021stylemeup, collomosse2019livesketch, yelamarthi2018zero, shen2018zero} has seen considerable attention within sketch-related research where the objective is to learn a joint-embedding space through a CNN \cite{bhunia2020sketch}, RNN \cite{collomosse2019livesketch}, Transformer \cite{sampaio2020sketchformer} or their combinations \cite{collomosse2019livesketch} for retrieving image(s) based on a sketch query.  In particular, sketch has been a successful query medium due to its fine-grained  attribution ability over text/tags \cite{song2017fine}. In recent times, sketches have been employed to design aesthetic pictionary-style gaming \cite{sketchxpixelor} and to mimic the creative angle \cite{ge2020creative} of human understanding. 
Growing research on the application of sketches to visual interpretation tasks establishes the representative \cite{wang2021sketchembednet} as well as discriminative \cite{sain2021stylemeup} ability of sketches to characterise the corresponding visual photo. These establish the fact that hand-drawn sketches have enough representative ability to characterise a visual photo efficiently. 
{Set upon this fact, we aim to explore how \emph{sketch} can act as a potential substitute to conventional photos in class-incremental learning  (Chapter \ref{chapter:SBIL}), in addition to advancing the traditional framework of Sketch-Based Image Retrieval (Chapter \ref{chapter:OTF}, \ref{chapter:SSL_FGSBIR} and \ref{chapter:NT_FGSBIR}).}


\section{Sketch Representation Learning}   
    
Learning a robust sketch representation benefits a variety of sketch-specific problems like classification \cite{yu2016sketch}, retrieval \cite{xu2018sketchmate}, scene understanding \cite{liu2020scenesketcher}, sketch-based image retrieval \cite{bhunia2020sketch, pang2019generalising, pang2020solving, dey2019doodle, dutta2019semantically, sain2020cross}, generative sketch modelling \cite{song2018learning, ha2017neural, sketchxpixelor} etc. While photos are pixel perfect depictions represented by 2D spatial matrices, sketches can be described either as 2D static pixel level \emph{rasterized images} or vector sketches with an ordered sequence of \emph{point coordinates}. Typically, sketch images are processed by convolutional neural networks \cite{bhunia2020sketch, dey2019doodle}, whereas vector sketches need Recurrent Neural Networks (RNNs) or Transformers \cite{xu2018sketchmate, sampaio2020sketchformer, sketchbert2020} for sequential modelling. There exists no consensus on which sketch modality (image or vector) is better than the other, as each has its own merits  based on the application scenario. While rasterized sketch images are usually claimed to be better for driving  fine-grained retrieval \cite{yu2016sketch, song2017deep}, they fail to model the varying level of abstraction \cite{song2018learning} in the sketch generative process. Conversely, vector sketches are more effective to simulate the human sketching style \cite{ha2017neural} for generation, however, it fails \cite{bhunia2020sketch} for fine-grained instance level image retrieval. From a  computational cost perspective, coordinate based models provide faster cost-effective real-time performance \cite{xu2019multi} for sketch-based human-computer interaction, compared to using rasterized sketch images that impose a costly rendering step and transfer cost of a large pixel array. Attempts have been directed towards combining representations from both sketch images and vector sketches for improved performance in sketch hashing and category level sketch-based image retrieval \cite{xu2018sketchmate}. 

Image-Net pretrained weights are widely used to initialise standard convolutional networks for sketch images, with the first self-supervised alternatives specifically designed for raster sketch images being proposed recently \cite{pang2020solving}. Vector sketches relying on RNN or Transformer do not have the ImageNet initialization option. Thus, Lin \etal \cite{sketchbert2020} employ BERT-like self-supervised learning on vector sketches.
Nevertheless, the existing self-supervised methods are proposed for specific modalities (raster image vs vector sketches) and do not generalize onto each other. We therefore propose a unified pipeline (Chapter \ref{chapter:Self}) that leverages this \emph{dual representation of sketches} to learn powerful features for encoding sketches represented in both vector and raster views.


\section{Category-Level SBIR}
Category-level SBIR aims at retrieving category-specific photos from user given query sketches. Like any other retrieval system, Deep Neural Networks have become a de-facto choice for any recent SBIR frameworks \cite{dutta2019semantically, dey2019doodle, sampaio2020sketchformer, zhang2018generative, collomosse2019livesketch} over early hand-engineered feature descriptors \cite{tolias2017asymmetric}.  Category-level sketch-photo retrieval is now well studied \cite{cao2011edgel, wang2015sketch, bui2018deep, dey2019doodle, collomosse2017sketching, yelamarthi2018zero, shen2018zero, dutta2019semantically, liu2019semantic, collomosse2019livesketch, liu2017deep, Sketch3T}. Overall, category level SBIR makes use of Siamese networks based on either CNN \cite{collomosse2019livesketch, dey2019doodle}, RNN \cite{xu2018sketchmate}, Transformer \cite{sampaio2020sketchformer} or their combinations \cite{collomosse2019livesketch} along with a triplet-ranking objective to learn a joint embedding space. A distance metric is used to rank the gallery photos against the learned embedding space for a given query sketch for retrieval. 

Contemporary research directions can be broadly classified into traditional SBIR, zero-shot SBIR and sketch-image hashing. In traditional SBIR \cite{cao2011edgel, collomosse2017sketching, collomosse2019livesketch, bui2018deep}, object classes are common to both training and testing. Whereas zero-shot SBIR \cite{yelamarthi2018zero, dey2019doodle, dutta2019semantically, liu2019semantic} asks models to generalise across disjoint training and testing classes in order to alleviate annotation costs. Sketch-image hashing \cite{liu2017deep, shen2018zero} aims to improve the computational cost of retrieval by embedding to binary hash-codes rather than continuous vectors.

While these SBIR works assume a single-step retrieval process, a recent study by Collomosse \etal \cite{collomosse2019livesketch} proposed an interactive SBIR framework.
Given an initial sketch query, if the system is unable to retrieve the user's goal in the first try, it resorts to providing some relevant image clusters to the user. The user can now select an image cluster in order to disambiguate the search, based on which the system generates new query sketch for following iteration. This interaction continues until the user's goal is retrieved. This system used Sketch-RNN \cite{ha2017neural} for sketch query generation after every interaction. However Sketch-RNN is acknowledged to be weak in \emph{multi-class} sketch generation  \cite{ha2017neural}. As a result, the generated sketches often diverge from the user's intent leading to poor performance. Note that though such interaction through clusters is reasonable in the case of \emph{category}-level retrieval, it is not applicable to our FG-SBIR task where all photos belong to a single class and differ in subtle ways only.

\section{FG-SBIR}
SBIR was initially posed as a category-level retrieval problem. However, it became apparent that the key advantage of sketch over text/tag-based retrieval was conveying \emph{fine-grained} details \cite{engilberge2018finding} -- leading to a focus on fine-grained SBIR that aims to retrieve a \emph{particular} photo within a gallery. FG-SBIR is a more recent addition to sketch analysis and also less studied compared to the category-level SBIR task until recently. One of the first studies  \cite{li2014fine} addressed it by graph-matching of deformable-part models. A number of deep learning approaches subsequently emerged \cite{yu2016sketch, song2017deep, pang2017cross, pang2019generalising, sain2020cross}. Yu \etal \cite{yu2016sketch} proposed a deep triplet-ranking model for instance-level FG-SBIR. This paradigm was subsequently improved through hybrid generative-discriminative cross-domain image generation \cite{pang2017cross}; and providing an attention mechanism for fine-grained details as well as more sophisticated triplet losses \cite{song2017deep}. Recently Pang \etal \cite{pang2019generalising} studied cross-category FG-SBIR in analogy to the `zero-shot' SBIR mentioned earlier. Mixed modality jigsaw \cite{pang2020solving} solving has been used as a self-supervised pre-text task for fine-grained SBIR over Image-Net based pretrained models. 


{In this thesis, we have made three significant contributions towards the design of fine-grained SBIR. Firstly, we explore framework design of FG-SBIR towards forging a new research direction (Chapter \ref{chapter:OTF}) towards \emph{on-the-fly and early} photo retrieval. 
Secondly, we introduce a novel semi-supervised FG-SBIR framework  (Chapter \ref{chapter:SSL_FGSBIR}) that can additionally leverage large-scale unlabelled photos to account for data scarcity. 
Lastly, we go deep into understanding the challenges arising from irrelevant/noisy strokes drawn by the users, which is a common issue amongst people not confident in drawing. With an aim to alleviate this, we design a stroke subset selector (Chapter \ref{chapter:NT_FGSBIR})  that detects noisy strokes, leaving out only those that positively contribute to successful retrieval.}


\section{Modelling Partial Sketches}

``Sketch'' being an interactive medium, is drawn sequentially in a stroke-by-stroke manner. Moreover, due to its subjective nature, the same sketch might be perceived as partial or complete based on the user's perception. Users can retrieve photos \cite{bhunia2020sketch}, create \cite{wang2021sketch} imaginative visual-art, or edit existing photos \cite{jo2019sc} through repeated interactions with the AI agent. Therefore, on-the-fly interaction with sketches requires sketch-based models to be capable of handling partial sketches.   One of the most popular areas for studying incomplete or partial data is image inpainting \cite{yu2018generative, zheng2019pluralistic}. Significant progress has been made in this area using contextual-attention \cite{yu2018generative} and Conditional Variational Autoencoder (CVAE) \cite{zheng2019pluralistic}. Following this direction, works have attempted to model partial sketch data \cite{liu2019sketchgan, ha2017neural, ghosh2019interactive}. Sketch-RNN \cite{ha2017neural} learns to predict multiple possible endings of incomplete sketches using a Variational Autoencdoer (VAE). While Sketch-RNN works on sequential pen-coordinates, Liu \etal \cite{liu2019sketchgan} extend conditional image-to-image translation to rasterized sparse sketch domain for partial sketch completion, followed by an auxiliary sketch recognition task. 
Ghosh \etal \cite{ghosh2019interactive} proposed an interactive sketch-to-image translation method, which completes an incomplete object outline, and thereafter it generates a final synthesised image. 
Overall, these works first try to complete the partial sketch by modelling a conditional distribution based on image-to-image translation, and subsequently focus on specific task objective, be it sketch recognition or sketch-to-image generation.
{Unlike these two-stage inference frameworks, in Chapter \ref{chapter:OTF} , we focus on instance-level photo retrieval with a minimum number of sketch strokes, thus enabling partial sketch queries in a single step. In Chapter \ref{chapter:NT_FGSBIR}, we take a step further and aim to answer the question \emph{``whether a partial sketch has sufficient representative information/discriminative potential to retrieve photos faithfully''}. Additionally, we aim to quantify the instant at which a sequentially drawn sketch would reach the optimum threshold point where it is representative enough for downstream tasks (e.g., retrieval). By doing so, we can faithfully train models with sufficiently representative partial sketches instead of randomly dropping strokes and ignoring instances where the synthetic partial sketch is too coarse to convey any meaning.}

 \section{Data-Scarcity for FG-SBIR}

Earlier works have tried resolving the lack of instance-level photo-sketch paired data, by using edge-maps for training \cite{radenovic2018deep} or synthetic sketch stroke deformation \cite{yu2016sketchAnet, yu2016sketch} for data augmentation. Umar \etal \cite{riaz2018learning, muhammad2019goal} leveraged reinforcement-learning (RL) in an attempt to augment sketches from edge-maps under the assumption that real sketch-strokes are a subset of edge-maps, which however is negated by the highly abstracted nature of real sketches. Very recently, mixed-modality jigsaw solving \cite{pang2020solving} has been used as a pre-training task for FG-SBIR to exploit additional photo images and their edge maps for cross-modal matching. Its efficacy remains limited however as edge-maps are not sketches.  In Chapter \ref{chapter:SSL_FGSBIR}, we have tried to address the data-scarcity issue in FG-SBIR through a semi-supervised learning framework. 

\section{Photo-to-Sketch Generation}

A plausible solution to data scarcity for fine-grained SBIR through semi-supervised learning (Chapter \ref{chapter:SSL_FGSBIR}) is synthesising sketches for unlabelled photos to form pseudo photo-sketch pairs. Existing photo-to-sketch generation methods can be classified into two types: the first employs image-to-image translation  \cite{li2019photo}, which however merely works as a contour detection paradigm, thus failing to model the hierarchically abstracted     nature of   human-drawn sketch. The second follows the seminal work of Sketch-RNN \cite{ha2017neural}, and generates sequential sketch-coordinates given a photo, thus mimicking subjective human sketching style. The basic design \cite{chen2017sketch}, involving a CNN encoder and RNN decoder, has been further augmented with both self-domain and two way cross-modal reconstruction losses \cite{song2018learning}. 
    

\section{Reinforcement Learning in Vision}
       
Reinforcement Learning (RL) \cite{kaelbling1996reinforcement} has been applied in different vision problems \cite{liang2017deep, wang2019reinforced}. 
Lately there has been a growing interest in applying reinforcement learning (RL) \cite{kaelbling1996reinforcement} to computer vision problems \cite{liang2017deep, wang2019reinforced}. 
{In the scenario of existent non-differentiability in a training framework, RL comes in quite handy to circumvent it and quantify the \textit{goodness} of the network's state.}
Instead, learning progresses via interactions \cite{duong2019automatic, han2019deep} with the environment. Particularly in the sketch community, RL has been leveraged for modelling sketch abstraction \cite{muhammad2018learning,muhammad2019goal}, retrieval \cite{bhunia2020sketch, bhunia2021more}, and  designing competitive sketching agent \cite{sketchxpixelor}. {In our Chapters \ref{chapter:OTF}, \ref{chapter:SSL_FGSBIR} and \ref{chapter:NT_FGSBIR}, we have thus adopted RL based training to tackle the non-differentiability issue in the learning paradigm.}


\section{Semi-supervised Learning}
In Chapter \ref{chapter:SSL_FGSBIR}, our learning framework is semi-supervised in the sense that the majority of training data are unlabelled photos without their paired sketches. This makes it quite different from most existing semi-supervised learning methods which are designed for classification rather than cross-modal retrieval. Therefore these methods, based on either  pseudo-labelling  \cite{lee2013pseudo,muller2019does,sohn2020fixmatch,pham2020meta} or consistency regularisation \cite{berthelot2019mixmatch, berthelot2020remixmatch}, offer little insight towards solving our problem.  
{On the contrary, prior works on semi-supervised cross-modal learning such as image captioning \cite{chen2016semi, kim2019image} are more relevant to our semi-supervised FG-SBIR framework in Chapter \ref{chapter:SSL_FGSBIR}, where we uniquely address a cross-modal instance-level retrieval problem, and jointly train a sketch-generator along with the retrieval model, rather than merely providing model pre-training.}


    
\section{Self-supervised Learning}
    
Self-supervised learning is now a large field, too big to review in detail here, with recent surveys \cite{jing2020selfSup} providing a broader overview. Reviewing briefly, generative models such as VAEs \cite{kingma2013auto} learn representations by modelling the distribution of the data. Contrastive learning \cite{grill2020bootstrap, chen2020simple, he2020momentum} aims to learn discriminative features by minimising the distance between different augmented views of the same image while maximizing it for views from different images. Clustering based approaches \cite{caron2018deep} on the other hand, first clusters the data based on the features extracted from a network, and re-trains the same network thereafter using the cluster-index as pseudo-labels for classification.

Different pre-text tasks have been explored for self-supervised feature learning in imagery, e.g., image colorization \cite{zhang2016colorful},  super-resolution \cite{ledig2017photo}, solving jigsaw puzzles \cite{noroozi2016unsupervised,pang2020solving},  in-painting \cite{pathak2016context}, relative patch location prediction \cite{doersch2015unsupervised}, frame order recognition \cite{misra2016shuffle}, etc. Compared to these approaches, our work is more similar to the few approaches addressing multi-modal data. For instance,  pre-text tasks like visual-audio correspondence  \cite{arandjelovic2017look,korbar2018cooperative}, or RGB-flow-depth correspondence \cite{tian2019contrastiveCMC} within vision. However, these approaches use contrastive losses, which raise a host of complex design issues in batch size, batch sampling strategies, and positive/negative balancing \cite{chen2020simple,patacchiola2020selfsupervised, he2020momentum, henaff2019data, tian2019contrastiveCMC} that are necessary to tune, in order to obtain good performance. Furthermore they tend to be extremely expensive to scale due to the ultimately quadratic cost of comparing sample \emph{pairs} \cite{chen2020simple,patacchiola2020selfsupervised, tian2019contrastiveCMC, li2020prototypical}. In contrast, our work on self-supervised learning on sketches avoids all of these design issues and compute costs through a simple cross-modal synthesis task, while achieving state of the art performance in both vector and raster view downstream performance. We attribute this to the transformation between pixel-array and vector-sequence models of sketches being a much more challenging mapping to learn, and thus providing a stronger representation in both modalities. 


\section{Learning from Noisy-Labels}
Despite significant progress from the community-generated labelled data, accurate labelling is still challenging even for experienced domain experts \cite{song2021learning}. Therefore, a separate topic of study \cite{song2021learning,zhou2021robust,zhou2021robust} emerged, which aims at learning robust models even from the noisy data distribution. Despite significant progress from the community-generated labelled data, accurate labelling is challenging even for experienced domain experts \cite{song2021learning}. The existing works \cite{han2018masking,tanno20219learning} mainly consider having access to a large, noisy dataset as well as a subset of carefully cleaned data for validation. 
{The situation intensifies further in Chapter \ref{chapter:NT_FGSBIR}, where we assume that every annotated sketch is not an absolutely perfect matching sketch of the paired photo. We therefore aim to develop a noise-tolerant framework for FG-SBIR.}

\section{Incremental Learning}

Incremental Learning (IL) \cite{polikar2001learn++, kuzborskij2013n} is a machine learning paradigm where a model adapts itself to learn new tasks sequentially while retaining the previously learnt knowledge. Although deep networks have demonstrated incredible achievements in a variety of tasks \cite{santoro2016meta, snell2017prototypical}, sequentially learning different tasks remains a key challenge. Consequently, IL continues to receive considerable research attention \cite{hsu2018re, icart2017, chaudhry2018efficient, kirkpatrick2017overcoming, aljundi2018memory}. Majority of the present research either use memory-based \cite{hsu2018re, icart2017}, distillation-based \cite{cheraghian2021semantic, dong2021few}, or regularisation-based \cite{kirkpatrick2017overcoming} approaches to tackle the IL task. Based on the task at hand, IL can be categorised into (a) Incremental domain learning \cite{rosenfeld2018incremental}, which aims at performing incremental domain adaptation. (b) Incremental task learning \cite{aljundi2018memory}, where each task consists of separate classification layers, and a task descriptor selects the appropriate layer during the testing phase. (c) Class incremental learning (CIL), the most challenging IL task that operates in a single-head setup with no available task descriptors. In CIL, the model needs to learn a unified classifier to fit all the new unseen classes incrementally. Distillation \cite{li2017learning} and memory-based \cite{hsu2018re} methods are more effective than regularisation-based ones \cite{kirkpatrick2017overcoming} in the CIL setting. Our work is mainly concerned with CIL setup, which is the most challenging task among its variants.
 
 Few shot learning ({{FSL}}) aims at adapting a trained model to learn patterns from novel classes (unseen during training) using only a \emph{few labelled} samples \cite{wang2020generalizing}. Recently, it has experienced rapid proliferation \cite{rezende2016one, snell2017prototypical, vinyals2016matching} in the research community. There are three major swim lanes of the FSL problem: (a) recurrent-based \cite{rezende2016one, santoro2016meta} (b) optimisation-based \cite{rusu2018meta, vuorio2019multimodal}, and (c) metric-based frameworks \cite{gidaris2018dynamic, koch2015siamese}. Our work falls under metric-based methods in which similarity is drawn between the query sample and the novel support classes. Conventional CIL presumes that the incrementally provided novel classes have access to a substantial amount of labelled data. Although in the FSCIL paradigm \cite{tao2020few}, the initial dataset contains sufficient training data (base classes), the subsequently provided novel classes contain only a few labelled samples. Very few methods are present  to tackle the FSCIL problem like, pseudo incremental learning \cite{zhang2021few}, knowledge distillation \cite{cheraghian2021semantic,dong2021few}, neural-gas network \cite{tao2020few}. While existing works intend to build a model to incrementally learn novel classes, we aim at building a model for a much harder and practically applicable sketch-based FSCIL setting in Chapter  \ref{chapter:SBIL} that addresses user's privacy concerns. 
\chapter{On-the-fly Fine-Grained SBIR}
    \label{chapter:OTF}
    \minitoc
    \newpage
    
    \section{Overview}
    \label{chapter:OTF:Overview}

\aneeshan{Following the first theme our first chapter begins with exploring the potential issues encountered during practical deployment of FG-SBIR system.}
Despite the great progress made in FG-SBIR~\cite{yu2016sketch, song2017deep, pang2019generalising}, two barriers hinder its widespread adoption in practice -- the time taken to draw a complete sketch, and the drawing skill shortage of the user. Firstly, while sketch can convey fine-grained appearance details more easily than text,  drawing a complete sketch is slow compared to clicking a tag or typing a search keyword. Secondly, although state-of-the-art vision systems are good at recognising badly drawn sketches \cite{sangkloy2016sketchy,yu2016sketchAnet}, users who perceive themselves as someone who ``can't sketch'' worry about getting details wrong and receiving inaccurate results.

\begin{figure}[!hbtt]
\begin{center}
  \includegraphics[width=0.75\linewidth]{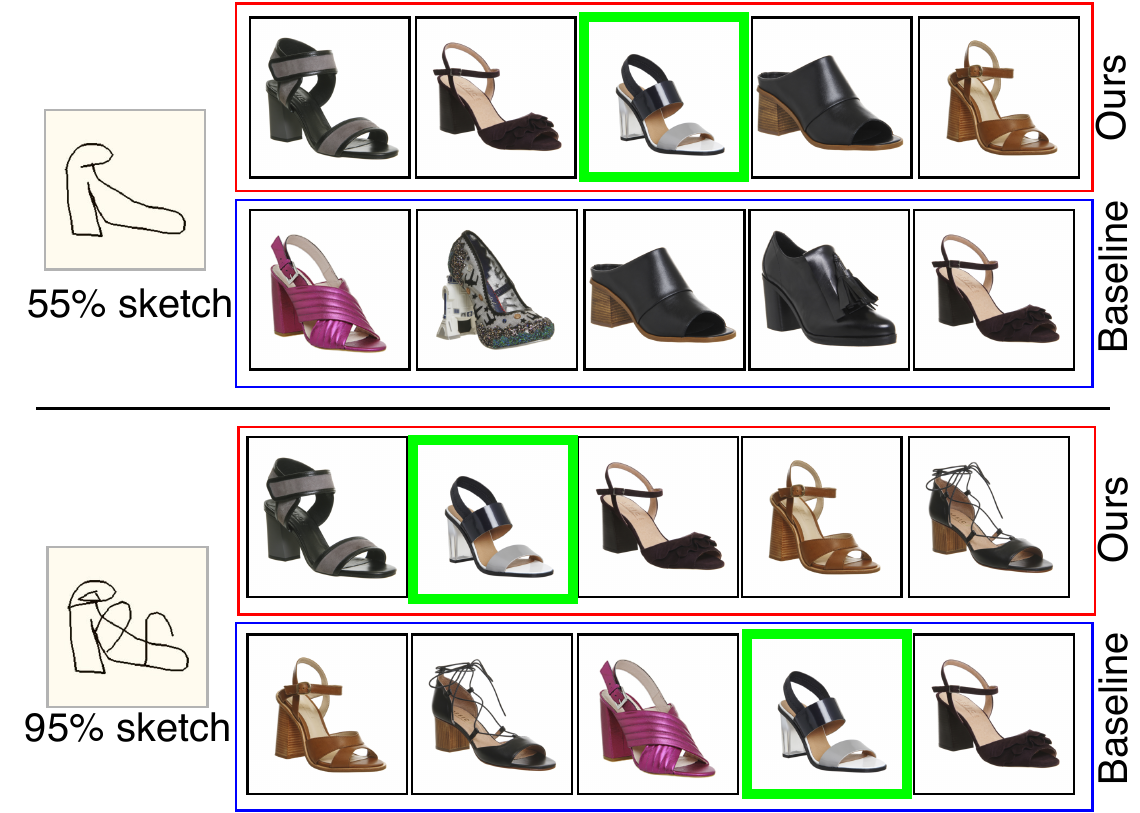} 
\end{center}
\vspace{-0.6cm}
  \caption{Examples showing the potential of our framework that can retrieve (top-5 list) target photo using fewer number of strokes than the conventional baseline method.}
\label{fig:Fig1_a2.pdf}
\end{figure}

In this chapter we break these barriers by taking a view of ``less is more'' and propose to tackle a new fine-grained SBIR problem that aims to retrieve the target photo with just a few strokes, as opposed to requiring the complete sketch. This problem assumes a ``on-the-fly'' setting, where retrieval is conducted at every stroke drawn. Figure \ref{fig:Fig1_a2.pdf} offers an illustrative example of our on-the-fly FG-SBIR framework. Due to stroke-by-stroke retrieval, and a framework optimised for few-stroke retrieval, users can usually ``stop early'' as soon as their goal is retrieved. This thus  makes sketch more comparable with traditional search methods in terms of time to issue a query, and \emph{more easily} -- as those inexperienced at drawing can retrieve their queried photo based on the easiest/earliest strokes possible \cite{berger2013styleAbstraction}, while requiring fewer of the detailed strokes that are harder to draw correctly.

Solving this new problem is non-trivial.  One might argue that we can directly feed incomplete sketches into the off-the-shelf FG-SBIR frameworks \cite{yu2016sketch, sangkloy2016sketchy}, perhaps also enhanced by including synthesised sketches in the training data. However, those frameworks are not fundamentally designed to handle incomplete sketches. This is particularly the case since most of them employ a triplet ranking framework where each triplet is treated as an independent training example. So they struggle to perform well across a whole range of sketch completion points. Also, the initial sketch strokes could correspond to many possible photos due to its highly abstracted nature, thus more likely to give a noisy gradient. Last, there is no specific mechanism that can guide existing FG-SBIR model to retrieve the photo with minimal sketch strokes, leaving it struggling to perform well across a complete sketching episode during on-the-fly retrieval.

A novel on-the-fly FG-SBIR framework is proposed in this chapter. First and foremost, instead of the de facto choice of triplet networks that learn an embedding where sketch-photo pairs lie close, we introduce a new model design that optimizes the rank of the corresponding photo over a sketch drawing episode. Secondly, the model is optimised specifically to return the true match within a minimum number of strokes. Lastly, efforts are taken to mitigate the effect of misleading noisy strokes on obtaining a consistent photo ranking list as users add details towards the end of a sketch.

More concretely, we render the sketch at different time instants of drawing, and feed it through a deep embedding network to get a vector representation.  While the other SBIR frameworks \cite{yu2016sketch,sangkloy2016sketchy} use triplet loss \cite{weinberger2009metric_learn_margin} in order to learn an embedding suited for comparing sketch and photo, we optimise the \emph{rank} of the target photo with respect to a sketch query. By calculating the rank of the ground-truth photo at each time-instant $t$ and maximizing the sum of $\frac{1}{rank_t}$ over a complete sketching episode, we ensure that the correct photo is retrieved as early as possible.
Since ranking is a non-differential operation, we use a Reinforcement Learning (RL) \cite{kaelbling1996reinforcement} based pipeline to achieve this goal. Representation learning is performed with knowledge of the whole sequence, as we optimize the reward non-myopically over the sketch drawing episode. This is unlike the triplet loss used for feature learning that does not take into account the temporal nature of the sketch. We further introduce a global reward to guard against harmful noisy strokes especially during later stages of sketching where details are typically added. This also stabilises the RL training process, and produces  smoother retrieval results.

Our contributions can be summarised as follows: (a) We introduce a novel \textit{on-the-fly} FG-SBIR framework trained using reinforcement learning to retrieve photo using an incomplete sketch, and do so with the minimum possible drawing. (b) To this end, we develop a novel reward scheme that models the early retrieval objective, as well as one based on Kendall-Tau \cite{knight1966computer} rank distance that takes into account the completeness of the sketch and associated uncertainty. (c) Extensive experiments on two public datasets demonstrate the superiority of our framework.

\begin{figure*}[t]
	\begin{center}
		\includegraphics[width=1\linewidth]{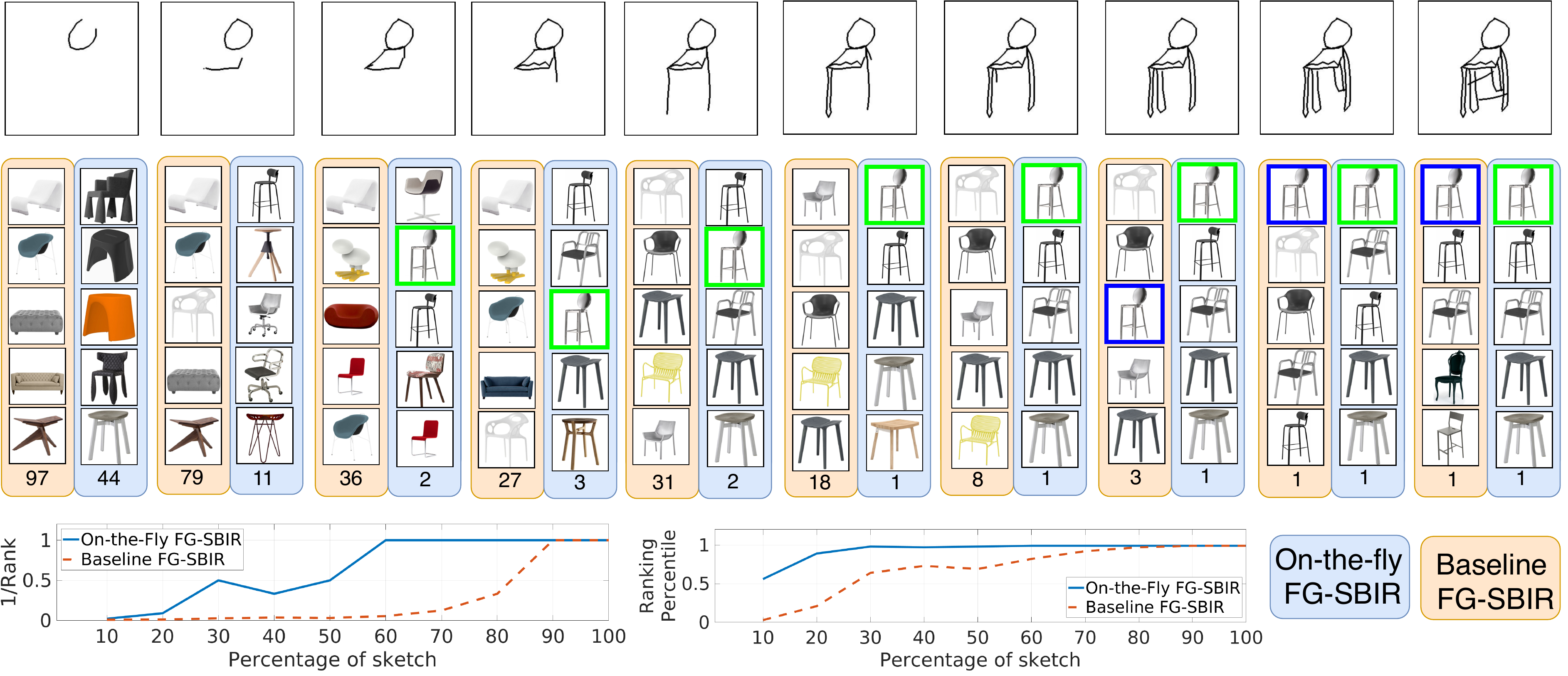} 
	\end{center}
	\vspace{-0.25cm}
	\caption{Illustration of proposed \textit{on-the-fly} framework's efficacy over a baseline FG-SBIR method \cite{song2017deep, yu2016sketch} trained with completed sketches only.  For this particular example, our method needs only $30\%$ of the complete sketch to include the true match in the top-10 rank list, compared to $80\%$ for the baseline. Top-5 photo images retrieved by either framework are shown here, in progressive sketch-rendering steps of $10\%$. The number at the bottom denotes the paired (true match) photo's rank at every stage.}
	\label{fig:Fig1}
	\vspace{-0.2cm}
\end{figure*}

\section{Problem Formulation}

Our objective is to design an `on-the-fly' FG-SBIR framework, where we perform live analysis of the sketch as the user draws. The system should re-rank candidate photos based on the sketch information up to that instant and retrieve the target photo at the earliest stroke possible (see Figure \ref{fig:Fig1} for an example of how the framework works in practice). To this end, we first pre-train a state-of-the-art FG-SBIR model \cite{yu2016sketch, song2017deep} using triplet loss. Thereafter, we keep the photo branch fixed, and fine-tune the sketch branch through a non-differentiable ranking based metric over complete sketch drawing episodes using reinforcement-learning.

Formally, a pre-trained  FG-SBIR model learns an embedding function $F(\cdot): I \rightarrow \mathbb{R}^{D}$ mapping a rasterized sketch or photo $I$ to a $D$ dimensional feature. Given a gallery of $M$ photo images $G = \left \{ X_{i} \right \}_{i=1}^{M}$, we obtain a list of $D$ dimensional vectors $\hat{G}= \left \{ F(X_{i}) \right \}_{i=1}^{M}$ using $F(.)$. Now, for a given query sketch $S$, and some pairwise distance metric, we obtain the top-$q$ retrieved photos from $G$, denoted as $Ret_{q}(F(S), \hat{G})$. If the ground truth (paired) target photo appears in the top-$q$ list, we consider top-$q$ accuracy to be true for that sketch sample. Since we are dealing with on-the-fly retrieval, a sketch is represented as $S \in \left \{p^{1}, p^{2}, p^{3}, \cdots,  p^{N}  \right \}$, where $p^{i}$ denotes one sketch coordinate tuple $(x,y)$, and $N$ stands for maximum number of points. We assume that there exists a sketch rendering operation $ \varnothing $, which takes a list $S^{K}$ of the \textit{first $K$ coordinates} in $S$, and produces one rasterized sketch image. Our objective is to train the framework so that the ground-truth paired photo appears in $Ret_{q}(F(\varnothing(S^{K})), \hat{G})$ with a minimum value of $K$.

\section{Background: Base Models} \label{basemodel}
\begin{figure}[!hbt]
\begin{center}
  \includegraphics[width=\linewidth]{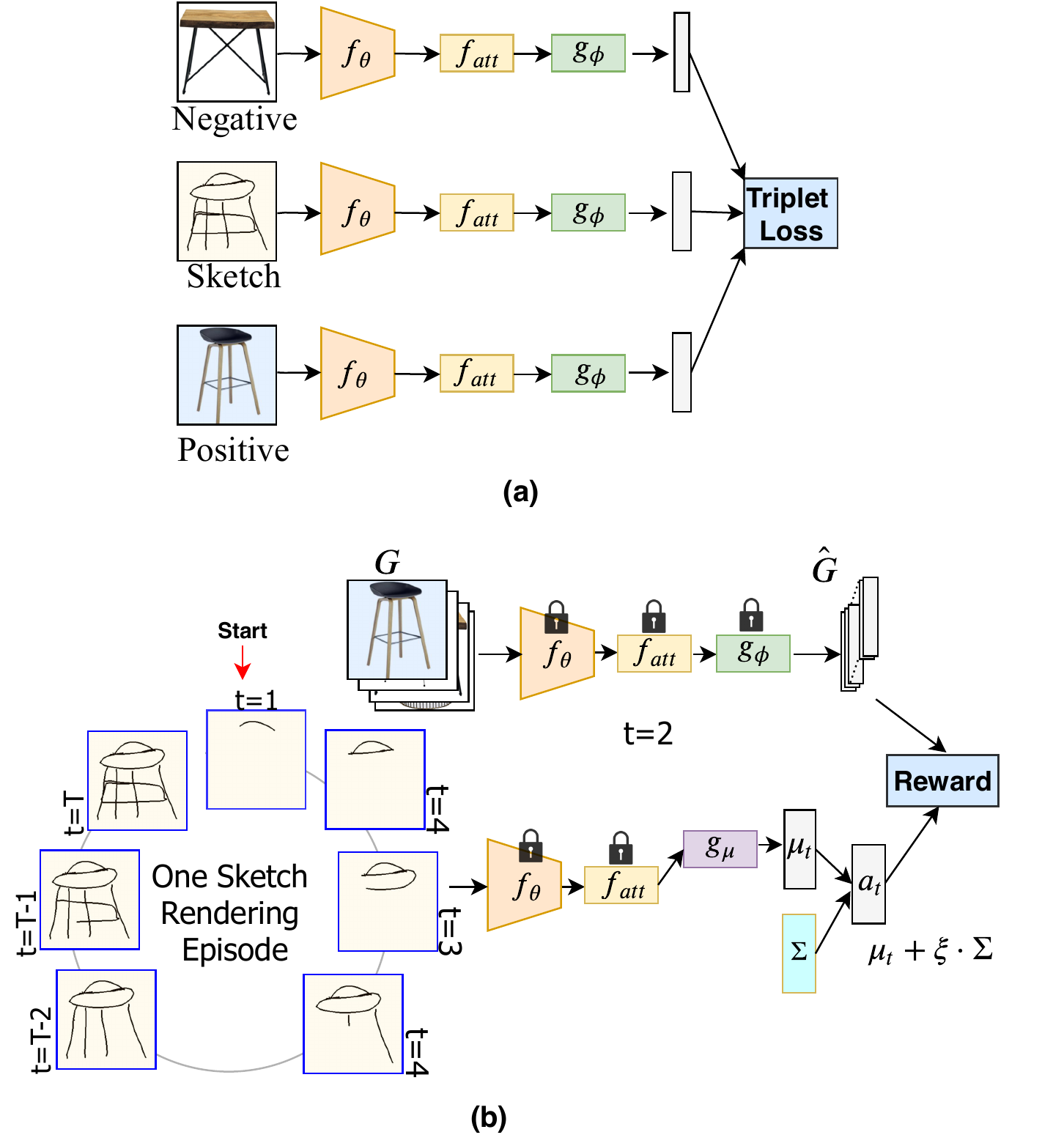} 
\end{center}
\vspace{-.2in}
  \caption{(a) A conventional FG-SBIR framework trained using triplet loss. (b) Our proposed reinforcement learning based framework that takes into account a complete sketch rendering episode. Key locks signifies particular weights are fixed during RL training.}
\label{fig:Fig123_2}
\end{figure}

For pre-training, we use a state-of-the-art Siamese network \cite{song2017deep} with three CNN branches with shared weights, corresponding to a query sketch, positive and negative photo respectively (see Figure \ref{fig:Fig123_2}(a)). Following recent state-of-the-art sketch feature extraction pipelines \cite{dey2019doodle, song2017deep}, we use soft spatial attention \cite{xu2015show} to focus on salient parts of the feature map. Our baseline model consists of three specific modules: (a) $f_{\theta}$ is initialised from pre-trained InceptionV3 \cite{szegedy2016rethinking} weights, (b) $f_{att}$ is modelled using 1x1 convolution followed by a softmax operation, (c) $g_{\phi}$ is a final fully-connected layer with $l_2$ normalisation to obtain an embedding of size D. Given a feature map $B = f_{\theta}(I)$, the output of the attention module is computed by $B_{att} = B + B\cdot f_{att}(B) $. Global average pooling is then used to get a vector representation, which is again fed into $g_{\phi}$ to get the final feature representation used for distance calculation. We considered $f_{\theta}$, $f_{att}$, and $g_{\phi}$ to be wrapped as an overall embedding function $F$. The training data are triplets $\{a, p, n\}$ containing sketch anchor, positive and negative photos respectively. The model is trained using \textit{triplet loss} \cite{weinberger2009metric_learn_margin} that aims to reduce the distance between sketch anchor and positive photo $\beta^{+} = \left \| F(a) -F(p) \right \|_{2}$, while increasing the distance between sketch anchor and negative photo $\beta^{-} = \left \| F(a) -F(n) \right \|_{2}$. Hence, the triplet loss can be formulated as $max\{0, \mu + \beta^{+} - \beta^{-}\}$, where $\mu$ is the margin hyperparameter.

\section{Methodology}
\subsection{Overview} We model an on-the-fly FG-SBIR model as a sequential decision making process \cite{kaelbling1996reinforcement}. Our agent takes \emph{actions} by producing a feature vector representation of the sketch at each rendering step, and is \emph{rewarded} by retrieving the paired photo early. Due to computation overhead, instead of rendering a new sketch at every coordinate instant, we rasterize the sketch a total of $T$ times, i.e., at steps of interval $\left \lfloor \frac{N}{T} \right \rfloor$. As the photo branch remains constant, we get $\hat{G}$ using the baseline model. We train the agent (sketch-branch) to deal with partial sketches. In this stage we fine-tune the sketch branch only, aiming to make it competent in dealing with partial sketches. 
Considering one sketch rendering episode as $S \in \{ p_{1}, p_{2}, p_{3}, ..., p_{T} \}$, the agent takes state $s_{t} = \varnothing(p_{t})$ as input at every time step $t$, producing a continuous `action' vector $a_{t}$. Based on that, the retrieval environment returns one reward $r_{t}$, mainly taking into account the pairwise distance between $a_{t}$ and $\hat{G}$. The goal of our RL model, is to find the optimal policy for the agent that maximises the total reward under a complete sketch rendering episode.

Triplet loss \cite{song2017deep,weinberger2009metric_learn_margin} considers only a single instant of a sketch. However, due to creation of multiple partially-completed instances of the same sketch, a diversity is created which confuses the triplet network. In contrast, our approach takes into account the complete episode of progressive sketch generation before updating the weights, thus providing a more principled and practically reliable way to model partial sketches.

\subsection{Model} The sketch branch acts as our agent in RL framework, based on a stochastic continuous Gaussian policy \cite{duan2016benchmarking}, where action generation is represented with a multivariate Normal distribution. Following the typical RL notation, we define our policy as $\pi_{\Theta }(a|s)$. $\Theta$ encases the parameters of policy network comprised of pre-trained $f_{\theta}$ and $f_{att}$ which remains fixed, and a fully-connected trainable layer $g_{\mu}$ that finally predicts the mean vector \textbf{$\mu$} of the multivariate Gaussian distribution. Please refer to Figure \ref{fig:Fig123_2}(b) for an illustration. At each time step $t$, a policy distribution $\pi_{\Theta }(a_{t}|s_{t})$ is constructed from the  distribution parameters predicted by our deep network. Following this, an action is sampled from this distribution, acting as a $D$ dimensional feature representation of the sketch at that instant, i.e. $a_{t} \sim \pi_{\Theta }(\cdot |s_{t})$. Mathematically, this Gaussian policy is defined as:

\vspace{-0.1cm}
 \begin{multline}\label{policy_eqn}
 \mathrm{\pi_{\Theta }(a_{t}|s_{t}) = \sqrt{\frac{1}{(2\pi)^{D} \left | \Sigma \right |}} \times}   \mathrm{exp\left \{-\frac{1}{2}(a_{t} - \mu_{t})^{\top} \Sigma^{-1} (a_{t} - \mu_{t}) \right \}}
 \end{multline}

\noindent where the mean ${\mu_{t}} = g_{\mu}(s'_{t}) \in \mathbb{R}^{D}$, and $s'_{t}$ is obtained via a pre-trained  $f_{\theta}$ and $f_{att}$ that take state $s_{t} = \varnothing(p_{t})$ as its input. Meanwhile, $\Sigma$ is a standalone trainable diagonal covariance matrix. We sample action $a_{t} = \mu_{t} +  \xi \cdot \Sigma $, where $\xi \sim \mathcal{N}(\mathbf{0},\mathbf{I})$ and $a_{t} \in \mathbb{R}^{D}.$

\subsection{Local Reward} In line with existing works leveraging RL to optimize non-differentiable task metrics in computer vision (e.g.,~\cite{rennie2017self}), our  optimisation objective is the non-differentiable \textit{ranking} metric. The distance between a query sketch embedding and the paired photo should be lower than the distance between the query and all other photos in $G$. In other words, our objective is to minimise the rank of paired photo in the obtained rank list. Following the notion of maximising the reward over time, we maximise the \textit{inverse rank}. For $T$ sketch rendering steps under a complete episode of each sketch sample, we obtain a total of $T$ scalar rewards that we intend to maximize:
\begin{equation}\label{reward_1}
R_{t}^{Local}  =  \frac{1}{rank_{t}}
\end{equation}
From a geometric perspective, assuming a high value of T, this reward design can be visualised as maximising the area under a curve, where the $x$ and $y$ axes correspond to \textit{percentage of sketch} and $\frac{1}{rank_t}$ respectively. Maximising this area therefore requires the model to achieve early retrieval of the required photo.

\subsection{Global Reward}  
During the initial steps of sketch rendering, the uncertainty associated with the sketch representation is high because an incomplete sketch could correspond to various photos (e.g., object outline with no details yet). The more it progresses towards completion, the representation becomes more concrete and moves towards one-to-one mapping with a corresponding photo. 
To model this observation, we use Kendall-Tau-distance \cite{knight1966computer} to measure the distance between two rank lists obtained from sequential sketching steps $L_t$ and $L_{t+1}$. Kendall-Tau measures the distance between two ranking lists \cite{pedronette2013image} as the number of pairwise disagreements (pairwise ranking order change) between them.
Given the expectation of more randomness associated with early ambiguous partial sketches, the Kendall-Tau-distance between two successive rank lists from the initial steps of an episode is expected to be higher.
Towards its completion, this value should decrease as the sketch becomes more unambiguous. 
With this intuition, we add a regularizer that encourages the \emph{normalised} Kendall-Tau-distance $\tau$ between two successive rank lists to be monotonically decreasing over a sketch rendering episode: 
\begin{equation}\label{reward_2}
R_{t}^{Global}= -max(0,  \tau(L_{t}, L_{t+1}) -\tau(L_{t-1}, L_{t}))
\end{equation}

This global regularisation reward term serves three purposes: (a) It models the extent of uncertainty associated with the partial sketch. (b) It discourages excessive change in the rank list later in an episode, making the retrieved result more consistent. This is important for user experience: if the returned top-ranked photos are changing constantly and drastically as the user adds more strokes, the user may be dissuaded from continuing.   (c) Instead of simply considering the rank of the target, it considers the behaviour of the full ranking list and its consistency at each rendering step.


\subsection{Training Procedure} 

We aim at maximising the sum of two proposed rewards
\begin{equation}\label{reward_3}
R_{t}  = \gamma_{1} \cdot R_{t}^{Local} + \gamma_{2} \cdot R_{t}^{Global}
\end{equation}

The RL literature provides several options for optimisation. The vanilla policy gradient \cite{mnih2016asynchronous} is the simplest, but suffers from poor data efficiency and robustness. Recent alternatives such as trust region policy optimization (TRPO) \cite{schulman2015trust} are more data efficient, but involves complex second order derivative matrices. We therefore employ Proximal Policy Optimization (PPO) \cite{schulman2017proximal}. Using only first order optimization, it is data efficient as well as simple to implement and tune. PPO tries to limit how far the policy can change in each iteration, so as to reduce the likelihood of taking wrong decisions. More specifically, in vanilla policy gradient, the current policy is used to compute the policy gradient whose objective function is given as:
\begin{equation}\label{ppo_1}
J(\Theta)= \mathrm{\hat{\mathop{\mathbb{E}}}_{t}\left [  log\pi_{\Theta}(a_{t}|s_{t}) R_{t}\right ]}
\end{equation}

\noindent PPO uses the idea of importance sampling \cite{neal2001annealed} and maintains two policy networks, where evaluation of the current policy $\pi_{\Theta}(a_{t}|s_{t})$ is done by collecting samples from the older policy $\pi_{old}(a_{t}|s_{t})$, thus helping in sampling efficiency. Hence, along-with importance sampling, the overall objective function  written as:

\begin{equation}\label{ppo_2}
J(\Theta)= \mathrm{\hat{\mathop{\mathbb{E}}}_{t}\left [  \frac{\pi_{\Theta}(a_{t}|s_{t})}{\pi_{old}(a_{t}|s_{t})} r_{t}\right ]} = \mathrm{\hat{\mathop{\mathbb{E}}}_{t}\left [ m_{t}(\Theta) R_{t}\right ]}
\end{equation}

\noindent where $m_{t}(\Theta)$ is the probability ratio $m_{t}(\Theta) = \frac{\pi_{\Theta}(a_{t}|s_{t})}{\pi_{old}(a_{t}|s_{t})}$, which measures the difference between two polices. Maximising Eqn.~\ref{ppo_2} would lead to a large policy update, hence it penalises policies moving $m_{t}(\Theta)$ away from 1, and the new clipped surrogate objective function becomes: 
\begin{equation}\label{ppo_3}
J(\Theta)= \mathrm{\hat{\mathop{\mathbb{E}}}_{t}\left [min(m_{t}(\Theta)r_{t}, clip(m_{t}(\Theta), 1 - \varepsilon, 1 + \varepsilon)R_{t}) \right ]}
\end{equation}
\noindent where $\varepsilon$ is a hyperparameter  set to 0.2 in this work. Please refer to \cite{schulman2017proximal} for more details. Empirically, we found the actor-only version of PPO with clipped surrogate objective to work well for our task. More analysis is given in Sec.~\ref{abla_124}.

\section{Experiments}
    \label{chapter:SBIL:Experiments}
\subsection{Dataset} 
We use QMUL-Shoe-V2 \cite{pang2019generalising, riaz2018learning, song2018learning} and QMUL-Chair-V2 \cite{song2018learning} datasets that have been specifically designed for FG-SBIR.
Both datasets contain coordinate-stroke information, enabling us to render the rasterized sketch images at intervals, for training our RL framework and evaluating its retrieval performance over different stages of a complete sketch drawing episode. QMUL-Shoe-V2 contains a total of 6,730 sketches and 2,000 photos, of which we use 6,051 and 1,800 respectively for training, and the rest for testing. For QMUL-Chair-V2, we split it as 1,275/725 sketches and 300/100 photos for training/testing respectively.

\subsection{Implementation Details} 
We implemented our framework in PyTorch \cite{paszke2017automatic} conducting experiments on a 11 GB Nvidia RTX 2080-Ti GPU. An Inception-V3 \cite{szegedy2016rethinking} (excluding the auxiliary branch) network pre-trained on ImageNet datasets \cite{russakovsky2015imagenet}, is used as the backbone network for both sketch and photo branches. In all experiments, we use Adam optimizer \cite{kingma2014adam} and set $D = 64$ as the dimension of final feature embedding layer. We train the base model with a triplet objective having a margin of $0.3$, for $100$ epochs with batch size $16$ and a learning rate of $0.0001$. During RL based fine-tuning of sketch branch, we train the final $g_{\mu}$ layer of sketch branch (keeping $f_{\theta}$ and $f_{att}$ fixed) with $\Sigma$ for $2000$ epochs with an initial learning rate of 0.001 till epoch 100, thereafter reducing it to 0.0001. The rasterized sketch images are rendered at $T = 20$ steps, and the gradients are updated  by averaging over a complete sketch rendering episode of $16$ different sketch samples. In addition to normalising the sampled action vector $a_{t}$, $l_{2}$ normalisation is also used after global adaptive average pooling layer as well as after the final feature embedding layer $g_{\phi}$ in image branch. The diagonal elements of $\Sigma$ are initialised with $1$, and $\gamma_{1} $,  $\gamma_{2}$  and $\varepsilon$ are set to 1, $\mathrm{1e-4}$  and 0.2 respectively.

\subsection{Evaluation Protocol} 
In line with on-the-fly FG-SBIR setting, we consider results appearing at the top of the list to matter more. Thus, we quantify the performance using Acc.@$q$ accuracy, i.e. percentage of sketches having true-match photos appearing in the top-$q$ list. Moreover, in order to capture the early retrieval performance, shadowing some earlier image retrieval works \cite{kovashka2012whittlesearch}, we use plots of (i) \textit{ranking percentile} and (ii) \textit{$\frac{1}{rank}$} versus \textit{percentage of sketch}. In this context, a higher value of the mean area under the curve for (i) and (ii) signifies better early sketch retrieval performance, and we use m@A and m@B as shorthand notation for them in the rest of the work, respectively.

\subsection{Competitor} 

To the best of our knowledge, there has been no prior work dealing with early retrieval in SBIR. Thus, based on some earlier works, we chose existing FG-SBIR baselines and their adaptations towards the new task to verify the contribution of our proposed RL based solution.

\textbullet \ \textbf{B1}: Here, we use the baseline model \cite{song2017deep, yu2016sketch} trained only with triplet loss. This basically represents our model (see Section~\ref{basemodel}) before RL based fine-tuning.\ \textbullet \  \textbf{B2}: We train a standard triplet model, but use all intermediate sketches as training data, so that the model also learns to retrieve incomplete sketches.\ \textbullet \   \textbf{B3}: We train $20$ different models (as $T=20$) for the sketch branch, and each model is trained to deal with a specific percentage of sketch (like 5\%, 10\%, ..., 100\%), thus increasing the number of trainable parameters $20$ times than the usual baseline. Different models are deployed at different stages of completion -- again not required by any other compared methods.\ \textbullet \  \textbf{B4}: While RL is one of the possible ways of dealing with non-differentiable metrics, the recent work of Engilberge \etal \cite{engilberge2019sodeep} introduced a generalized deep network based solution to approximate non-differentiable objective functions such as ranking. This can be utilized in a plug-and-play manner within existing deep architectures. We follow a similar setup of cross-modal retrieval as designed by Engilberge \etal \cite{engilberge2019sodeep} and impose combination of triplet loss and ranking loss at T different instants of the sketch.  

\begin{figure}[!hbt]
	\begin{center}
		\includegraphics[width=1\linewidth]{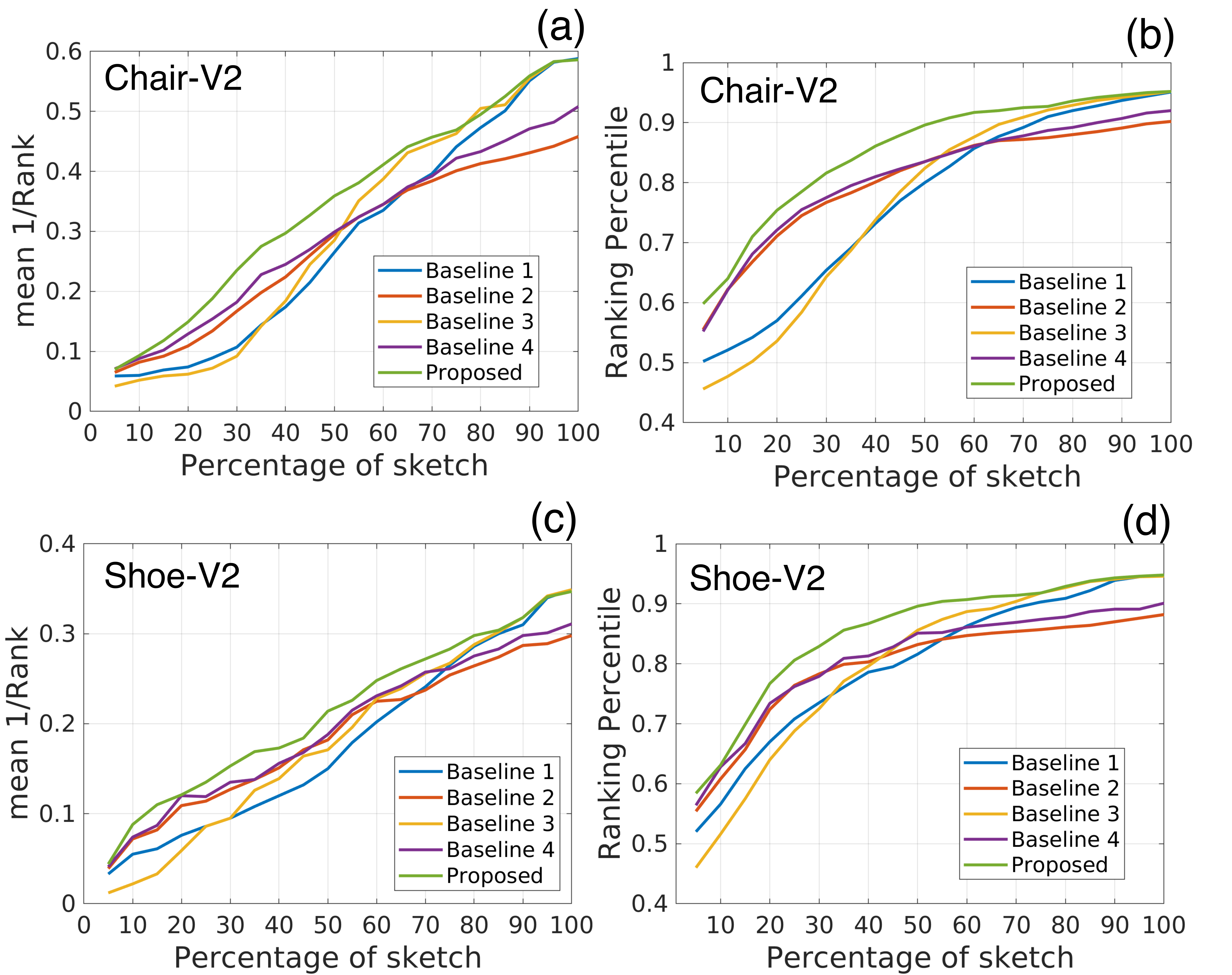}
	\end{center}
	\vspace{-0.5cm}
	\caption{Comparative results. Note that instead of showing T=20 sketch rendering steps, we visualize it through percentage of sketch here. A higher area under these plots indicates better \emph{early retrieval performance}.}
	\label{fig:Graph1_Full123}
	\vspace{-0.2cm}
\end{figure}

\vspace{-0.2cm}
\subsection{Performance Analysis}

The performance of our proposed on-the-fly sketch based image retrieval is shown in Figure~\ref{fig:Graph1_Full123} against the baselines methods. We observe: 
(i) State-of-the-art triplet loss based baseline \textbf{B1} performs poorly for early incomplete sketches, due to the absence of any mechanism for learning incomplete retrieval. 
(ii) \textbf{B2}'s imposition of triplet-loss at every sketch-rendering step provides improved retrieval performance over \textbf{B1} for a few initial instants, but its performance declines towards the completion of sketch. This is mainly due to the fact that imposing triplet loss over incomplete sketches destabilises the learning process as it generates noisy gradients.
In contrast, our RL based pipeline takes into account a complete sketch rendering episode along-with the associated uncertainty of early incomplete sketches before updating the gradients. 
(iii) Designing 20 different sketch models in \textbf{B3} for $T=20$ different sketch rendering steps, improves performance towards the end of sketch rendering episode after 40\% of sketch rendering in comparison to \textbf{B1}. However, it is poor for sketches before that stage due to its incompleteness which could correspond to various possible photos. 
(iv) An alternative of RL method -- differential sorter described in \textbf{B4}, fares well against baseline \textbf{B1}, but is much weaker in comparison to our RL based non-differentiable ranking method. A qualitative result can be seen in Figure \ref{fig:Fig1} where \textbf{B1} is the baseline.

In addition to the four baselines,  following the recent direction of dealing with partial sketches  \cite{ghosh2019interactive, liu2019sketchgan}, we tried a two stage framework for our early retrieval objective referred as \textbf{TS} in Table \ref{tab:my-table1}. At any given drawing step, a conditional image-to-image translation model \cite{isola2017image} is used to generate the complete sketch. Thereafter, it is fed to an off-the-shelf baseline model for photo retrieval. However, this choice of using an image translation network to complete the sketch from early instances, fails miserably. Moreover, it merely produces the input sketch with a few new random noisy strokes.

To summarise, our RL framework outperforms a number of baselines by a significant margin in context of early image retrieval performance as seen from the quantitative results in Figure~\ref{fig:Graph1_Full123} and Table \ref{tab:my-table1}, without deteriorating top-5 and top-10 accuracies in retrieval performance.

\subsection{Ablation Study}\label{abla_124}
\keypoint{Different RL Methods:} We compare Proximal Policy Optimization (PPO), used here for continuous-action reinforcement learning, with some alternative RL algorithms. Although we intended to use a complete actor-critic version of PPO combining the policy surrogate Eqn.~\ref{ppo_3} and value function error \cite{schulman2017proximal} term, using the actor-only version works better in our case. Additionally we evaluate this performance with (i) vanilla policy gradient \cite{mnih2016asynchronous}\cut{ which apparently suffers from poor data efficiency,} and (ii) TRPO \cite{schulman2015trust}\cut{which requires evaluating a complicated second order derivative matrix}. Empirically we observe a higher performance with clipped surrogate objective when compared with the adaptive KL penalty (with adaptive co-efficient 0.01) approach of PPO.  Table~\ref{tab:my-table2} shows that our actor-only PPO with clipped surrogate objective outperforms other alternatives.

\keypoint{Reward Analysis:} In contrast to the complex design of efficient optimization approaches for non-differentiable rank based loss functions \cite{mohapatra2018efficient, engilberge2019sodeep}, we introduce a simple reinforcement learning based pipeline that can optimise a CNN to achieve any non-differentiable metric in a cross modal retrieval task. 
To justify our contribution and understand the variance in retrieval performance with different plausible reward designs, we conduct a thorough ablative study. 
The positive scalar reward value is assigned to $1$ (otherwise zero), when the paired photo appears in top-$q$ list. This $q$ value could be controlled based on the requirements. 
Instead of reciprocating the rank value, taking its negative is also a choice. To address the concern that our inverse rank could produce too small a number, we alternatively evaluate the square root of  reciprocal rank. From the results in Table~\ref{tab:my-table3}, we can see that our designed reward function (Eqn.~\ref{reward_3}) achieves the best performance.

 \begin{figure}[t]
\begin{center}
  \includegraphics[width=0.8\linewidth]{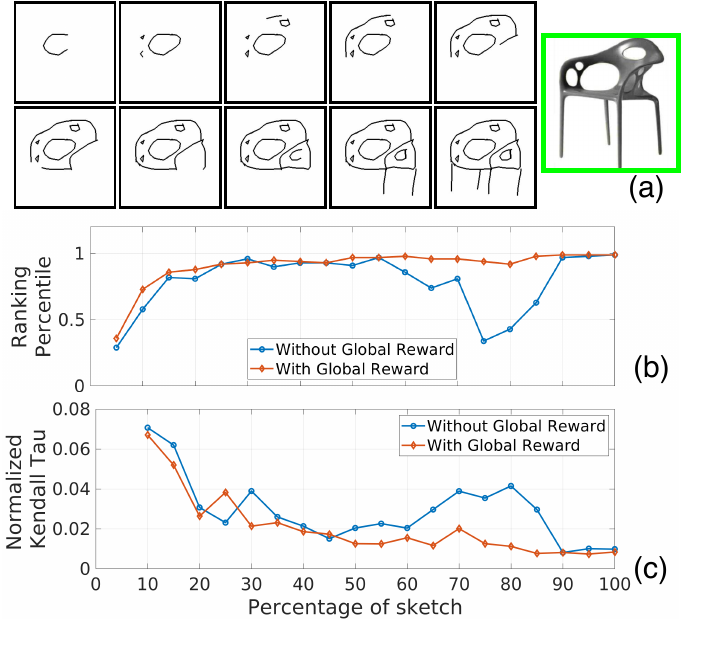}
\end{center}
\vspace{-.4 in}
  \caption{
  (a) Example showing progressive order of completing a sketch.
  (b) shows the drop in percentile whenever an irrelevant stroke is introduced while drawing (blue).
  (c) shows the corresponding explosive increase of Kendall-Tau distance signifying the percentile drop (blue). Our global reward term (red) nullifies these negative impacts of irrelevant sketch strokes thus maintaining consistency of the rank-list overall.
  }
\label{fig:Fig3}
\end{figure}

\begin{table}[!hbt]
\normalsize
\centering
\caption{Comparative results with different baseline methods. Here A@5 and A@10 denotes top-5 and top-10 retrieval accuracy for complete sketch (at t=T), respectively, whereas, m@A and m@B quantify the retrieval performance over a sketch rendering episode (Metric definition is previously mentioned).}
\vspace{-.2cm} 
\label{tab:my-table1}
\scriptsize
\begin{tabular}{ccccccccc}
\hline
 & \multicolumn{4}{c}{Chair-V2} & \multicolumn{4}{c}{Shoe-V2} \\ \cline{2-9} 
 & m@A & m@B & A@5 & A@10 & m@A & m@B & A@5 & A@10 \\ \hline
B1 & 77.18 &  29.04 & \textbf{76.47}  & 88.13 & 80.12 & 18.05  & 65.69 & \textbf{79.69}  \\ 
B2 & 80.46 &  28.07 &  74.31 &  86.69 &79.72  & 18.75 &  61.79 & 76.64  \\
B3 & 76.99 &  30.27 &  \textbf{76.47} & 88.13 & 80.13 &  18.46 &  65.69 & 79.69   \\ 
B4 & 81.24 &  29.85 &  75.14 &  87.69 & 81.02 & 19.50 & 62.34 &  77.24 \\  
TS & 76.01 &  27.64 & 73.47  & 85.13 & 77.12 & 17.13  & 62.67 & 76.47  \\ 
Ours & \textbf{85.44} &  \textbf{35.09} & 76.34 & \textbf{89.65} & \textbf{85.38} & \textbf{21.44} &   \textbf{65.77}&  79.63  \\ \hline
\end{tabular}%
\end{table}

\begin{table}[!hbt]
\scriptsize
\centering
\caption{Results with different Reinforcement Learning (RL) methods, where A stands for actor-only version of the algorithm, and AC denotes the complete actor-critic design.}
\label{tab:my-table2}
\vspace{-.2cm} 
\begin{tabular}{ccccc}
\hline
\multirow{2}{*}{RL Methods} & \multicolumn{2}{c}{Chair-V2} & \multicolumn{2}{c}{Shoe-V2} \\ \cline{2-5} 
 & m@A & m@B & m@A & m@B \\ \hline
Vanilla Policy Gradient & 80.36 & 32.34  & 82.56 & 19.67 \\ 
PPO-AC-Clipping & 81.54  & 33.71 & 83.47 & 20.84 \\ 
PPO-AC-KL Penalty & 80.99  & 32.64 & 83.84 & 20.04 \\ 
PPO-A-KL Penalty & 81.34  & 33.01 & 83.51 & 20.66 \\ 
TRPO &83.21& 33.68 & 83.61 & 20.31   \\ 
PPO-A-Clipping (Ours) & {\bf 85.44} & {\bf 35.09} & {\bf 85.38} & {\bf 21.44}   \\ \hline
\end{tabular}%
\end{table}

\begin{table}[!hbt]
\footnotesize
\centering
\caption{Results with different candidate reward designs}
\vspace{-.23cm} 
 \label{tab:my-table3}
\begin{tabular}{ccccc}
\hline
\multirow{2}{*}{Reward Schemes} & \multicolumn{2}{c}{Chair-V2} & \multicolumn{2}{c}{Shoe-V2} \\ \cline{2-5} 
 & m@A & m@B & m@A & m@B \\ \hline
$\mathrm{rank \leq 1\Rightarrow reward = 1}$ & 82.99  & 32.46 & 82.24 & 19.87 \\ 
$\mathrm{rank \leq 5\Rightarrow reward = 1}$ & 81.36  & 31.94 & 81.74 & 19.37 \\
$\mathrm{rank \leq 10\Rightarrow reward = 1}$ & 80.64 & 30.57 &  80.87 &  19.08  \\ 
$\mathrm{-rank}$  &83.71 & 32.84 & 83.81& 20.71 \\
$\mathrm{\frac{1}{\sqrt{rank}}}$  & 83.71 & 33.97 & 83.67 & 20.49 \\ 
$\mathrm{\frac{1}{rank}}$  & 84.33 & 34.11 & 84.07 & 20.54 \\ 
$\mathrm{Ours}$ (Eqn.~\ref{reward_3})   & {\bf 85.44} & {\bf 35.09} & {\bf 85.38} & {\bf 21.44}  \\ \hline
\end{tabular}%

\end{table}

\begin{table}[!hbt]
\footnotesize
\centering
\caption{Performance on varying feature-embedding spaces}
\vspace{-.23cm} 
\label{tab:my-table4}
\begin{tabular}{ c c c c c c c }
\hline
\multirow{2}{*}{} & \multicolumn{3}{c}{Chair-V2} & \multicolumn{3}{c}{Shoe-V2} \\ \cline{2-7} 
                  & m@A      & m@B      & A@5     & m@A      & m@B     & A@5     \\  \hline
32                &   82.61       & 34.67         &       72.67  &     82.94     &     19.61    &      62.31   \\ 
64                & {\bf  85.44}      &   {\bf 35.09 }     &    76.34     &    {\bf 85.38}       &   {\bf 21.44}      &     65.77     \\  
128               &    84.71      &     34.49     &     {\bf 78.61}       &  84.61     &      20.81   &       {\bf 67.64}  \\ 
256               &     81.39     &     31.37     &     77.41      &    80.69    &      19.68   &        66.49 \\ \hline
\end{tabular}%
\end{table}

\keypoint{Significance of Global Reward:} While using  our local reward (Eqn.~\ref{reward_1}) achieves an excellent rank in early rendering steps, we noticed that the rank of a paired photo might worsen at a certain sketch-rendering step later on, as illustrated in Figure~\ref{fig:Fig3}. As the user attempts to convey more fine-grained detail later on in the process, they may draw some noisy, irrelevant, or outlier stroke that degrades the
performance. Our global-reward term in Eqn.~\ref{reward_3} alleviates this issue by imposing a monotonically decreasing constraint on the \emph{normalised} Kendall-Tau-distance \cite{knight1966computer} between two successive rank lists over an episode
(Figure~\ref{fig:Fig3}). We quantify the identified adverse impacts of inconsistent strokes, via a new metric, termed \textbf{stroke-backlash index}. It is formulated as $\frac{\sum_{t = 2}^{T}\left | min(RP_{t} - RP_{t-1}  , 0)\right |}{T-1}$, where $RP_{t}$ denotes the ranking percentile of paired photo at $t^{th}$ sketch-rendering step which is averaged over all sketch samples in test split. Whenever a newly introduced stroke produces a decline in the ranking percentile, it is considered as a negative performance. Please note that \emph{the lower the value of this index, the better will be the ranking list consistency performance}. We get a decline in stroke-backlash index from $0.0421$ ($0.0451$) to $0.0304$ ($0.0337$) in Chair-V2 (Shoe-V2) dataset when including the global reward. Furthermore, as shown in Table \ref{tab:my-table3}, this global reward term improves the early retrieval performance m@A and (m@B) by 1.11\% (0.98\%) and 1.31\% (0.90\%) for Chair-V2 and Shoe-V2 respectively.
{Instead of imposing the monotonically decreasing constraint over Kendall-Tau-distance that actually considers the relative ranking position of all the photos, we could have imposed the same monotonically decreasing constraint on the specific ranking of the paired-photo only. However, we notice that the stroke-backlash index arises to $0.0416$ ($0.0446$) and overall m@A value decreases by 0.78\% (0.86\%) for Chair-V2 (Shoe-V2), thus justifying the importance of using Kendall-Tau distance in our context.} 

\keypoint{Further Analysis:} (i) We evaluate the performance of our framework with a varying embedding space dimension in Table \ref{tab:my-table4}, confirming our choice of $D=64$. (ii) Instead of using a standalone-trainable diagonal covariance matrix $\Sigma$ for the actor network, we tried employing a separate fully-connected layer to predict the elements of $\Sigma$. However, the m@A and m@B performance deteriorates by 5.64\% (4.67\%) and 4.48\% (3.51\%), for Chair-V2 and (Shoe-V2) datasets respectively. (iii) In the context of on-the-fly FG-SBIR where we possess online sketch-stroke information, a reasonable alternative could be using a recurrent neural network like \cite{collomosse2019livesketch} for modelling the sketch branch instead of CNN. Following SketchRNN's \cite{ha2017neural}  vector representation, a five-element vector is fed at every LSTM unit, and the hidden state vector is passed through a fully connected layer at any arbitrary instant, thus predicting the sketch feature representation. This alleviates the need of feeding a rendered rasterized sketch-image every time. However, replacing the CNN-sketch branch by RNN and keeping rest of the setup unchanged, performance drops significantly.  As a result a top@5 accuracy of 19.62\% (15.34\%) is achieved compared to 76.34\% (65.77\%) in case of CNN for Chair-V2 (Shoe-V2) dataset. (iv) Different people have different stroke orders for sketching. 
Keeping this in mind we conducted an experiment by randomly shuffling stroke orders to check the consistency of our model.
We obtain m@A and m@B values of 85.04\% (34.84\%) and 85.11\% (20.92\%) for Chair-V2 (Shoe-V2) datasets, respectively,  demonstrating our robustness to such variations.

\section{Conclusion}
\label{chapter:OTF:Conclusion}
We have introduced a fine-grained sketch-based image retrieval framework designed to mitigate the practical barriers to FG-SBIR by analysing user sketches on-the-fly, and retrieve photos at the earliest instant. To this end, we have proposed a reinforcement-learning based pipeline with a set of novel rewards carefully designed to encode the `early retrieval' scheme and stabilise the learning procedure against inconsistently drawn strokes. This provides considerable improvement on conventional baselines for on-the-fly FG-SBIR.


{On-the-fly retrieval system starts retrieving as soon as the user starts drawing, and therefore the users do not need to draw a complete sketch to get faithful retrieval results. Thus it makes the retrieval process much faster and provides more flexibility to users who are less confident with sketching. 
Overall, this on-the-fly retrieval framework guided fine-grained SBIR literature towards better practicality of using the sketch as a query medium for fine-grained image retrieval. 
Talking of \aneeshan{practical deployment} however, an important aspect remains to be addressed -- the time-consuming and labour-intensive process of collecting instance-level photo-sketch pairs. Therefore, in our next chapter, we introduce a semi-supervised framework for fine-grained sketch retrieval to deal with the data-scarcity issue that additionally uses unlabelled photos to improve retrieval performance.}
\chapter{Semi-Supervised Learning for Fine-Grained SBIR}
    \label{chapter:SSL_FGSBIR}
    \minitoc
    \newpage
    
    \section{Overview}
    \label{chapter:SSL_FGSBIB:Overview}

\aneeshan{Looking further towards the practical application of Fine-grained SBIR, one would agree that there will hardly be sufficiently large sketch-photo pairs to train a good model.} 
{We first} test the hypothesis that -- freely-available \textit{unlabelled photo data} would help to mitigate the performance gap imposed by the lack of specifically collected \textit{photo-sketch paired data}. Our utmost contribution {in this chapter} is therefore a semi-supervised FG-SBIR framework where unlabelled photo data (i.e., photos without matching sketches) are used alongside photo-sketch pairs for model training. We differ significantly to conventional semi-supervised classification methods \cite{sohn2020fixmatch, pham2020meta} -- other than learning pseudo photo labels via a learnable classifier, our ``label'' for a photo is in the form of a visual sketch which needs to be \textit{generated} rather than \textit{classified}. Thus, at the core of our design is a sequential photo-to-sketch \textit{generation model} that outputs pseudo sketches for unlabelled photos. The hope is therefore that such pseudo sketch-photo pairs could augment the training of a \textit{retrieval model}. 
 
 \begin{figure}[!hbt]
\begin{center}
  \includegraphics[width=0.8\linewidth]{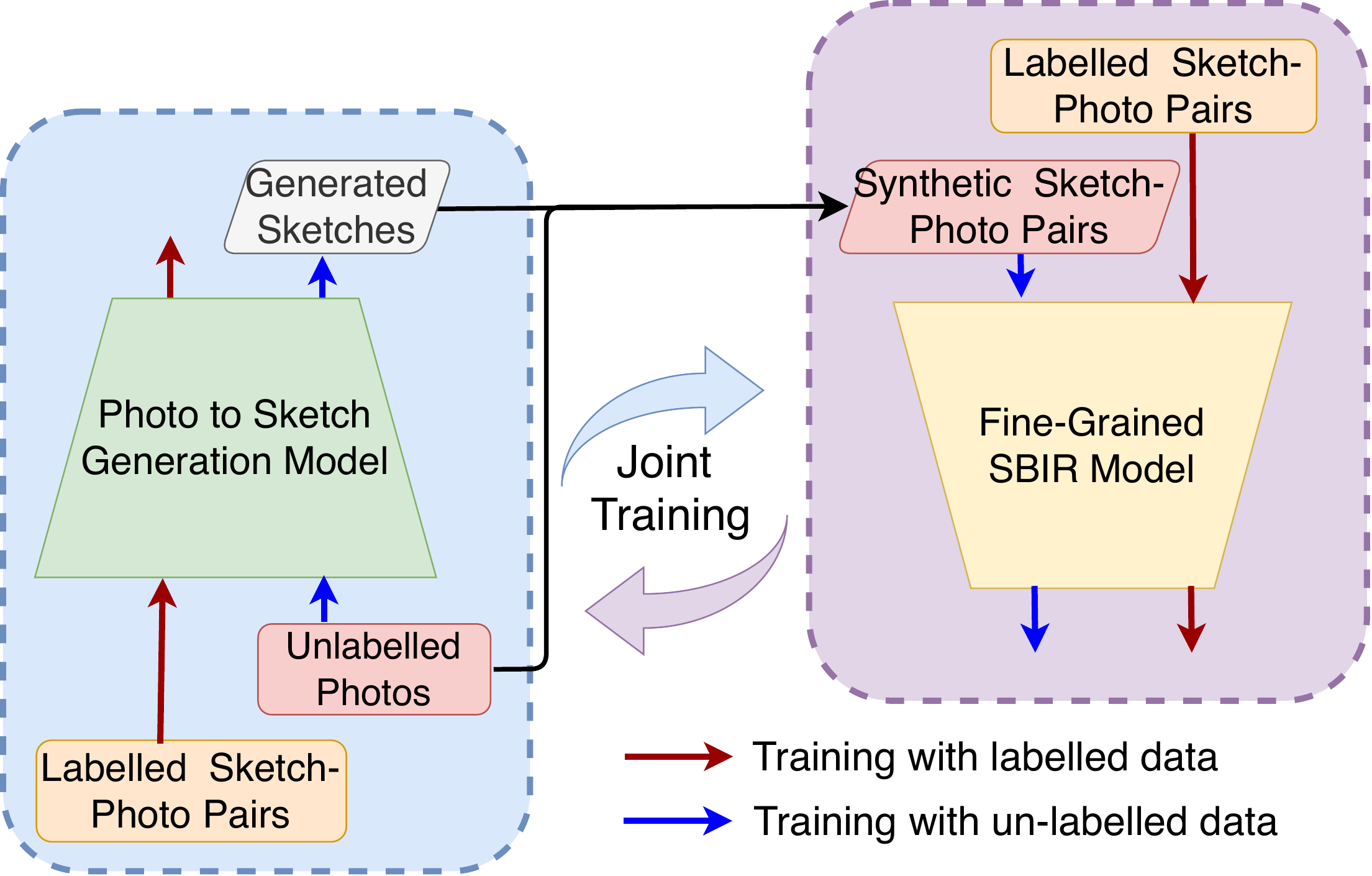}
\end{center}
\vspace{-0.6cm}
  \caption{Our proposed method additionally leverages large scale photos without any manually labelled paired sketches to improve FG-SBIR performance. Moreover, we show that the two conjugate process, \emph{photo-to-sketch} generation and \emph{fine-grained SBIR}, could improve each other by joint training.}
\label{fig:Fig1_a.pdf}
\vspace{-0.3cm}
\end{figure}

Naively cascading a generator with a retrieval model however would not work. This is mainly because off-the-shelf photo-to-sketch generation models \cite{song2018learning, chen2017sketch} could sometimes generate unfaithful sketches that may not resemble their corresponding photos, especially when it comes to fine-grained visual features. The downstream retrieval model would then have no way of knowing which pseudo sketch and photo pairs are worth training with, ultimately resulting in performance degradation. This leads to an important design consideration of ours -- we advocate that there is positive complementarity between generation and retrieval that can be explored via \emph{joint learning} (Figure \ref{fig:Fig1_a.pdf}). The intuition is simple -- pseudo sketches automatically generated from unlabelled photos can help to semi-supervise a better retrieval model, \textit{and vice versa} that better retrieval model can feed back to the generator in producing more faithful sketch-photo pairs.

The key therefore lies with \textit{how} such positive exchange cycle can be facilitated between the generator and retrieval model. 
To this end, novelty lies in the components introduced in both generator and retrieval model designs, and in how\cut{ two models} they are jointly trained.
\emph{First}, we formulate a novel sequential photo-to-sketch generator with spatial resolution preservation and a cross-modal 2D-attention mechanism. \emph{Second}, a discriminator is formulated in the retrieval model, to quantify the reliability of generated pseudo photo-sketch pairs. Reliability scores are then used\cut{to perform} for instance-wise weighting of triplet-loss values upon updating the retrieval model. A consistency loss (via distillation) is further introduced to simultaneously suppress the noisy training signal coming from pseudo photo-sketch pairs. \emph{Third}, to enable exchange from retrieval to generation, we rely on the following intuition -- good synthetic pairs would trigger a low value on the resulting triplet loss and a higher output of the discriminator. Feeding these training signals back to the generator would however involve passing through a non-differentiable rasterization operation (Figure \ref{fig:Fig12352}). We thus employ a policy-gradient \cite{sutton2000policy} based reinforcement learning scheme that feeds back these signals as \textit{rewards}.

In summary, our contributions are: (a) For the first time, we propose to solve the data scarcity problem in FG-SBIR by adopting   \textit{semi-supervised} approach that additionally leverages large scale unlabelled photos to improve retrieval accuracy. (b) To this end, we couple sequential sketch generation process with fine-grained SBIR model in a joint learning framework based on reinforcement learning. (c) We further propose a novel photo-to-sketch generator and introduce a discriminator guided \emph{instance weighting} along with \emph{consistency loss}  to retrieval model training with  noisy synthetic photo-sketch pairs. (d) Extensive experiments validate the efficacy of our approach for overcoming data scarcity in FG-SBIR (Figure \ref{fig:ablation3}) -- we can already reach performances at par with prior arts with just a fraction ($\approx$60$\%$) of the training pairs, and obtain state-of-the-art performances on both QMUL-Shoe and QMUL-Chair with the same training data (by $\approx$6$\%$ and $\approx$7$\%$  respectively).

\section{Problem Formulation}

For semi-supervised fine-grained SBIR, we consider having access to a limited amount of labelled photo-sketch pairs $\mathrm{\mathcal{D}_{L} = \{ ({p_{L}^{i}}; {s_{p,L}^{i}})\}_{i}^{N_{L}}}$ and a much bigger set of unlabelled photos  $\mathrm{\mathcal{D}_{U} = \{ {p_{U}^{i}} \}_{i}^{N_{U}}}$, where $N_{U} \gg N_L$. The key objective is to improve retrieval performance using both $\mathcal{D}_{L}$ and unlabelled photos $\mathcal{D}_{U}$ (having no corresponding sketches). More specifically, our framework consists of two models learned jointly: a FG-SBIR model, and a photo-to-sketch generation model. The retrieval model tries to learn an embedding function $ \mathcal{F(\cdot)}: \mathbb{R}^{H \times W \times 3} \rightarrow \mathbb{R}^{d}$ mapping any rasterized sketch or photo having height $H$ and width $W$ to a $d$-dimensional feature.

Instead of image-to-image translation \cite{isola2017image, li2019photo}, sketch generation process needs to be designed by sequential sketch-coordinate decoding \cite{ha2017neural} in order to model the hierarchical abstract nature of sketch. In particular, the FG-SBIR model requires rasterized sketch-images to obtain the sketch embedding, as  performance can collapse on using sketch-coordinate instead \cite{bhunia2020sketch, sain2020cross}. Thus, the generator learns a function $\mathcal{G}(\cdot): \mathbb{R}^{H \times W \times 3} \rightarrow \mathbb{R}^{T\times2}$, mapping a photo to equivalent sequential sketch-coordinate points $\mathrm{s_c = \{(x_1, y_1),(x_2, y_2), \cdots, (x_T, y_T) \}}$, where $T$ is the number of points.
Note that in order to feed the generated sketches to the feature embedding network $ \mathcal{F}$ of the retrieval model,  a  \emph{rasterization} (sketch-image redrawing from coordinates) operation is required which is denoted as  $s_p = \phi(s_c) : \mathbb{R}^{T\times2} \rightarrow \mathbb{R}^{H \times W \times 3}$. Finally, we can create synthesised photo-sketch pairs  $\mathrm{\mathcal{D}_{U}' = \{ ({p_{U}^{i}}; {s_{p,U}^{i}})\}_{i}^{N_{U}}}$ for unlabelled photos to train the retrieval model. Once trained, only $\mathcal{F(\cdot)}$ is used for retrieval during inference,  while $\mathcal{G(\cdot)}$ augments pseudo/synthetic photo-sketch pairs for training.

\section{Photo to Sketch Generation Model} \label{p2s}
The existing sequential photo-to-sketch generation models \cite{song2018learning, chen2017sketch} comprise a convolutional image encoder, followed by an LSTM decoder. This however has two major limitations: Firstly, it reduces the photo representation to a latent vector, leading to significant spatial information loss. Secondly, one fixed global representation is given as input at every time step of the LSTM decoding. To overcome these limitations two novel designs are introduced: (a) keeping the spatial feature map while ignoring global average pooling; (b) looking back at the specific part of photo which it \emph{draws}. Overall, it consists of three major components, a CNN encoding the photo, a 2-D attention module, and an LSTM decoder generating the coordinates sequentially.  

Given a photo $p$, let the extracted convolutional feature map be $\mathcal{B} \in \mathbb{R}^{h'\times  w' \times c}$ where $h'$, $w'$ and $c$ denotes the height, width and number of channels, respectively. Next, we perform a global average pooling on $\mathcal{B}$ to obtain a vector of size $\mathbb{R}^c$, and project it as two vectors $\mu$ and $\sigma$, each having size $\mathbb{R}^{N_z}$. The global embedding of photo is obtained through a reparameterization trick as $z = \mu + \sigma \odot \mathcal{N}(0,1)$. The initial hidden state $h_{0}$ (and optional cell state $c_{0}$) of decoder RNN is initialised as $[h_0; c_0] = \tanh(W_{z}z + b_z)$. 

Instead of predicting the absolute coordinates $\{(x_i, y_i)\}_{i}^{T}$, we model every point as 5 element vectors $(\Delta x, \Delta y, p_1, p_2, p_3)$  where $\Delta x$ and $\Delta y$ represents the off-set distances \cite{ha2017neural} in the $x$ and $y$ directions from the previous point. The last three elements represent a binary one-hot vector of three pen-state situations: pen touching the paper, pen being lifted and end of drawing. Each offset-position ($\Delta x, \Delta y$) is modelled  using a  Gaussian mixture model (GMM) with $M=20$ bivariate normal distributions \cite{ha2017neural}  given by:
\vspace{-0.25cm}
\begin{equation}
p(\Delta x, \Delta y) = \sum_{j=1}^{M}\Pi_j \mathcal{N}(\Delta x, \Delta y\mid  \lambda_{j});  \; \sum_{j=1}^{M}\Pi_j= 1.
\end{equation}
\vspace{-0.4cm}

\noindent Here, we consider each M bivariate normal distribution to have \emph{five} parameters $\lambda = \{ \mu_{x}, \mu_{y}, \sigma_{x}, \sigma_{x}, \rho_{xy}\}$ with mean ($\mu_x, \mu_y$), standard deviation ($\sigma_x, \sigma_y$) and correlation ($\rho_{xy}$). The mixture weights of the GMM is modelled by a categorical distribution of size $\mathbb{R}^M$. Thus every time step's output $y_t$ modelled is of size  $\mathbb{R}^{5M + M + 3}$, which includes 3 logits for pen-state.  At time step $t$, a recurrent decoder network updates its state $s_t = (h_{t}, c_{t})$ as follows: $\mathrm{ s_t = RNN(s_{t-1} ; [g_{t}, P_{t-1}] )}$ where $g_{t}$ is the glimpse vector  encoding the information from specific relevant parts of the feature map $\mathcal{B}$ to predict $y_{t}$; $P_{t-1}$ is the last predicted point (start-token $P_0 =\{0, 0, 1, 0, 0\}$),  $[\cdot]$ signifies a concatenation operation. The glimpse/context vector is obtained by 2D attention as follows:

\vspace{-0.5cm}
\begin{align}
    \begin{cases}
      J =  \tanh(W_{\mathcal{B}} \circledast  \mathcal{B} + W_{S}h_{t-1});  \\ 
     \alpha_{i,j}  =  \mathrm{softmax}(W_{a}^T  J_{i,j})   \\
      g_{t} =  \sum_{i,j} \alpha_{i,j} \cdot \mathcal{B}_{i,j};  \; i=[1,..h'], \; j=[1,..w']
    \end{cases}
\end{align}
\vspace{-0.3cm}

\noindent where $W_{B}$, $W_{S}$, $W_{a}$ are the learnable weights. Calculating the attention weight $\alpha_{i,j}$ at every spatial position $(i,j)$, we employ a convolution operation ``$\circledast $" with $3\times3$ kernel $W_{\mathcal{B}}$ to consider the neighbourhood information in the 2D attention module, and $g_t$ is obtained by weighted summation operation at the end. A fully-connected layer over every hidden state outputs $y_t = W_yh_t + b_y$ , where $y_t \in \mathbb{R}^{6M+3}$. We refer to \cite{ha2017neural} for more details. Like the standard VAE, our generator $\mathcal{G}$ is trained from the weighted summation of a reconstruction loss ($\mathrm{L^R_{\mathcal{G}}}$) and a KL-divergence loss ($\mathrm{L^{kl}_{\mathcal{G}}}$)  with unit normal distribution as follows: 

\vspace{-0.5cm}
\begin{equation}\label{vae_loss}
\mathrm{L_{\mathcal{G}}^{vae} = L^{R}_{\mathcal{G}} + \omega_{kl} L^{kl}_{\mathcal{G}}},    
\end{equation}
\vspace{-0.5cm}

\noindent where  $L^{R}_{\mathcal{G}}$ is composed of the negative log-likelihood loss of the offsets $\Delta z = (\Delta x, \Delta y)$ and the pen states $(p_1, p_2, p_3)$: $\mathrm{L^{R}_{\mathcal{G}}} = - \frac{1}{T} \Big[ \sum_{i=1}^{T}\log p(\Delta z_i\mid \lambda_i) + \hat{p}_i\log(p_i) \Big]$.

\section{Baseline FG-SBIR Model} \label{baseline_fgbsir}
 
For the discriminative retrieval module $\mathcal{F(\cdot)}$, we use the state-of-the-art Siamese network \cite{song2017deep, dey2019doodle, bhunia2020sketch} (multi-branch with weight-sharing)  with soft spatial attention \cite{xu2015show} to focus on salient parts of the feature map. 
Concretely, given a photo or rasterized sketch image $I$, we use a pre-trained InceptionV3 model \cite{szegedy2016rethinking} to extract feature map $F' = f_{\theta}(I)$. This is followed by a residual connection between backbone feature and attention normalised feature to give $F = F' + F'\cdot f_{attn}(F')$, upon which global average pooling is performed to obtain final feature representation of size $\mathbb{R}^d$; and $f_{attn}$ is  modelled using 1x1 convolution with softmax across the spatial dimensions. For training, the distance to a sketch anchor (a) from a negative photo (n), denoted as $\beta^{-} = \left \| \mathcal{F}(a) - \mathcal{F}(n) \right \|_{2}$ should increase while that from the positive photo (p), $\beta^{+} = \left \| \mathcal{F}(a) - \mathcal{F}(p) \right \|_{2}$ should decrease. This is brought about by the triplet loss with a margin $\mu > 0$ as a hyperparameter:

\begin{equation}\label{triplet}
\mathrm{L_{\mathcal{F}}^{trip} = max\{0, \mu + \beta^{+} - \beta^{-}\}}.
\vspace{-0.1cm}
\end{equation}

\section{Semi-Supervised Framework for FG-SBIR}
\subsection{Overview} 
\begin{sidewaysfigure}
\begin{center}
  \includegraphics[width=\linewidth]{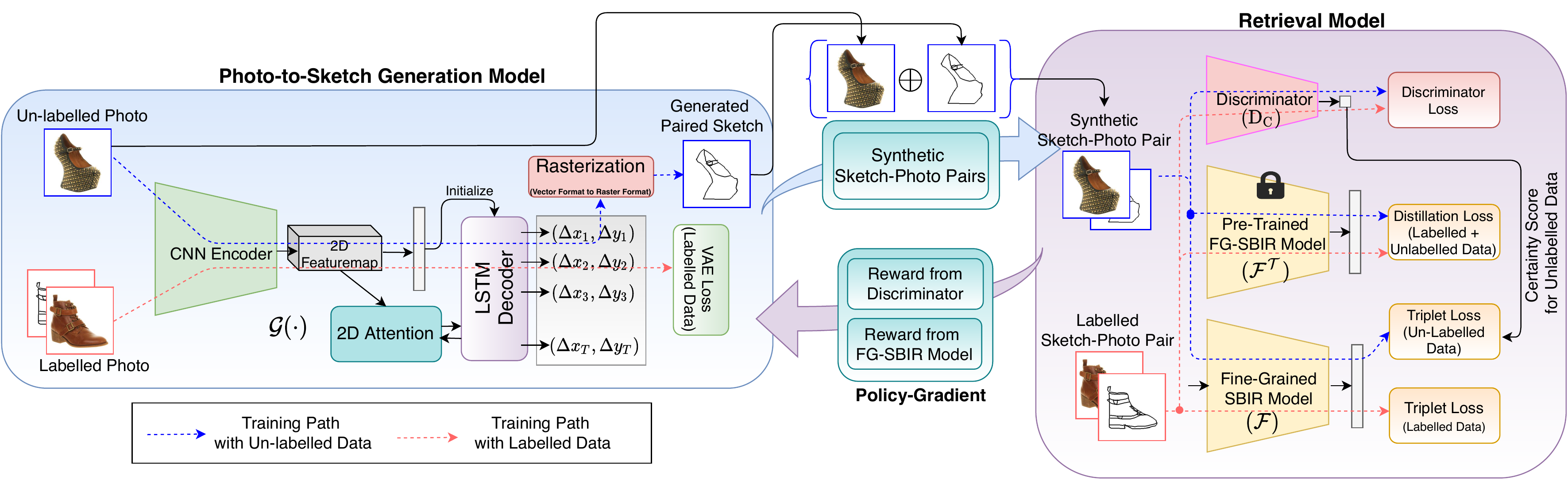}
\end{center}
\vspace{-.20in}
  \caption{Our framework: a FG-SBIR model ($\mathcal{F}$) leverages large scale unlabelled photos using a \emph{sequential}   photo-to-sketch generation model ($\mathcal{G}$) along with labelled pairs. Discriminator ($\mathrm{D_C}$) guided instance-wise weighting and distillation loss are used to guard against the noisy generated data. Simultaneously, $\mathcal{G}$ learns by taking reward from $\mathcal{F}$ and $\mathrm{D_C}$ via policy gradient (over both labelled and unlabelled) together with supervised VAE loss over labelled data. Note rasterization (vector to raster format) is a non-differentiable operation.}
\vspace{-.55cm}
\label{fig:Fig12352}
\end{sidewaysfigure}

Firstly, we train the  photo-to-sketch generation model and discriminative fine-grained FG-SBIR model independently using the labelled training set $\mathrm{\mathcal{D}_{L}}$. Thereafter, through our semi-supervised learning framework,  $\mathcal{F}(\cdot)$ starts exploiting the unlabelled photos to improve its retrieval performance, while enhancing sketch generation quality of $\mathcal{G}(\cdot)$ by using $\mathcal{F}(\cdot)$ as a critic to provide training signal to the sketch-generation model $\mathcal{G}(\cdot)$ using both unlabelled and labelled photos simultaneously. Hence, both $\mathcal{G}(\cdot)$  and  $\mathcal{F}(\cdot)$ can now improve itself with the help of each other and by exploiting unlabelled photos (see Figure \ref{fig:Fig12352}).  

\subsection{Certainty Score for Synthetic Photo-Sketch Pair} The generated photo-sketch pairs $\mathcal{D}_{U}'$ of unlabelled photos are sometimes noisy compared to real labelled photo-sketch pairs $\mathcal{D}_{L}$. This is mainly due to large possible output space \cite{song2018learning} of sketch drawing even with respect to a particular photo, as well as  difficulties in predicting the sketch ending token \cite{ha2017neural} in the sequential decoding process. Every synthetic photo-sketch instance pair needs to be handled individually based on their quality, thus requiring a specific \emph{certainty score} -- signifying the reliability of synthetic photo-sketch pair to train the retrieval model. Existing semi-supervised classification approach usually considers the probability distribution over classes to filter out noisy samples based on a predefined threshold \cite{yalniz2019billion}, top-K selection \cite{sohn2020fixmatch}, or uses entropy-based instance-wise weighting \cite{iscen2019label} to deal with noisy synthetic labels. 
A new solution is thus needed to not just measure the  quality of generated sketch itself, but to quantify how the generated sketch matches with the particular photo input, in order to help training the retrieval model. 

Inspired by the generative adversarial network \cite{chongxuan2017triple} where the sigmoid normalised output of discriminator shows the probability of being a real vs fake input sample, we use the \emph{discriminator's confidence} to quantify the quality of synthetic photo-sketch pairs. Specifically, the discriminator $\mathrm{D_{C}}$ learns to classify between real photo sketch pairs and generated pseudo photo-sketch pairs  (concatenated across channels). Thus, the learning objective for $\mathrm{D_{C}}$ is 

\vspace{-0.7 cm}
\begin{multline}\label{discri}
\mathrm{L_{D_C} = -\mathbb{E}_{({p_{L}; \,{s_{p,L}}}) \sim  \mathcal{D}_{L}} \big[\log \mathrm{D_{C}}\big(p_{L}, s_{p, L}\big)\big]}\\[-2pt]
\mathrm{-\mathbb{E}_{({p_{U};\, {s_{p,U}}}) \sim  \mathcal{D}_{U}'}\big[\log\big(1- \mathrm{D_{C}}\big(p_{U}, s_{p, U}\big)\big)\big]}.
\end{multline}

\noindent This objective is computed via a binary cross-entropy loss using label $1$ for real pairs, and $0$ for synthetic ones. Thus, the discriminator's output  $\mathrm{\mathrm{D_{C}}(p_{U}, s_{p, U}) \in [0,1]}$ signifies the extent to which the synthetic photo-sketch pairs match with the distribution of real labelled photo-sketch pairs. Therefore, values closer to $1$ indicate better quality synthetic photo-sketch pairs. 

\subsection{Tolerance against Noisy Pseudo-Labelled Data} \label{NT}To further avoid over-fitting to noisy synthetic photo-sketch pairs, we introduce a consistency loss with respect to a pre-trained (on labelled dataset) retrieval model  as \emph{weak teacher} \cite{furlanello2018born}. More specifically, once the baseline FG-SBIR model is trained from labelled data, we keep a copy as $\mathcal{F}^T$ with weights frozen. As $\mathcal{F}^T$ has been trained from real clean photo-sketch pairs only, we expect that the feature embedding vector obtained from it would act as an additional supervision via distillation \cite{hinton2015kd} to regularise the main FG-SBIR model ($\mathcal{F}$) which is to be trained from both labelled  and synthetic photo-sketch pairs in a semi-supervised manner. This distillation process is expected to improve the tolerance against noisy information of synthetic data. Compared to cross-entropy loss \cite{hinton2015distilling} used in distillation of classification network, a naive choice to design distillation for feature embedding network is to  minimise the distance between learnable student's embedding and teacher's embedding of a particular photo or sketch image individually. We term it \emph{absolute teacher}. However, instead of considering the actual embedding, we hypothesise that the \emph{relative distance} between paired photo and sketch, minimising which is the major purpose of embedding network, could be a better knowledge to be distilled. We term this as \emph{relative teacher}. Thus, given a photo-sketch image pair ($p, s_p$) and  $d(\cdot, \cdot)$ being a ${l_2}$ distance function,  the consistency loss for learnable student $\mathcal{F}$ with respect to pre-trained teacher $\mathcal{F}^T$ becomes as follows:  
 
\begin{equation}\label{eqn_3}
\mathrm{L}_{\mathcal{F}}^{KD}  =   \big\| d\big(\mathcal{F}^T(p),\mathcal{F}^T(s_p)\big) - d\big(\mathcal{F}(p),\mathcal{F}(s_p)\big) \big\|_{2}.
\end{equation}
\vspace{-0.75 cm}

\subsection{Joint Training} \label{jointtraining}
 \keypoint{Optimising FG-SBIR Model:} We train the fine-grained SBIR model $\mathcal{F}$ using triplet loss over labelled photo-sketch pairs $\mathcal{D}_L$, \emph{instance weighted triplet loss} over generated pseudo photo-sketch pairs $\mathcal{D}_U$, and pre-trained teacher based consistency loss over both  $\mathcal{D}_L$ and $\mathcal{D}_U$. Given sampled data  $\mathcal{D}_L^i=\{p_L^i, s_{p,L}^i\} \sim \mathcal{D}_L$ and 
 $\mathcal{D'}_U^j=\{p_U^j, s_{p,U}^j\} \sim \mathcal{D}_U'$ (on \emph{same} ratio), the instance-wise weight is calculated as $\mathrm{\omega_{j} = D_C(\mathcal{D'}_U^j)}$. The overall semi-supervised loss to train the retrieval model becomes: 
 
 \vspace{-0.1 cm}
\begin{equation}\label{train_retrieval}
\mathrm{L}_{\mathcal{F}}^{all} =  L_{\mathcal{F}}^{trip}(\mathcal{D}_L^i) + \omega_j \cdot L_{\mathcal{F}}^{trip}(\mathcal{D'}_U^j)  + \lambda_{kd}  \cdot L_{\mathcal{F}}^{KD}(\mathcal{D}_L^i, \mathcal{D'}_U^j)
\vspace{0.05cm}
\end{equation}

\keypoint{Optimising Photo-to-Sketch Model:} Besides the fully-supervised VAE loss $L_{\mathcal{G}}^R$, during joint training, the photo-to-sketch generation model is also learned considering $\mathcal{F}$ and $\mathrm{D_C}$ as critics. In particular, if the generated sketch from $\mathcal{G}$ correctly depicts the corresponding input photo, the triplet loss for that generated photo-sketch pair from retrieval model would be low, signifying a better photo-sketch matching and generated sketch quality. Similarly, the higher the discriminator's output, the better the quality of generated photo-sketch pairs. However, these training signals from $\mathcal{F}$ and $\mathrm{D_C}$ cannot  be directly back-propagated to $\mathcal{G}$, as there exists a non-differentiable \emph{rasterization} operation $s_p$ before feeding the sketch-image to both retrieval model and discriminator.  Hence, we employ reinforcement learning  based on  policy-gradient  \cite{sutton2000policy}  with REINFORCE \cite{williams1992simple} deployed to estimate gradients with respect to parameters $\theta_{\mathcal{G}}$ of  $\mathcal{G}$ given some \emph{reward}. As $\mathcal{G}$ aims to lower this triplet loss value $L_{\mathcal{F}}^{trip}$ (Eqn.~\ref{triplet}), the reward should be negative of  $L_{\mathcal{F}}^{trip}$ that needs to be maximised. Similarly, the discriminator's output quantifying the goodness of photo-sketch pairs needs to be maximised. Thus the weighted joint reward is:
 
  \vspace{-0.3 cm} 
\begin{equation}\label{reward}
\mathrm{\mathrm{R}_{\mathcal{G}} = -\lambda_{r1} \cdot L_{\mathcal{F}}^{trip}(\mathcal{D}^i) + \lambda_{r2} \cdot D_C(\mathcal{D}^i)}
\end{equation}

\noindent where  $\mathcal{D}^i \sim \mathcal{D}_L \cup \mathcal{D}_U$. This reward could be computed for both labelled and unlabelled data as it does not need any ground-truth sketch-coordinates unlike the $L_{\mathcal{G}}^{vae}$ loss (Eqn.~\ref{vae_loss}). Thus two types of gradients are computed to update the parameter $\theta_{\mathcal{G}}$, one using policy gradient  \cite{sutton2000policy}  based on joint-reward guided by the retrieval model and the discriminator, and the other using back-propagation over only the labelled photos: 

  \vspace{-0.3 cm}
\begin{equation}\label{g_train}
\begin{split}
 \nabla_{\theta_{\mathcal{G}}}L(\theta_{\mathcal{G}}) &= \underbrace{\nabla_{\theta_{\mathcal{G}}}L_{\mathcal{G}}^{vae}(\theta_{\mathcal{G}})}_\text{over only labelled data} \\
& \mathrm{-}\lambda_\mathcal{G}\sum_{i=1}^{T}\underbrace{\mathbb{E}_{\substack{p_i\sim p(q_i) \\ \Delta z_i\sim p(\Delta z_i \mid \lambda_i)}}\nabla_{\theta_{\mathcal{G}}} \Big( \log p(\Delta z_i \mid \lambda_i) + \log p(p_i) \Big) \cdot R_{\mathcal{G}}}_\text{over both labelled and unlabelled data (via policy gradient)} 
\end{split}
\end{equation}
\vspace{-0.4 cm}

\setlength{\textfloatsep}{0pt}
\vspace{0.5cm}
\begin{algorithm}[!b]
\sloppy
	\caption{Training of Semi-Supervised FG-SBIR}
	\begin{algorithmic}[1] 
        \State \textbf{Input}: Labelled photo-sketch pairs $\mathcal{D}_L$ and Unlabelled photos $\mathcal{D}_U$.  
        \label{algo}
        \State \textbf{Initialise hyper params}: $k_r$, $k_g$.
        \State \textbf{Pre-training}: $\mathcal{G}$ and $\mathcal{F}$ from $\mathcal{D}_L$ (using $\mathrm{L_{\mathcal{G}}^{vae}}$ \& $\mathrm{L_{\mathcal{F}}^{trip}}$). \cut{using $\mathrm{L_{\mathcal{G}}^{vae}}$ and $\mathrm{L_{\mathcal{F}}^{trip}}$ respectively.} 
        \While {not done training}
            \For {$k_{r}$ steps}
                \State Sample data $\mathcal{D}_L^{i} \sim \mathcal{D}_L$ and $\mathcal{D}_U^j \sim \mathcal{D}_U$.
                \State Get synthetic paired images $\mathcal{D'}_U^j$ using $\mathcal{G(\cdot)}$.
                \State \textsc{Train} $\mathcal{F}$ using $\{\mathcal{D}_L^i, \mathcal{D'}_U^j\}$ \Comment{Eqn.~\ref{train_retrieval}}
                \State \textsc{Train} $D_C$ using $\{\mathcal{D}_L^i, \mathcal{D'}_U^j\}$ \Comment{Eqn.~\ref{discri}}
            \EndFor
            \For {$k_{g}$ steps}
                \State Sample data $\mathcal{D}_L^{i} \sim \mathcal{D}_L$ and $\mathcal{D}_U^j \sim \mathcal{D}_U$.
                \State Get reward $R_{\mathcal{G}}$ using $\mathcal{F}$ and $D_C$.
                \State \textsc{Train} $\mathcal{G}$ using $\{\mathcal{D}_L^i, \mathcal{D}_U^j\}$ \Comment{Eqn.~\ref{g_train}}
            \EndFor
        \EndWhile
     \State \textbf{Output}: Optimised models $\mathcal{F}$, $\mathcal{G}$ and $\mathrm{D_C}$.
	\end{algorithmic}
\end{algorithm}  
\vspace{0.5cm}

\noindent In our experiments, we only update the final, fully-connected layer of sketch-decoder (with weights $W_y, b_y$ predicting 6M+3 outputs at every time step), at times using policy gradient, keeping rest of the parameters of $\mathcal{G}$ fixed. {We use a single global reward for the whole sketch-coordinate sequence, instead of local reward at every time step, that would otherwise need costly Monte Carlo roll-outs \cite{yu2017seqgan}}. Note that in our design, $\mathcal{G}$  and $\mathrm{D_C}$ are connected in a GAN-like fashion \cite{goodfellow2014generative} having adversarial objective. Moreover, the retrieval and generative models are trained alternatively improving each other over time (Algorithm \ref{algo}).  

\vspace{-0.2cm} 
\section{Experiments}
    \vspace{-0.2cm} 
\subsection{Dataset} 
 Two publicly available datasets, QMUL-Shoe-V2 \cite{pang2019generalising, riaz2018learning, song2018learning, bhunia2020sketch} and QMUL-Chair-V2 \cite{bhunia2020sketch, song2018learning} are used, which contain stroke-level coordinate information of sketches in addition to instance-wise paired sketch-photo labels, thus enabling us to train both retrieval and sketch-generative models. Out of the 6,730 sketches and 2,000 photos in Shoe-V2,  6,051 and 1,800 for training respectively, and the rest are for testing \cite{bhunia2020sketch, song2018learning}. The splits \cite{bhunia2020sketch, song2018learning} for Chair-V2 dataset are 1,275/725 sketches and 300/100 photos for training/testing respectively. In addition to these  labelled training data, we further use all 50,025 UT-Zap50K images \cite{yu2014fine} as unlabelled photos for shoe retrieval, and 7,800 unlabelled chair photos \cite{pang2020solving} are collected from shopping websites, including IKEA, Amazon and Taobao.  \\

\subsection{Implementation Details} 
Firstly, for sketch-generation, we use ImageNet pre-trained VGG-16 as encoder, excluding any global average pooling operation. We keep the  dimension ($N_z$) of $z$ as 128, the hidden state of the decoder LSTM as 512, the embedding dimension of the 2D-attention module as 256 respectively. \cut{A similar pre-training strategy to \cite{song2018learning} is adopted using rasterized sketches of QuickDraw dataset \cite{ha2017neural}.} We set the max sequence length to $100$, and the generative model is trained with a batch size of 64 with $\omega_{kl} = 1$ using the pre-training strategy from \cite{song2018learning}. Secondly, the retrieval model (ImageNet pre-trained Inception-V3 \cite{szegedy2016rethinking} ) is trained with a batch-size of 16 with a margin value of 0.3. Finally, after completing individual training from labelled data, we start \emph{joint training} (Section \ref{jointtraining}) by additionally exploiting unlabelled data using $k_g$ and $k_r$ as 5. Architecture of  $\mathrm{D_C}$ is from  \cite{isola2017image}. The discriminator used in the joint-training process has a similar design to \cite{isola2017image}.
We set $\lambda_{kd}=0.1$, $\lambda_{r1}=1$, $\lambda_{r2}=1$, and $\lambda_{\mathcal{G}}=10$ respectively. All images are resized to $256\times256$, with rasterization from sketch-coordinate involving a window of same size having centre scaling as well. We use Adam optimiser for both the generation and retrieval models with a learning rate of $0.0001$. 

\subsection{Evaluation Protocol} 
\textbf{(a) FG-SBIR:} Following existing FG-SBIR works \cite{yu2016sketch, pang2020solving}, we use $\mathrm{Acc@q}$, i.e. percentage of sketches having true-paired photo appearing in the top-q list. \textbf{(b) Sketch Generation:} Following \cite{song2018learning, sketchxpixelor}, sketch-generation is quantified from three perspectives (i) \emph{Recognition:} Using a ResNet-50 classifier trained on 250-classes from TU-Berlin sketch dataset, a generated sketch getting recognised as the same class as that of corresponding photo signifies category-level realism. (ii) \emph{Retrieval:}\footnote{Note: Retrieval accuracy is used to quantify both FG-SBIR and sketch generation performance. Please refer to \cite{song2018learning} for more details.} To judge whether the generated sketch has object-instance specific agreement, we check the retrieval accuracy $\mathrm{Acc@q}$ via a pre-trained FG-SBIR model using the generated sketches to retrieve corresponding photos of the testing set. (iii) \emph{Generation:} Following a recent sketch generation work \cite{sketchxpixelor}, we further calculate FID-score \cite{heusel2017gans} using a pre-trained sketch-classifier that captures both the quality and diversity of generated data compared to real human sketches.

\subsection{Competitor} 

 \keypoint{Sketch Generation:}  Sketch Generation could  be approached
in two following ways:  (a) \textit{Image-to-image translation} pipeline: \textbf{Pix2Pix} \cite{isola2017image} could be adapted to perform cross-modal translation in the image space. \textbf{PhotoSketch} \cite{li2019photo} extends further to handle the one-to-many possible nature of photo-conditioned sketch image generation problem, by calculating a mean loss over multiple sketches corresponding to a particular photo. (b) \textit{Image-to-sequence generation} pipeline:  \textbf{Pix2Seq} \cite{chen2017sketch} is the ablated version of our model having a convolutional encoder and LSTM decoder, without involving 2D-attention. \textbf{L2S} \cite{song2018learning} is an extension over \cite{chen2017sketch} that uses two-way cross domain translation with self-domain reconstruction for better regularisation.  \textbf{Ours-G} is a \emph{supervised} model with 2D-attention, trained independently from labelled data only. \textbf{Ours-G-full} is our final sketch-generative model involving joint-training to learn from both labelled and unlabelled  data. 

\keypoint{Fine-Grained SBIR:} We  compare with three groups of competitors. (a) \emph{state-of-the-art:} \textbf{SN-Triplet} \cite{yu2016sketch} employs triplet ranking loss with Sketch-a-Net as its baseline feature extractor. \textbf{SN-HOLEF} \cite{song2017deep}  is an extension over \cite{yu2016sketch}  employing spatial attention along with higher order ranking loss. \textbf{SN-RL} \cite{bhunia2020sketch} is a very recent work employing reinforcement learning based fine-tuning for on-the-fly retrieval. As early retrieval is not our objective, we cite result at sketch-completion point. (b) \emph{Exploiting unlabelled photos:}  There has been no prior work addressing semi-supervised learning for  FG-SBIR, and model designed for category-level retrieval \cite{jang2020generalized} does not fit here. We thus adopt a few works that could be used to leverage unlabelled photos. \textbf{Edgemap-Pretrain} \cite{radenovic2018deep} is a naive-approach to use edge-maps of unlabelled photos to pre-train the retrieval model.  While edge-maps hardly have any similarity to real free-hand sketches, they could be converted to better pseudo-sketches using the work \cite{riaz2018learning} that learns how to abstract sketches based on subset-stroke selection. We term it as \textbf{Edge2Sketch} \cite{riaz2018learning}. Recently, \textbf{Jigsaw-Pretrain}  \cite{pang2020solving} used jigsaw solving over the mixed patches between a particular photo (unlabelled) and its edge-map, as a pre-text task for self-supervised learning (SSL) to improve FG-SBIR performance. Furthermore, we term our self-implemented \emph{supervised} FG-SBIR model trained only on labelled data as  \textbf{Ours-F}. \textbf{Ours-F-Full} is our final retrieval model employing joint training over both labelled and unlabelled photos. We also replace our 2D-attention based sketch-generation process by baseline sketch-generative model Pix2Pix \cite{isola2017image} and L2S \cite{song2018learning}, and term them as \textbf{Ours-F-Pix2Pix} and \textbf{Ours-F-L2S}  respectively. Finally, we design a naive semi-supervised FG-SBIR baseline (\textbf{Vanilla-SSL-F}), where we \emph{blindly} (without instance-weighting and distillation) use the generated sketch to additionally train the retrieval model.

\setlength{\tabcolsep}{1.5pt}
\begin{table}[!htbp]
    \caption{Quantitative results of photo-to-sketch generation}
      \centering
    \scriptsize
    \vspace{-0.25cm}
    \begin{tabular}{cccccc}
        \midrule
        \multirow{2}{*}{Chair-V2} & \multicolumn{2}{c}{Recognition ($\uparrow$)} & \multicolumn{2}{c}{Retrieval($\uparrow$)} & \multirow{2}{*}{FID Score($\downarrow$)} \\
        \cmidrule{2-5}
         & $\mathrm{Acc.@1}$ & Acc.@10 & Acc.@1 & Acc.@10 & \\
        \midrule
        Pix2Pix \cite{isola2017image} & 4.5\% & 12.1\% & 2.4\% & 16.2\% & 33.4 \\
        PhotoSketch \cite{li2019photo} & 7.1\% & 14.3\% & 4.2\% & 17.9\% & 25.7 \\
        Pix2Seq \cite{chen2017sketch} & 5.4\% & 52.1\% & 4.0\% & 31.8\% & 14.5 \\
        L2S \cite{song2018learning} & 12.3\% & 53.8\% & 8.3\% & 36.7\% & 12.7 \\\hdashline
        Ours-G  (only labelled data)  & \blue{15.2\%} & \blue{56.9\%} & \blue{13.4\%} & \blue{40.7\%} & \blue{10.1} \\
        Ours-G-Full & \red{16.4\%} & \red{58.6\%} & \red{14.9\%} & \red{42.6\%} & \red{8.9} \\
        \toprule
        \multirow{2}{*}{Shoe-V2} & \multicolumn{2}{c}{Recognition($\uparrow$)} & \multicolumn{2}{c}{Retrieval($\uparrow$)} & \multirow{2}{*}{FID Score ($\downarrow$)} \\
        \cmidrule{2-5}
         & Acc.@1 & Acc.@10 & Acc.@1 & Acc.@10 & \\
        \midrule
        Pix2Pix \cite{isola2017image} & 6.2\% & 14.5\% & 1.8\% & 8.4\% & 31.7\% \\
        PhotoSketch \cite{li2019photo} & 8.9\% & 17.3\% & 3.4\% & 10.2\% & 24.3\% \\
        Pix2Seq \cite{chen2017sketch} & 51.3\% & 86.6\% & 5.1\% & 25.8\% & 11.3\% \\
        L2S \cite{song2018learning} & 53.7\% & 89.7\% & 6.2\% & 28.6\% & 10.7\% \\\hdashline
        Ours-G (only labelled data) & \blue{56.3\%} & \blue{91.9\%} & \blue{9.7\%} & \blue{33.6\%} & \blue{9.5\%} \\
        Ours-G-Full & \red{58.1\%} & \red{93.4\%} & \red{12.3\%} & \red{35.4\%} & \red{8.3\%} \\
        \midrule
    \end{tabular}
    \label{tab:generation}
     \vspace{0.4cm}
\end{table}

\subsection{Performance Analysis}

\begin{figure*}[!hbt]
\begin{center}
  \includegraphics[width=\linewidth]{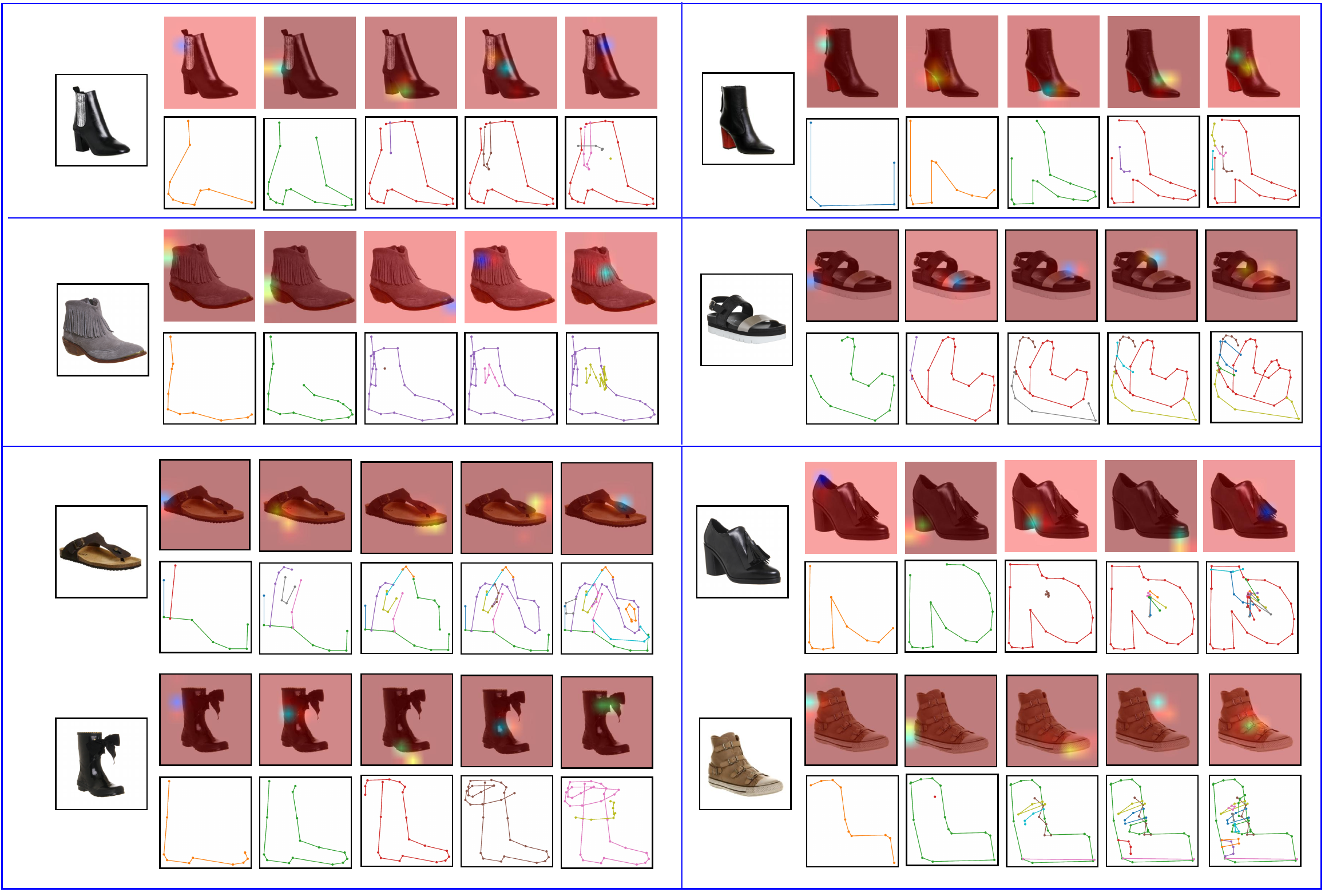}
\end{center}
\vspace{-.20in}
\caption{Qualitative results on our photo-to-sketch generation process, where sketch is shown with attention-map at progressive instances.}
\label{fig:ablation2}
  \vspace{0.2cm}
\end{figure*}

\noindent \textbf{Photo-to-Sketch Generation:} From Table \ref{tab:generation}, we observe: \textbf{(i)} \emph{Pix2Pix} and \emph{PhotoSketch}  based on cross-modal translation in pixel space perform poorly. They fail to capture the abstraction in human sketching style, where distribution gap with real sketches is reflected in their significantly poor FID scores. \textbf{(ii)} \emph{Pix2Seq} and \emph{L2S} outputs vector sketches by sequentially predicting sketch coordinates, thus possessing higher similarity towards human sketches. 
They however, still lag behind our ablated version \emph{Ours-G} in scores.
As both of them reduce the spatial dimension of the convolutional feature-map to a global context vector, spatial information is significantly compromised, with the decoder receiving little guidance from the vector on exact drawing content. In contrast, we retain the spatial dimension of feature-map, and employ 2D-attention to focus on that specific part of the photo it draws at any time step.  \textbf{(iii)} \emph{L2S} is a notably close competitor to ours in terms of recognition accuracy, but better information passage between every time step of decoder and convolutional encoder delivers much better sketches with fine-grained details (reflected by retrieval accuracy).  Furthermore, our final model \textit{Ours-G-Full} employs joint training with a retrieval model to additionally exploit the unlabelled photos, improving sketch generation performance  (retrieval Acc@1) from $9.7\%$ to $12.3\%$ by $2.6\%$ over our baseline \emph{Ours-G} on Shoe-V2, thus justifying the benefits of our semi-supervised learning. Some qualitative results are shown in Figure \ref{fig:ablation2}. Blue denotes a supervised baseline, while red is \textit{Ours-G(F)-Full}.
\vspace{0.1cm}\\
\setlength{\tabcolsep}{6pt}
\begin{table}[!htbp]
    \centering
    \caption{Quantitative results of fine-grained SBIR}
    \scriptsize
    \vspace{-0.25cm}
    \begin{tabular}{ccccc}
        \hline
        \multirow{3}{*}{Methods} & \multicolumn{2}{c}{Chair-V2} & \multicolumn{2}{c}{Shoe-V2} \\
         \cmidrule{2-5}
         & Acc.@1 & Acc.@10 & Acc.@1  & Acc.@10 \\
        \midrule 
        SN-Triplet \cite{yu2016sketch} & 47.4\% & 84.3\% & 28.7\% & 71.6\%  \\
        SN-HOLEF \cite{song2017deep} & 50.7\% & 86.3\% & 31.2\% & 74.6\%  \\
        SN-RL \cite{bhunia2020sketch} & 51.2\% & 86.9\% & 30.8\% & 74.2\%\\
        \hdashline
        Edgemap-Pretrain \cite{radenovic2018deep} & 53.9\% & 87.7\% & 33.8\% & 80.9\% \\
        Edge2Sketch-Pretrain \cite{riaz2018learning} & 54.3\% & 88.2\% & 34.2\% & 81.2\% \\
        Jigsaw-Pretrain \cite{pang2020solving} & 56.1\% & 88.7\% & 36.5\% & 85.9\% \\
        \hdashline
        Ours-F  (only labelled data)  & \blue{53.3\%} & \blue{87.5\%} & \blue{33.4\%} & \blue{80.7\%} \\
        Vanilla-SSL-F  & 49.6\% & 85.6\% & 30.6\% & 74.3\% \\
        Ours-F-Pix2Pix  & 53.2\% & 87.5\% & 33.2\% & 80.1\%  \\
        Ours-F-L2S & 57.6\% & 89.4\% & 36.6\% & 84.7\%  \\
        Ours-F-Full & \red{60.2\%} & \red{90.8\%} & \red{39.1\%} & \red{87.5\%} \\
        \hline
    \end{tabular}
    \label{tab:retrieval}
    \vspace{0.4cm}
\end{table}
\noindent \textbf{Fine-grained SBIR:} From Table \ref{tab:retrieval}, we observe: \textbf{(i)} Our baseline retrieval model is noticeably better than \emph{SN-Triplet}, and lies at par with recent state-of-the-art FG-SBIR baselines like \emph{SN-HOLEF} and \emph{SN-RL}. \textbf{(ii)} With regards to exploiting unlabelled photos, \emph{Edgemap-Pretrain} offers marginal improvement while using it on top of our baseline with ImageNet pretrained weights. Aligning with the intuition, while edge-maps are further augmented with \emph{Edge2Sketch} by a subset of stroke selection to model the abstracted nature of sketch over edge-maps, it increases retrieval performance by a reasonable margin. In context of using edge-maps for pre-training, \emph{Jigsaw-Pretrain} provides maximum benefits, but still lags behind our final model \emph{Ours-F-Full}. \textbf{(iii)} While edge-map does not posses sketch abstraction knowledge of human sketching style, our approach of using a sequential photo-to-sketch generation model to generate synthetic photo-sketch pairs for unlabelled photo encodes better knowledge to enhance generalisation.  However, it is noteworthy that  \emph{Vanilla-SSL-F} blindly using synthetic sketch-photo pairs yields performance lower than the supervised one due to overfitting on noisy information. Overall, for fine-grained SBIR, due to our proposed semi-supervised learning, the retrieval accuracy $\mathrm{Acc@1}$ of \emph{Ours-F-Full } increases from   $33.4\%$ to $39.1\%$ by a margin of $5.7\%$ over our baseline \emph{Ours-F} on Shoe-V2. Moreover, replacing our photo-to-sketch generation model by \emph{L2S} and \emph{Pix2Pix} reduces  the same by $2.5\%$  and $5.9\%$ respectively, thus justifying the importance of our sketch generative model with 2D attention.  \textbf{(iv)} Note that policy-gradient based RL scheme could be avoided by using Pix2Pix for sketch generation, and gradient can directly be back-propagated from retrieval to generative model. However, that is still found to be inferior to ours.

\subsection{Ablation Study}\label{abla}
\begin{figure} 
    \centering
    \includegraphics[width=0.49\linewidth, height=4.2cm]{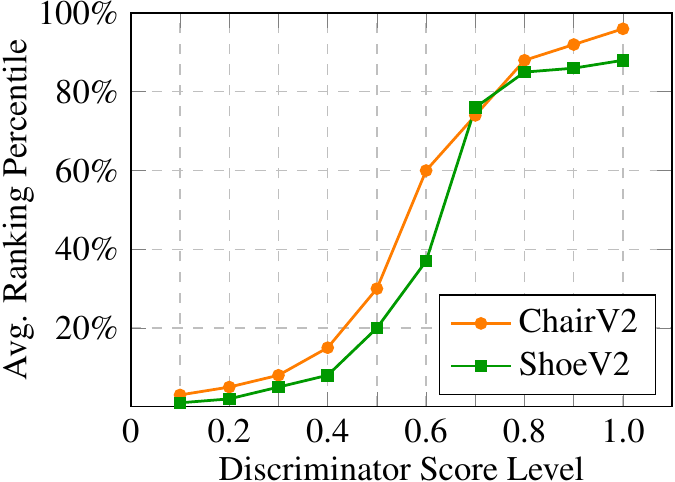}
    \includegraphics[width=0.49\linewidth, height=4.2cm]{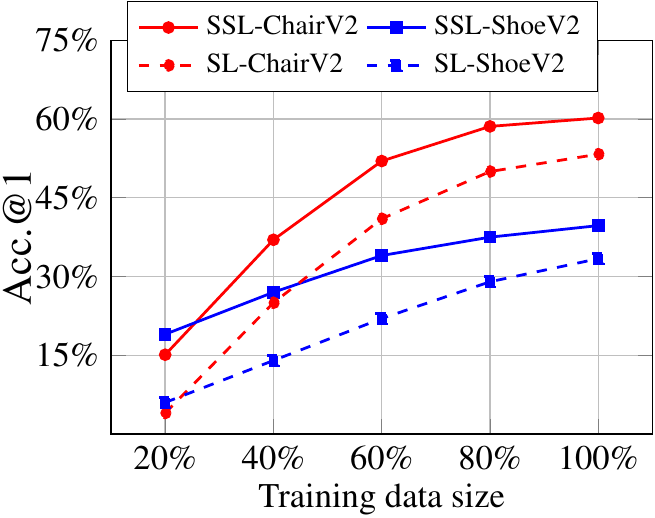}
    \vspace{-0.1cm}
    \caption{(a): Consistency of discriminator's certainty score. (b): Varying training data size for FG-SBIR - Semi-Supervised Learning (SSL) vs Supervised-Learning (SL).}
    \label{fig:ablation3}
    \vspace{0.4cm}
\end{figure}

 A thorough ablative study on Shoe-V2 dataset verifies contributions of individual design components in Table \ref{tab:ablation123}. \textbf{{\emph{[i]} Instance weighting for retrieval}:} To simply judge the contribution of discriminator ($\mathrm{D_C}$) guided instance weighting we remove it, and adapt the framework accordingly. Consequently $\mathrm{Acc@1}$ retrieval performance significantly drops to $36.8\%$ with a decrease of $2.3\%$ on Shoe-V2. Due to sigmoid normalisation \cite{chongxuan2017triple}, the output of $\mathrm{D_C}$ falls in [0,1]. We quantify it as 10 discrete levels with a step size $0.1$. We calculate the average ranking percentile (ARP) of synthetic sketch-photo pairs from testing set which fall under the same discrete level, and plot it against $10$ different levels. From Figure \ref{fig:ablation3} (a), it is evident that the synthetic sketch-photo pairs having higher discriminator score (towards $1$) tend to have much better ARP \cite{bhunia2020sketch} values (i.e better quality), while those with lesser ARP values are assigned with lesser (towards $0$) certainty score by the discriminator. This observation is consistent with our assumption that $\mathrm{D_C}$ should quantify quality of synthetic sketch-photo pairs for instance-wise weighting.  \textbf{{\emph{[ii]}  Distillation based noise tolerance for retrieval}:}  Removing knowledge distillation based regularisation, which additionally tries to provide tolerance against noisy synthetic sketch-photo pairs, $\mathrm{Acc@1}$ is decreased by $1.8\%$ to $37.3\%$ on Shoe-V2 dataset. Our  \emph{relative teacher} based distillation process (Section \ref{NT}) for retrieval network surpasses \emph{absolute teacher} alternative by a margin of $0.9\%$ ($\mathrm{Acc@1}$) on Shoe-V2, thus confirming its usefulness. \textbf{{\emph{[iii]} 2D-attention for sketch-generation}:} The use of 2D attention significantly improves the sketch generation performance, providing better fine-grained agreement with the input photo. While we employ a 3x3 convolutional kernel to aggregate neighbourhood information, using an 1D attention that treats feature maps as 1D sequence, the retrieval accuracy $\mathrm{Acc@1}$ of generated sketches on Shoe-V2 drops to $8.1\%$ by margin of $4.2\%$. We conjecture that 2D-spatial attention has higher efficiency in generating fine-grained sequential sketches from input photo than two-way translation based regularisation as done in \emph{L2S} \cite{song2018learning}.

\setlength{\tabcolsep}{4.5pt}
\begin{table}[!hbt]
    \centering
    \caption{Ablative study on Shoe-V2: Instance Weighting (IW), Teacher Regularisation (TR), Attention (AT), Joint-Training (JT).}
    \scriptsize
    \begin{tabular}{cccccccc}
        \hline
        \multirow{3}{*}{IW} & \multirow{3}{*}{TR} & \multirow{3}{*}{AT} & \multirow{3}{*}{JT} & \multicolumn{2}{c}{Fine-Grained SBIR} & \multicolumn{2}{c}{Sketch Generation} \\
        \cline{5-8}
         & & & & \multirow{2}{*}{Acc.@1} & \multirow{2}{*}{Acc.@10} & Recognition & Retrieval \\
         \cline{7-8}
         & & & & & & Acc.@1 & Acc.@1 \\
        \hline
        \cmark & \cmark & \cmark & \cmark & 39.1\% & 87.5\% & 58.1\% & 12.3\% \\
        \hline
        \xmark & \cmark & \cmark & \cmark & 36.8\% & 85.4\% & 57.3\% & 11.2\% \\
        \cmark & \xmark & \cmark & \cmark & 37.3\% & 86.1\% & 57.8\% & 12.1\% \\
        \cmark & \cmark & \xmark & \cmark & 37.6\% & 86.1\% & 51.3\% & 5.1\% \\
        \cmark & \cmark & \cmark & \xmark & 37.9\% & 86.6\% & 56.3\% & 9.7\% \\
        \hline
        \xmark & \xmark & \xmark & \xmark & 31.1\% & 75.4\% & 51.3\% & 5.1\% \\
        \hline
    \end{tabular}
    \label{tab:ablation123}
     \vspace{0.4cm}
\end{table}%

\noindent \textbf{{\emph{[iv]} Significance of joint-training}:} (a) A direct way of judging efficiency of joint training is employing separately trained photo-to-sketch generation model to augment synthetic sketch-photo pairs, and using them \emph{blindly} to train the retrieval model along with labelled data \emph{without} instance weighting or teacher-regularisation.  This however lags behind the baseline (supervised) fine-grained SBIR model by $2.8\%$ (i.e. $30.6\%$), as the model over-fits to the noisy information present in synthetic sketch-photo data. This confirms that naively using sketch generation does not help at all. (b) The retrieval accuracy ($\mathrm{Acc@1}$) from sketch generation performance improves by $1.4\%$ with an additional policy-gradient based training taking reward (Eqn.~\ref{reward}) from the retrieval model and the discriminator $\mathrm{D_C}$ as critic. Individually, they help to improve by $0.9\%$ and $0.8\%$, respectively under the same metric.  (c) Furthermore, we compute the performance of our semi-supervised framework at varying training data size for both processes in Figure \ref{fig:ablation3} (b). We notice a significant overhead compared to our supervised baseline model for each dataset individually.

\section{Conclusion}
\label{chapter:SSL_FGSBIR:Conclusion}
We have proposed a semi-supervised fine-grained sketch-based image retrieval framework to solve the data scarcity problem. To this end, we proposed to treat sequential photo-to-sketch generation and fine-grained sketch-based image retrieval as two conjugate problems along with various regularizers to address the intricate issues of reliability and tolerance to noisy synthetic sketch-photo pairs. This leads to substantial improvement on existing baselines in sparse data-scenarios for FG-SBIR.


{Besides improving the performance over existing baselines, our semi-supervised framework pushes fine-grained SBIR literature towards a more practical design where good performance could be achieved even in a low-data regime by collecting fewer sketches than earlier. Exploring the practicality angle further, we discover a common challenge: most users find it difficult to sketch and usually come up with sketches that are difficult as queries for efficient retrieval. Turns out, it is not only the quality of the sketch, but the unnecessary addition of irrelevant strokes in pursuit of perfection, that pose a significant bottleneck for retrieval performance.
Therefore, in the following chapter, we design a neural stroke subset selection strategy to eliminate the noisy strokes, thus providing more flexibility and confidence to the user for using the sketch to search and retrieve fine-grained commercial objects.}

\chapter{Noise Tolerant Fine-Grained SBIR}
    \label{chapter:NT_FGSBIR}
    \minitoc
    \newpage
    
    \section{Overview}
    \label{chapter:NT_FGSBIB:Overview}

\aneeshan{Keeping in line with the first theme of practical deployment of FG-SBIR we now focus on what factors restraining the growth of FG-SBIR from the user perspective.}
We observe that despite great strides made \cite{bhunia2021more, pang2019generalising, PartialSBIR}, the \textit{fear-to-sketch} has proven to be fatal for its omnipresence -- a \emph{``I can't sketch"} reply is often the end of it. This \emph{``fear"} is predominant for fine-grained SBIR (FG-SBIR), where the system dictates users to produce even more faithful and diligent sketches than that required for category-level retrieval \cite{collomosse2019livesketch}. In this chapter, we tackle this \emph{``fear"} head-on and propose for the first time a \emph{pre-processing} module for FG-SBIR that essentially let the users sketch without the worry of \emph{``I can't''}. We first experimentally show that, in most cases it is not about how bad a sketch is -- most \textit{can} sketch (even a rough outline) -- the devil lies in the fact that users typically draw irrelevant (noisy) strokes that are detrimental to the overall retrieval performance\cut{(see Section \ref{sec3})}. This observation has largely inspired us to alleviate the \emph{``can't sketch"} problem by \emph{eliminating} the noisy strokes through selecting an optimal subset that \textit{can} lead to effective retrieval.

\begin{figure}[!hbt]
\centering
\includegraphics[width=0.9\linewidth]{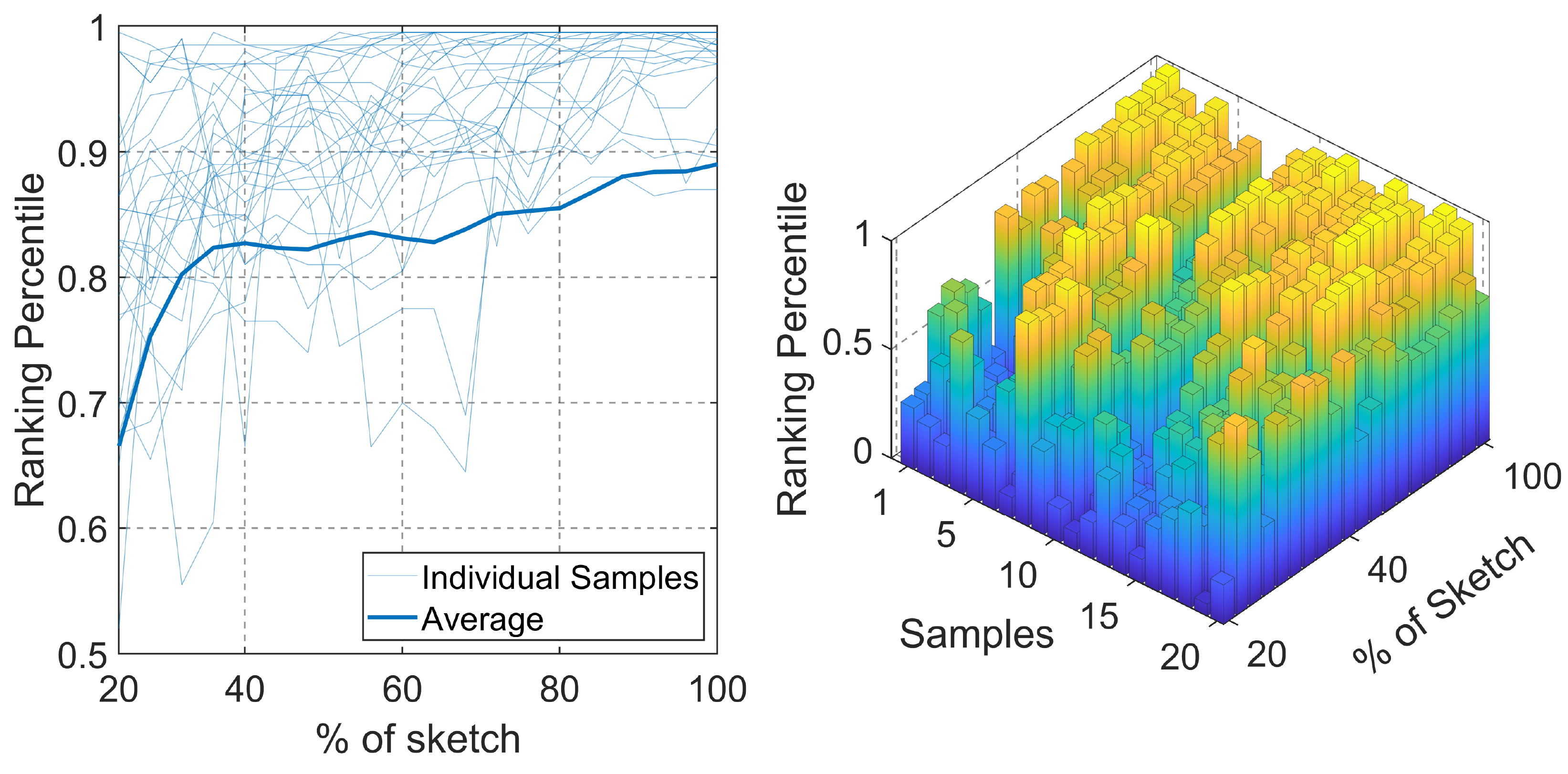}
\vspace{-0.25cm}
\caption{(a) While the \emph{average} ranking percentile increases as the sketching proceeds from starting towards completion, \emph{unwanted sudden drops} have been noticed for many individual sketches due to noisy/irrelevant strokes drawn. (b) The same thing is visualised with number of samples in the third axis to get an overall statistics on QMUL-Shoe-V2 dataset.}
\label{fig1}
\vspace{0.1cm}
\end{figure}
 
This problem might sound trivial enough -- e.g., how about considering all possible stroke subsets as training samples to gain model invariance against noisy strokes? Albeit theoretically possible, the highly complex nature of this process (i.e., $\mathcal{O}(2^N)$) quickly renders this naive solution infeasible, especially when the number of strokes in free-hand sketches can range from an average of $N={9}$ to a max of $N={15}$ in fine-grained SBIR datasets (QMUL-ShoeV2/ChairV2 \cite{yu2016sketch, song2017deep}). Most importantly, augmenting the training data by random stroke dropping would lead to a noisy gradient during training. This is because out of all possible subsets, many of these augmented sketch subsets are too coarse/incomplete to convey any meaningful information to represent the paired photo. Therefore, instead of naively learning the invariance, we advocate for finding meaningful subsets that can sustain efficient retrieval.

Our solution generally rests with detecting noisy strokes and leaving only those that positively contribute to successful retrieval. {We achieve that by proposing a mechanism to quantify the \emph{importance} of each stroke present in a given stroke-set, based on the extent to which that stroke is \emph{worthy for retrieval} (i.e, makes a positive contribution).} We work on vector sketches\cite{sketch2vec} in order to utilise stroke-level information, and propose a sketch stroke subset selector that learns to determine a binary action for every stroke -- whether to include that particular stroke to the query stroke subset, or not. The stroke subset selector is designed via a hierarchical Recurrent Neural Network (RNN) that models the compositional relationship among the strokes. Once the stroke subset is obtained, it is first \emph{rasterized} then passed through a pre-trained FG-SBIR model \cite{yu2016sketch} to obtain a ranking of target photos against the ground-truth photo. The main objective is to select a particular subset that will rank the paired ground-truth photo towards the top of the ranking list. We use Reinforcement Learning (RL) based training due to the non-differentiability of rasterization operation. As explicit stroke-level ground-truth for the optimal subset is absent, we seek to train our stroke-subset selector with the help of pre-trained FG-SBIR for reward computation.  \cut{ that exists between stroke subset selector and pre-trained FG-SBIR.} In particular, we use the actor-critic version of proximal policy optimisation (PPO) to train the stroke subset selector.

Apart from the main objective of noisy stroke elimination, the proposed method also enables a few secondary sketch applications \cut{(Section \ref{byproduct})} in a plug-and-play manner. First, we show that a pre-trained stroke selector can be used as a \emph{stroke importance quantifier} to guide users to produce a sketch ``just'' enough for successful retrieval. 
Second, we demonstrate that it can significantly speed up existing works on interactive ``on-the-fly'' retrieval \cite{bhunia2020sketch} removing the need for incomplete rasterized sketch to be unnecessarily passed for inference multiple times. Third, besides benefiting FG-SBIR, our subset selector module can also act as a faithful \emph{sketch data augmenter} over random stroke dropping without much computational overhead. That is, instead of costly operation like sketch deformation \cite{yu2016sketchAnet} or unfaithful approximation like edge/contour-map as soft ground-truths \cite{chen2018sketchygan}, users can effortlessly generate $n$ most representative subsets to augment training for many downstream tasks. \cut{(e.g., sketch generation \cite{ha2017neural}, vectorization \cite{das2020beziersketch}, combination of both \cite{das2021cloud2curve}).}

In summary our contributions are, (a) We tackle the fear-to-sketch problem for sketch-based image retrieval for the first time, (b) We formulate the ``can't sketch" problem as stroke subset selection problem following detailed experimental analysis, (c) We propose a RL-based framework for stroke subset selection that learns through interacting with a pre-trained retrieval model. (d) We demonstrate our pre-trained subset selector can empower other sketch applications in a plug-and-plug manner.

\section{Pilot Study: What's Wrong with FG-SBIR?}\label{sec3}

\subsection{Baseline FG-SBIR} Instead of complicated pre-training \cite{pang2020solving} or joint-training \cite{bhunia2021more}, we use a three branch state-of-the-art Siamese network \cite{bhunia2021more} as our baseline retrieval model, which is considered to be a strong baseline till date. Each branch starts from ImageNet pre-trained VGG-16 \cite{simonyan2015very}, sharing equal weights. Given an input image $I \in \mathbb{R}^{H \times W \times 3}$, we extract the convolutional feature-map $\mathcal{F}({I})$, which upon global average pooling followed by  $l_2$ normalisation generates a $d$ dimensional feature embedding. This model has been trained with an anchor sketch (a), a positive (p) photo, and a negative (n) photo triplets $\{\bar{a},\bar{p},\bar{n}\}$ using \textit{triplet-loss} \cite{weinberger2009distance}. Triplet-loss aims at increasing the distance between anchor sketch and negative photo $\delta^{-}={||\mathcal F(\bar{a})-\mathcal F(\bar{n})||}_2$, while simultaneously decreasing the same between anchor sketch and positive photo $\delta^{+}={||\mathcal F(\bar{a})-\mathcal F(\bar{p})||}_2$. Therefore, the triplet-loss with margin $\mu>0$ can be written as:
\vspace{-0.2cm}
\begin{equation}
\label{eq1}
    \mathcal{L}_{Triplet}=max\{0, \delta^{+}-\delta^{-}+\mu\}
\vspace{-0.2cm}
\end{equation}
\vspace{-0.6cm}

\subsection{Dual representation of sketch} Recent study has emphasised on the dual representation \cite{sketch2vec} of sketch for self-supervised feature learning. In rasterized pixel modality $\mathcal{I}$, sketch can be represented as spatially extended image of size $\mathbb{R}^{H\times W\times 3}$. On the other side, in vector modality $\mathcal{V}$, the same sketch can be characterised by a sequence of strokes $(s_1, s_2, \cdots, s_K)$ where each stroke is a sequence of successive points $s_i = (v_1^i, v_2^i, \cdots, v_{N_i}^i)$, and each point is represented by an absolute 2D coordinate $v_n^i = (x_n^i, y_n^i)$ in a $H\times W$  canvas. Here, $K$ is number of strokes and $N_i$ is the number of points inside $i^{th}$ stroke. Individual strokes arise due to pen up/down \cite{ha2017neural} movement. 
Although sketch vectors can easily be recorded through touch screen-devices, generation of the corresponding rasterized sketch image needs a costly \cite{xu2021multigraph} \emph{rasterization} operation  $\mathcal{R}: \mathcal{V} \rightarrow \mathcal{I}$. Either modality, raster or vector,  has its own merits and demerits \cite{sketch2vec}. Apart from being more computationally efficient \cite{xu2021multigraph} than raster domain, vector modality also contains the stroke-by-stroke temporal information \cite{ha2017neural}. Nonetheless, sketch vectors lack the spatial information \cite{sketch2vec}  which is critical to model the fine-grained details \cite{bhunia2021more, sketch2vec}. Consequently, rasterized sketch image is the standard choice \cite{pang2020solving, sain2021stylemeup, sain2020cross, yu2016sketch} for FG-SBIR despite having a higher computational overhead and lacking temporal information.

\subsection{Preliminary analysis} The performance barrier due to irrelevant strokes gets noticed under on-the-fly FG-SBIR \cite{bhunia2020sketch} setup. Instead of only evaluating the complete sketch, we start rendering  at the end of every new $k^{th}$ stroke drawn as the rasterized sketch image $S^{\mathcal{I}}_{k} = \mathcal{R}([s_1, s_2, \cdots, s_k])$ where $k=\{1,2, \cdots, K\}$, and pass it through the \emph{pre-trained} baseline FG-SBIR model to get the feature representation $\mathcal{F}(S^{\mathcal{I}}_{k})$, followed by ranking the gallery images against it. We make these following observations on Shoe-V2 \cite{yu2016sketch} dataset \emph{(Linear Limit)}: (i) As the sketch proceeds towards completion, the rank is supposed to be improved, however, we notice some unexpected dips in the performance in the later part of the drawing episode. This signifies that the later irrelevant strokes play a detrimental role, thereby degrading the retrieval performance (Fig.~\ref{fig1}). (ii) Compared to top@1(top@5) accuracy of $33.43\%(67.81\%)$ on using complete sketch for retrieval, if we consider best rank achieved at any of the instant during the sketch drawing episode as the retrieved result, top@1(top@5) accuracy extends to $42.54\%(73.28\%)$. (iii) Further, we note that the percentage of instances where subsequently added strokes drops the performance compared to the previous version $S^{\mathcal{I}}_{k}$ of the same sketch is $43.44\%$, which is a critical number.

\subsection{Ablation on upper limit} 
Prior analysis unfolds the necessity of dealing with irrelevant stroke, and we \emph{hypothesise that in many cases a subset of the strokes $K' \leq K$ could better retrieve the paired photo by excluding the irrelevant ones}. Different people follow varying stroke order for sketching. Therefore, in order to simulate different possible stroke orders and to estimate the upper limit that we can achieve through the smart stroke-subset selector, we do the following study. Given $K$ strokes in a sketch, we form $(2^{K}-1)$ stroke subsets taking \emph{any} number of strokes at a time. Unlike the ``on-the-fly’’ \cite{bhunia2020sketch} protocol,  this setting does not stick to a pre-recorded sequential order, rather it aims to find if there exists \emph{any subset} that can retrieve the paired photo better than the entire sketch set. Under this setting, we achieve an exceptionally high top@1(top@5) accuracy of  $66.37\%(88.31\%)$. However, evaluating with every possible stroke combination during real-time inference is \emph{impractical}, and we do not have any explicit way to select one final result.  Therefore, in this work, we seek to build a \emph{smart stroke-subset selector} as a \emph{pre-processing} module which when plugged in before any pre-trained FG-SBIR model  \cite{yu2016sketch, song2017fine}, will aim to construct the most representative subset to improve the overall accuracy.

\section{Noisy Stroke Tolerant FG-SBIR}
\subsection{Overview} Our preliminary study motivates us to design a stroke-subset selector to eliminate the noisy strokes for FG-SBIR. While \emph{raster sketch image} is essential \cite{bhunia2021more} to model the fine-grained correspondence, the stroke-level sequential information is missing in raster modality. Therefore, taking advantage of the dual representation \cite{sketch2vec} of the sketch, we model the stroke subset selector on the sequential vector space. In summary, our noise-tolerant FG-SBIR consists of two following modules connected in cascade: (a) \emph{stroke-subset selector} as pre-processing module working in vector space and (b) \emph{pretrained FG-SBIR} $(\mathcal{F})$ that uses rasterized version of predicted subset for final retrieval.

\subsection{Stroke Subset Selector}
\keypoint{Model:} Given sketch-photo pair $(S,P)$, the sketch $S$ can be represented as both raster image $S_{{I}}$ and stroke-level sequential vector $S_{{V}}=(s_1, s_2, \cdots, s_K)$. We design a stroke-subset selector $\mathcal{X}(\cdot)$ that takes $S_{{V}}$ as input, and aim to predict an optimal subset $\overline{S}_{V} = \mathcal{X}( S_{V})$ with $K'$ strokes where $K' \leq K$. However, selecting the optimal subset of stroke is an ill-posed problem.  Firstly, there is \emph{no explicit label} which represents the optimal stroke subset. In fact, there might be many sub-sets which can lead to successful retrieval. Furthermore, annotating the optimal stroke-subsets for the whole training dataset via brute-force iteration is computationally impractical \cite{sketchxpixelor}.

In our framework, we treat stroke subset selector as a binary categorical classification problem.  In other words, for a sketch of K strokes, we get an output of size $\mathbb{R}^{K\times2}$, where every row is softmax normalised and it represents a probability distribution $p(a_i|s_i)$ over two classes: $a \in \{\texttt{select}, \texttt{ignore}\}$. However, we do not have any explicit one-hot labels for this binary classification task. Therefore, we let the stroke sub-set selector agent to interact with the pre-trained FG-SBIR model, and $\mathcal{X}$ is learned  using a pre-trained FG-SBIR model $\mathcal{F}$ as a \emph{critic} which provides the training signal to $\mathcal{X}$.

\begin{figure}[!hbt]
\centering
\includegraphics[width=0.9\linewidth]{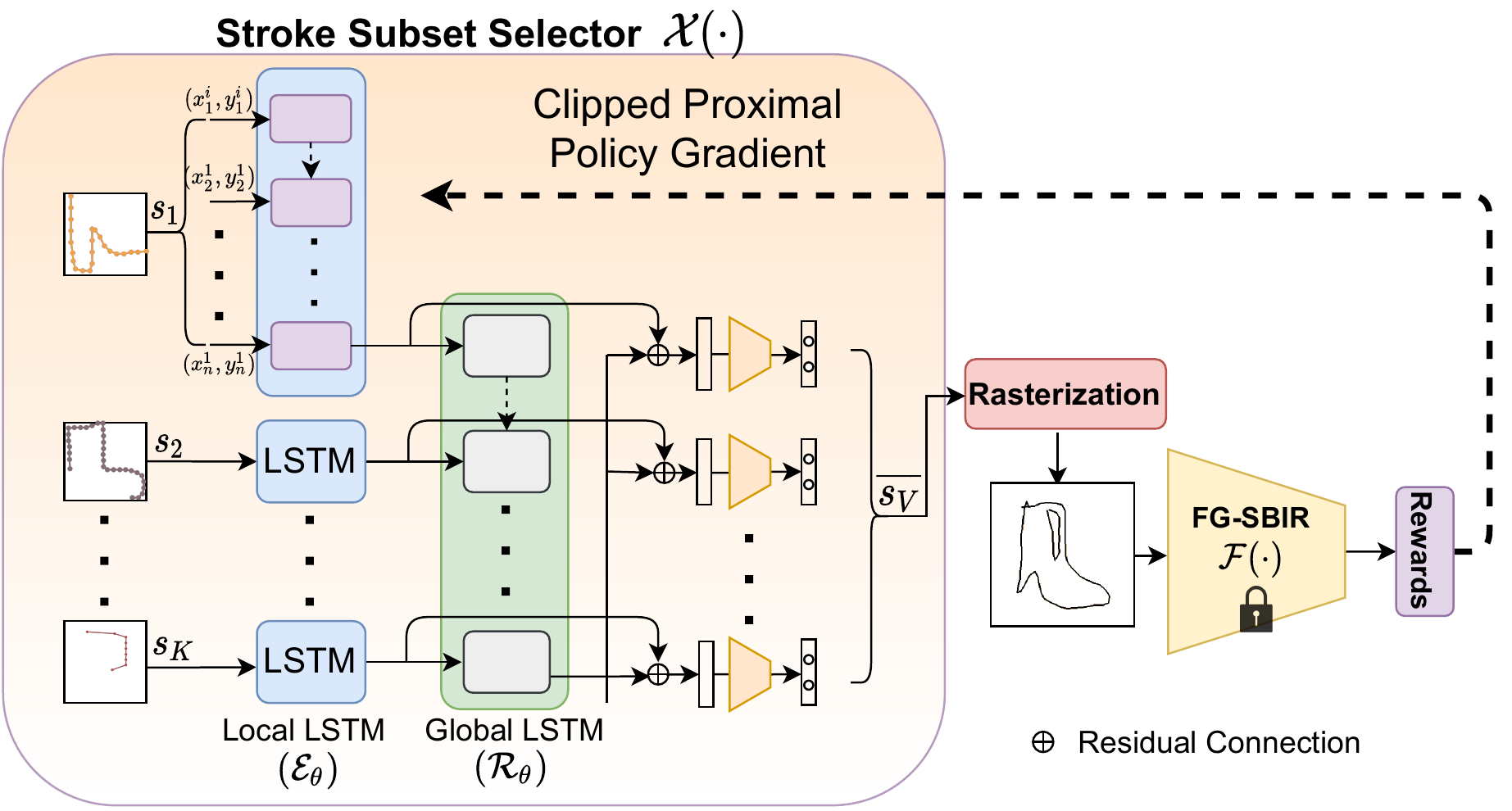}
\vspace{-0.3cm}
\caption{Illustration of Noise Tolerant FG-SBIR framework. Stroke Subset Selector $\mathcal{X(\cdot)}$ acts as a pre-processing module in the sketch vector space to eliminate the noisy strokes. Selected stroke subset is then rasterized and fed through an existing pre-trained FG-SBIR model for reward calculation, which is optimised by Proximal Policy Optimisation. For brevity, actor-only version is shown here. }
\label{fig_archi}
\vspace{0.2cm}
\end{figure}

\keypoint{Architecture:} To design the architecture of stroke-level selector, we aim at preserving localised stroke-level information, as well as the compositional relationship \cite{aksan2020cose} among the strokes, which  together conveys the overall semantic meaning. Therefore, we employ a two-level hierarchical model comprising of a local stroke-embedding network ($\mathcal{E}_\theta$) and global relational network ($\mathcal{R}_\theta$)  to enrich each stroke-level feature about the global semantics. In particular, we feed individual stroke of size $\mathbb{R}^{N_i\times2}$ having $N_i$ points though a local stroke-embedding network $\mathcal{E}_\theta$ (e.g. RNN, LSTM or Transformer) whose weights of $\mathcal{E}_\theta$ are shared across strokes. We take the final hidden-state feature as the localised representation ${f}^{l}_{s_i} \in \mathbb{R}^{d_s}$ for $i^{th}$ stroke. Thereafter, feature representation of $K$ such strokes having size of $\mathbb{R}^{K \times d_s}$ are further fed to a global relational network ($\mathcal{R}_\theta$) whose final hidden state $f^{g} \in  \mathbb{R}^{d_s}$ captures the global semantic information of the whole sketch. Taking inspiration from residual learning \cite{he2016deep}, we fuse the global feature with individual stroke-level feature through a residual connection with LayerNorm \cite{ba2016layer}. In concrete, every stroke feature enriched by global-local compositional hierarchy is represented by $\hat{f}_{s_i} = \texttt{LayerNorm}({f}^{l}_{s_i} + f^{g}) \in \mathbb{R}^{d}$. We implement both $\mathcal{E}_\theta$ and $\mathcal{R}_\theta$ through a one layer LSTM with hidden state size $128$. Further, we apply a shared linear layer ($\mathcal{C}_\theta$) to get $p(a_i|s_i) = \texttt{softmax} (W_\mathcal{X}\hat{f}_{s_i} + b_{\mathcal{X}})$, where $W_\mathcal{X} \in \mathbb{R}^{d_s\times2}$ and $b_\mathcal{X} \in \mathbb{R}^{2}$. We group three modules  $\{\mathcal{R}_{\theta}, \mathcal{E}_{\theta}, \mathcal{C}_{\theta}\}$ of stroke subset selector as $\mathcal{X_\theta}$. See Fig.~\ref{fig_archi}.

\subsection{Training Procedure}
\keypoint{Necessity of RL} Due to the unavailability of ground-truth for optimum strokes, we rely on the pre-trained FG-SBIR model to learn the optimum stroke-subset selection strategy.
In particular, given probability distribution $p(a_i|s_i) \in \mathbb{R}^{2}$ for \emph{every} stroke over $\{\texttt{select}, \texttt{ignore}\}$, we can sample from categorical distribution as $a_i \sim  \small{\texttt{Categorical}} \normalsize([p(a_{select}|s_i), p(a_{ignore}|s_i)])$, and thereby we will be getting a stroke subset as $\overline{S}_{V}$ with $K'$ strokes, where $K' \leq K$. In order to get the training signal from pre-trained FG-SBIR model $\mathcal{F}$, we need to feed the subset sketch through $\mathcal{F}$. For that, we need to convert the sequential sketch vector to raster sketch image through rasterization $\overline{S}_{I} = \mathcal{R}(\overline{S}_{V})$, as fine-grained SBIR model only \cite{bhunia2021more, bhunia2020sketch} works on raster image space. While subset sampling could be relaxed by Gumbel-Softmax \cite{jang2016categorical} operation for differentiablity, non-differentiable rasterization operation $\mathcal{R}(\cdot)$ squeeze us to use Policy-Gradient \cite{sutton2000policy} from Reinforcement Learning (RL) literature \cite{kaelbling1996reinforcement}. 

\keypoint{MDP Formulation}  In particular, given an input sketch $S_{V}$ (initial state), the stroke-subset selector $(\mathcal{X}_{\theta})$ acts a \emph{policy network} which takes \emph{action} on selecting every stroke, and we get an updated state as subset-sketch $\overline{S}_{V}$ (next state). In order to train the policy network, we calculate \emph{reward} using $\mathcal{F}$ as a critic. Therefore, we can form the tuple of four elements \small{(\texttt{initial\_state}, \texttt{action}, \texttt{reward}, \texttt{next\_state})} \normalsize that is typically required to train any RL model. In order to model the existence of multiple possible successful subsets, we unroll this sequential Markov Decision Process (MDP)  $T$ times starting from the complete sketch vector. In other words, for each sketch data, we sequentially sample the subset strokes $T$ times to learn the multi-modal nature of true stroke subsets. Empirically we keep \emph{episode length} $T=5$.

\keypoint{Reward Design} Our objective is to select the optimum set of stroke which can retrieve the paired photo with minimum rank (e.g. best scenario: rank 1). In other words,  pairwise-distance between the query sketch and paired photo embeddings should be lower than that of query sketch and rest other photos of the gallery.  {As $\mathcal{F}$ is fixed, we can pre-compute the features of all $M$ gallery photos as $G \in \mathbb{R}^{M\times D}$ -- thus eliminating the burden of repeatedly computing the photo features. During stroke subset selector training, we just need to calculate the feature embedding $\mathcal{F}(\overline{S}_{I})$ of rasterized version of predicted subset sketch, and we can calculate rank of paired photo using $G$ and paired-photo index efficiently.} We compute the reward both in the ranking space as well as in the feature embedding space using standard triplet loss on $\mathcal{F}(\overline{S}_{I})$ following Eqn. \ref{eq1}, which is found to give better stability and faster training convergence. In particular, we want to minimise the rank of the paired photo and triplet loss simultaneously. Following the conventional norm of reward maximisation, we define the \emph{reward} (R) as weighted summation of inverse of the rank and negative triplet loss as follows:

\vspace{-0.3cm}
\begin{equation}
\label{eq2}
R = \omega_1 \cdot \frac{1}{rank} + \omega_2 \cdot (-\mathcal{L}_{Triplet})
\end{equation}
\vspace{-0.4cm}

\keypoint{Actor Critic PPO:} We make use of actor-critic version of Proximal Policy Optimisation (PPO) with clipped surrogate objective \cite{schulman2017proximal}  to train our stroke-subset selector. In particular, the very basic policy gradient \cite{sutton2000policy}  objective that is to be minimised could be written as: \cut{$L^{PG}(\theta) = -\frac{1}{K}\sum_{i=1}^{K} \log p_{\theta}(a_i | s_i) \cdot R$ }

\vspace{-0.35cm}
\begin{equation}
\label{eq3}
L^{PG}(\theta) = -\frac{1}{K}\sum_{i=1}^{K} \log p_{\theta}(a_i | s_i) \cdot R
\end{equation}
\vspace{-0.35cm}

For sampling efficiency, using the idea of Importance Sampling \cite{neal2001annealed}, PPO maintains an older policy $p'_{\theta}(a_i | s_i)$, and thus
Conservative Policy Iteration (CPI) objective becomes  $L^{CPI}(\theta) = -\frac{1}{K}\sum_{i=1}^{K} r_i(\theta)\cdot R$, where \cut{$r_i(\theta) =\frac{\log p_{\theta}(a_i | s_i)}{\log p'_{\theta}(a_i | s_i)}$.} $r_i(\theta) = {\log p_{\theta}(a_i | s_i)}/{\log p'_{\theta}(a_i | s_i)}$. Further on, the clipped surrogate objective PPO can be written as $L^{CLIP}(\theta) = -\frac{1}{K}\sum_{i=1}^{K} \texttt{clip} (r_i(\theta), 1-\varepsilon, 1-\varepsilon))$, which aims to penalise too large policy update with hyperparameter $\epsilon=0.2$. We take a minimum of the clipped and unclipped objective, so the final objective is a lower bound (i.e., a pessimistic bound) on the unclipped objective. The final \emph{actor only} version PPO objective becomes:

\begin{equation}
L^{A}(\theta) = -\frac{1}{K}\sum_{i=1}^{K} \mathrm{min}(L^{CPI}, L^{CLIP})
\end{equation}

To reduce the variance, the \emph{actor-critic} version of PPO make use of a \emph{learned state-value function} $V(S)$ where $S$ is the sketch vector $S = (s_1, \cdots, s_K)$. {$V(S)$ shares parameter with actor network $\mathcal{X}_{\theta}$, where only the last linear layer ($C_\theta$) is replaced by a new linear layer upon a single latent vector (accumulated stroke-wise features by averaging), predicting a scalar value that tries to \emph{approximate the reward value}.} Thus, the final loss function combines the policy surrogate and value function error time together with a entropy bonus ($E_n$)  to ensure sufficient exploration is:

\begin{equation}
L^{AC}(\theta) = -\frac{1}{K}\sum_{i=1}^{K} (L^{A} - c_1 (V_{\theta}(S) - R)^2 + c_2 E_n )
\end{equation}

where, $c_1$ and $c_2$ are coefficients. As we unroll the sequential stroke-subset selection process for $T=5$, for every sample the loss accumulated over the MDP episode is $\frac{1}{T} \sum_{t=1}^{T}L^{AC}_t(\theta)$.

\section{Applications of Stroke-Subset Selector} \label{byproduct}

\keypoint{\emph{Resistance against noisy strokes:}} Collected sketch labels, which are used to train the initial fine-grained SBIR model are also noisy. The proposed stroke-subset selector \emph{not only assists during inference} by noisy-stroke elimination, but also \emph{helps in cleaning training data}, which in turn can boost the performance to some extent. In particular, we train the FG-SBIR model and Stroke-Subset Selector in stage-wise alternative manner, with the FG-SBIR model using clean sketch labels produced by the trained stroke-subset selector. Our method thus offers a plausible way to alleviate the latent/hidden noises of a FG-SBIR dataset \cite{yu2016sketch}.

\keypoint{\emph{Modelling ability to retrieve:}} As the critic network tries to approximate the scalar reward value which is a measure of retrieval performance, we can use the \emph{critic-network} to quantify the retrieval ability at any instant of a sketching episode. Higher scalar score from the critic signifies better retrieval ability. To wit, we ask the question whether a partial sketch is good enough for retrieval or not. Thus, instead of feeding rasterized partial sketch multiple times for on-the-fly \cite{bhunia2020sketch} retrieval, we can save significant computation cost by feeding \emph{only after} it gains a potential retrieval ability. Moreover, as both our actor and critic networks work in sketch vector modality, it adds less computational burden. 


\keypoint{\emph{On-the-fly FG-SBIR: Training from Partial Sketches:}} State-of-the-art on-the-fly FG-SBIR \cite{bhunia2020sketch} employs continuous RL for training using ranking objective. A supervised triplet-loss \cite{yu2016sketch} based training, augmented with synthetic partial sketches obtained through random stroke-dropping is claimed to be sub-optimal, as randomly dropped strokes frequently banish crucial details, resulting in the augmented partial sketch containing insufficient information to depict the paired photo. In contrast, we use our stroke-subset selector to create several augmented partial versions of the same sketch, each with \emph{sufficient retrievability}. While continuous RL is time intensive to train and allegedly unstable \cite{kaelbling1996reinforcement}, we can use simple triplet-loss based supervised learning \cut{to reduce SOTA on-the-fly FG-SBIR performance by creating}with multiple \emph{meaningful augmented partial sketches}.

\section{Experiments}
\subsection{Dataset} 
 Two publicly available FG-SBIR datasets  \cite{yu2016sketch, pang2019generalising, bhunia2020sketch} namely QMUL-Shoe-V2
and QMUL-Chair-V2 are used in our experiments. Apart from having instance-wise paired sketch-photo, these datasets also contain the sketch coordinate information, and thus would enable us to train the stroke-subset selector using sketch vector modality. We use the standard training/testing split used by the existing state-of-the-arts. In particular, out of $6,730$ ($1,800$) sketches and $2,000$ ($400$) photos from Shoe-V2 (Chair-V2) dataset, $6,051$ sketches ($1,275$) and $1,800$ ($300$) photos are used for training respectively, and the rest are for testing \cite{bhunia2020sketch}.

\subsection{Implementation Details} 
We have conducted all our experiments on an 11-GB Nvidia RTX 2080-Ti GPU with PyTorch. For \emph{fine-grained SBIR}, we have used  ImageNet \cite{russakovsky2015imagenet} pre-trained VGG-16 \cite{simonyan2015very} backbone with feature embedding dimension $d=512$. We train the FG-SBIR model using Adam optimiser \cite{kingma2014adam} with a learning rate of 0.0001, batch size 16, and margin value of 0.2 for triplet loss. For \emph{stroke subset selector}, we model local stroke embedding network and global relational network using one-layer LSTM with hidden state size $128$ for each. The critic network shares the same weights with that of the actor, with only the last linear layer $C_\theta$ being replaced by a new one that predicts a single scalar value. We train it for $2000$ epoch using Adam optimiser with initial learning rate $10^{-4}$ till $100$ epochs, then reducing to $10^{-5}$. We use a batch size of $16$ and keep an old policy network for importance sampling \cite{neal2001annealed} with episode length $T=5$, and sampled instances are stored in a replay buffer. We update the current policy network at every $20$ iteration using sampled instances from the replay buffer, and the old policy network's weights are copied from the current one for subsequent sampling.  We empirically set both $\omega_1$, $\omega_2$ to $1$, and keep $c_1=0.5$, $c_2=0.01$, $\epsilon=0.2$. 

\subsection{Evaluation Protocol} 
 \textbf{(a) Standard FG-SBIR:} Aligning to the existing state-of-the-art FG-SBIR frameworks \cite{pang2020solving, yu2016sketch}, we use percentage of sketches having true-matched photo in the top-1 (acc.@1) and top-5 (acc.@5) lists to assess the FG-SBIR performance. \textbf{(b) On-the-fly FG-SBIR:} Furthermore, to showcase the early retrieval performance from partial sketch, adhering to prior early-retrieval work \cite{bhunia2020sketch} we employ two plots namely, (i)\emph{ranking-percentile}  and (ii)  $\frac{1}{rank}$ vs. \emph{percentage of sketch}. Higher area under these curves indicate better early-retrieval potential. For the sake of simplicity, we call area under curves (i) and (ii) as r@A and r@B through the rest of the chapter.

\subsection{Competitor} 

 To the best of our knowledge, no earlier works have directly attempted to design a Noise-Tolerant FG-SBIR model in the SBIR literature.  Therefore, we compare with the existing standard FG-SBIR works appeared in the literature, as well as, we develop some self-designed competitive baselines under the assumption of \emph{`all sketches are sketchy'} -- which explicitly intend to learn invariance against noisy strokes. {(a)}  \emph{State-of-the-arts} (SOTA): While \textbf{Triplet-SN} \cite{yu2016sketch} uses Sketch-A-Net backbone along with triplet loss,  \textbf{Triplet-Attn-HOLEF}  extends \cite{yu2016sketch} with spatial attention and higher order ranking loss. Recent works include: \textbf{Jigsaw-Pretrain} with self-supervised pre-training, \textbf{Triplet-RL} \cite{bhunia2020sketch}  employing RL-based fine-tuning, \textbf{StyleMeUP} involving MAML training,  \textbf{Semi-Sup} \cite{bhunia2021more} incorporating semi-supervised paradigm, and \textbf{Cross-Hier} \cite{sain2020cross}  utilising cross-modal hierarchy with costly paired-embedding. {(b)}  \emph{Self-designed Baselines (BL)}:  We  create multiple version of the same sketch by randomly dropping strokes (ensuring percentage of sketch vector length never drops below 80\%) or  by synthetically adding random noisy stroke patches similar to \cite{liu2020unsupervised}. \textbf{Augment} aims to learn the invariance against noisy stroke by adding them inside training. This is further advanced by \textbf{StyleMeUp+Augment} where synthetic noisy/augmented sketches are mixed in the inner-loop of \cite{sain2021stylemeup} to learn invariance by optimising outer-loop on real sketches. \textbf{Contrastive+Augment} imposes an additional contrastive loss \cite{chen2020simple} such that the distance between two augmented versions of same sketch should be lower than that of with a random other sketch. Our pre-trained baseline FG-SBIR model is termed as \textbf{B-Siamese}.

\setlength{\tabcolsep}{3pt}
\begin{table}[!htbp]
    \centering
    \caption{Results under Standard FG-SBIR setup.}
    \vspace{-0.3cm}
    \footnotesize
    \begin{tabular}{ccccccc}
        \hline
         &   &   & \multicolumn{2}{c}{Chair-V2} & \multicolumn{2}{c}{Shoe-V2} \\
        \cline{4-7}
         & & & Acc@1 & Acc@5 & Acc@1 & Acc@5\\
        \hline
        \multirow{7}{*}{\rotatebox[origin=c]{0}{SOTA}} 
         & \multicolumn{2}{c}{Triplet-SN \cite{yu2016sketch}} & 47.4\% & 71.4\% & 28.7\% & 63.5\% \\
         & \multicolumn{2}{c}{Triplet-Attn-HOLEF \cite{song2017deep}} & 50.7\% & 73.6\% & 31.2\% & 66.6\% \\
        & \multicolumn{2}{c}{Triplet-RL  \cite{bhunia2020sketch}} & 51.2\% & 73.8\% & 30.8\% & 65.1\% \\ 
         & \multicolumn{2}{c}{Mixed-Jigsaw \cite{pang2019generalising}} & 56.1\% & 75.3\% & 36.5\% & 68.9\% \\
          & \multicolumn{2}{c}{Semi-Sup \cite{bhunia2021more}} & 60.2\% & 78.1\% & 39.1\% & 69.9\% \\
          & \multicolumn{2}{c}{StyleMeUp \cite{sain2021stylemeup}} & 62.8\% & 79.6\% & 36.4\% & 68.1\% \\
           & \multicolumn{2}{c}{Cross-Hier \cite{sain2020cross}} & {62.4\%} & {79.1\%} & {36.2\%} & {67.8\%} \\
         \hdashline
        \multirow{4}{*}{\rotatebox[origin=c]{0}{BL}} 
        & \multicolumn{2}{c}{\blue{(B)aseline-Siamese}} & {\blue{53.3\%}} & {\blue{74.3\%}} & {\blue{33.4\%}} & {\blue{67.8\%}}\\
         & \multicolumn{2}{c}{Augmnt} & {54.1\%} & {74.6\%} & {33.9\%} & {68.2\%}\\
          & \multicolumn{2}{c}{StyleMeUp+Augment} & {56.1\%} & {76.9\%} & {36.9\%} & {69.9\%}\\
        & \multicolumn{2}{c}{Contrastive+Augment} & {58.8\%} & {77.1\%} & {37.6\%} & {70.1\%}\\
        \hdashline
        \multirow{2}{*}{\rotatebox[origin=c]{0}{Limits}} 
        & \multicolumn{2}{c}{Upper-Limit} & 78.6\% & 90.3\% & 66.3\% & 88.3\% \\
         & \multicolumn{2}{c}{Linear-Limit} & 59.4\% & 77.3\% & 42.5\% & 73.2\% \\
         \hdashline
        & \multicolumn{2}{c}{Proposed} & \red{64.8\%} & \red{79.1\%} & \red{43.7\%} & \red{74.9\%} \\
        \hline
    \end{tabular}
    \label{tab:my_label2234}
\vspace{0.5cm}
\end{table}

\subsection{Performance Analysis}

The comparative analysis is shown in Table \ref{tab:my_label2234}. Overall, we observe a significantly improved performance of our proposed Noise-Resistant fine-grained SBIR employing a stroke-subset selector as a pre-processing neural agent compared to the existing state-of-the-art. The early works tried to address different \emph{architectural modifications} \cite{song2017fine, pang2019generalising}, and later on the field of fine-grained SBIR witnessed successive improvements through adaptation of different paradigms like \emph{self-supervised learning} \cite{pang2020solving}, \emph{meta-learning} \cite{sain2021stylemeup}, \emph{semi-supervised learning} \cite{bhunia2021more}, etc. As opposed to these works, we underpin an important phenomenon of noisy strokes, which is inherent to FG-SBIR. Most interestingly, our simple stroke-subset selector can improve the performance of baseline B-Siamese model by an approximate margin of $10.31\%$ without any complicated joint-training of \emph{Semi-Sup} \cite{bhunia2021more}, costly hierarchical paired embedding of \emph{Cross-Hier} \cite{sain2020cross}, or meta-learning cumbersome feature transformation layer of \emph{StyleMeUp} \cite{sain2021stylemeup}. Furthermore, the performance of \emph{Augmnt} baseline is slightly better than our baseline pre-trained FG-SBIR as it learns some invariance from augmented/partial sketch. While we experienced difficulty in stable training for \emph{StyleMeUp+Augment}, \emph{Contrastive+Augment} appears as a simple and straightforward way to learn the invariance against noisy strokes. Instead of modelling invariance, we aim to eliminate the noisy strokes, thus giving a freedom of explainability through visualisation. Despite using complicated architectures \cite{sain2020cross, bhunia2021more}, SOTA fails even to beat the accuracy of Linear-Limit (refer to section \ref{sec3}), while we can. Nevertheless, we suppress it by keeping the simple baseline FG-SBIR untouched and prepending a simple stroke-selector agent -- working on a cheaper vector modality for efficient deployment.

\subsection{Further Analysis and Insights}\label{ablation}
\begin{figure}[!hbt]
\centering
\includegraphics[width=0.85\columnwidth]{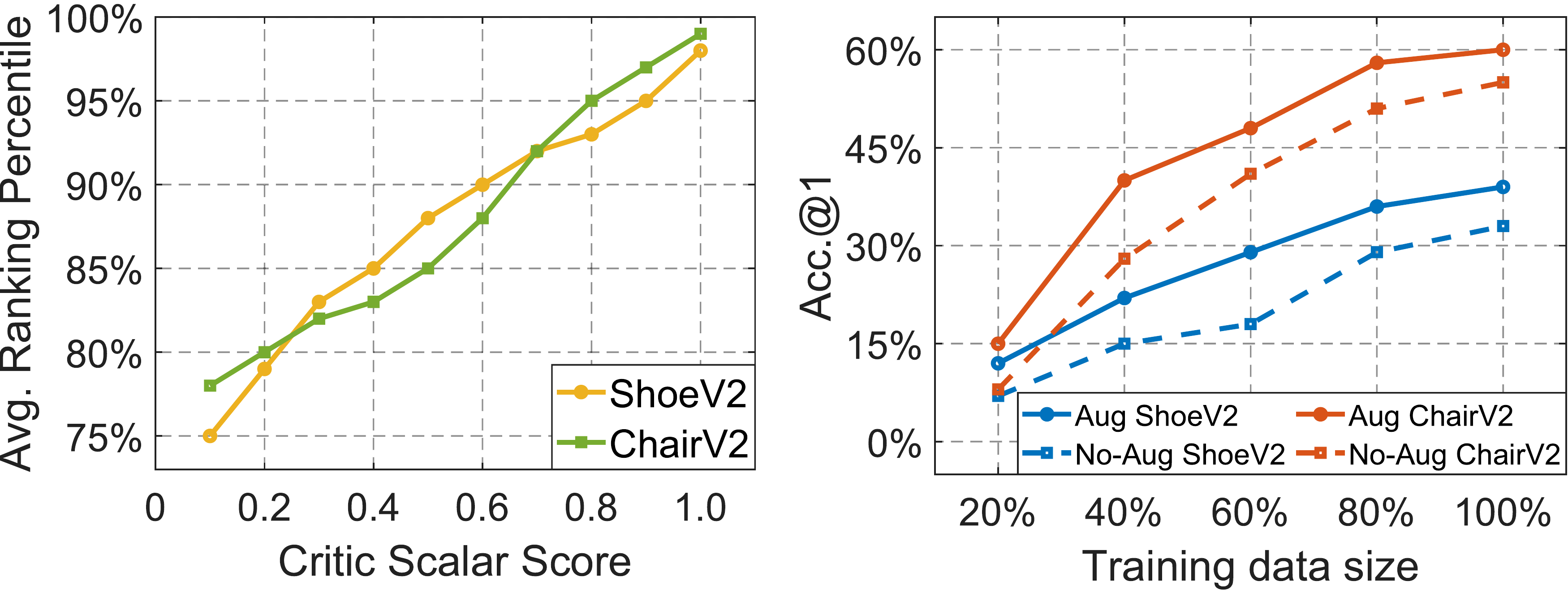} 
\vspace{-0.3cm}
\caption{ (a) Retrieval ability of partial sketch: correlation between critic network $\mathrm{V(S)}$ predicted score and ranking percentile (b) Performance at varying training data size with stroke-subset selector based data augmentation.}
\label{fig_partial}
\vspace{-0.4cm}
\end{figure}

\keypoint{Ability to retrieve/classify for partial sketches:} 
\cut{As described in Section \ref{byproduct}, we first aim to validate the potential of auxiliary critic network to quantify the retrieval ability of partial sketches.} The scalar value predicted by our learned state-value function (critic-network) \cite{schulman2017proximal} signifies the retrieval ability of partial sketch with the notion of higher being the better. We here train our model with a reward of $\frac{1}{rank}$ for easy interpretability. Once the stroke-subset selector with actor-critic version is trained, we feed the sketch to the critic network (in vector space) at a progressive step of $5\%$ completion, and record the predicted scalar value at every instant. At the same time, we rasterize every partial instance and feed through pre-trained FG-SBIR to calculate the resultant ranking percentile of the paired photo. In Fig.~\ref{fig_partial}, the high correlation demonstrates that the partial sketch with a higher scalar score by the critic network tends to have a higher average ranking percentile (ARP), while those with a lesser score result in lower ARP. Quantitatively, the top@5 accuracy for partial sketches is $80.1\%$, which have a higher predicted scalar score than a threshold of $\frac{1}{5}$. This validates the potential of our critic network in quantifying if a partial sketch is sufficient for retrieval. Suppose we repeat the same with the negative of the classification loss as a reward for a pre-trained classification network. In that case as well, we observe a similar consistent behaviour for partial sketch classification, indicating our approach to be generic for various sketch-related downstream tasks.\textbf{}
 
\vspace{-0.2cm}
 
\keypoint{Data Augmentation:} Our elementary study reveals that there exists multiple possible subsets which can retrieve the paired photo faithfully. In particular, we use our policy network to get stroke wise importance measure using $p(a_i|s_i)$ towards the retrieval objectives. Through categorical sampling of $p(a_i|s_i)$, we can create multiple augmented versions of the same sketch to increase the training data size. To validate this, we compute the performance of baseline retrieval model at varying training data size with our sketch augmentation strategy in Fig.~\ref{fig_partial}. While accuracy remains marginally better towards the high data regime, stroke subset selection based strategy excels the standard supervised counter-part by a significant margin, thus proving the efficacy of our smart data-augmentation approach.

\begin{figure}[!hbt]
\includegraphics[width=1.0\columnwidth]{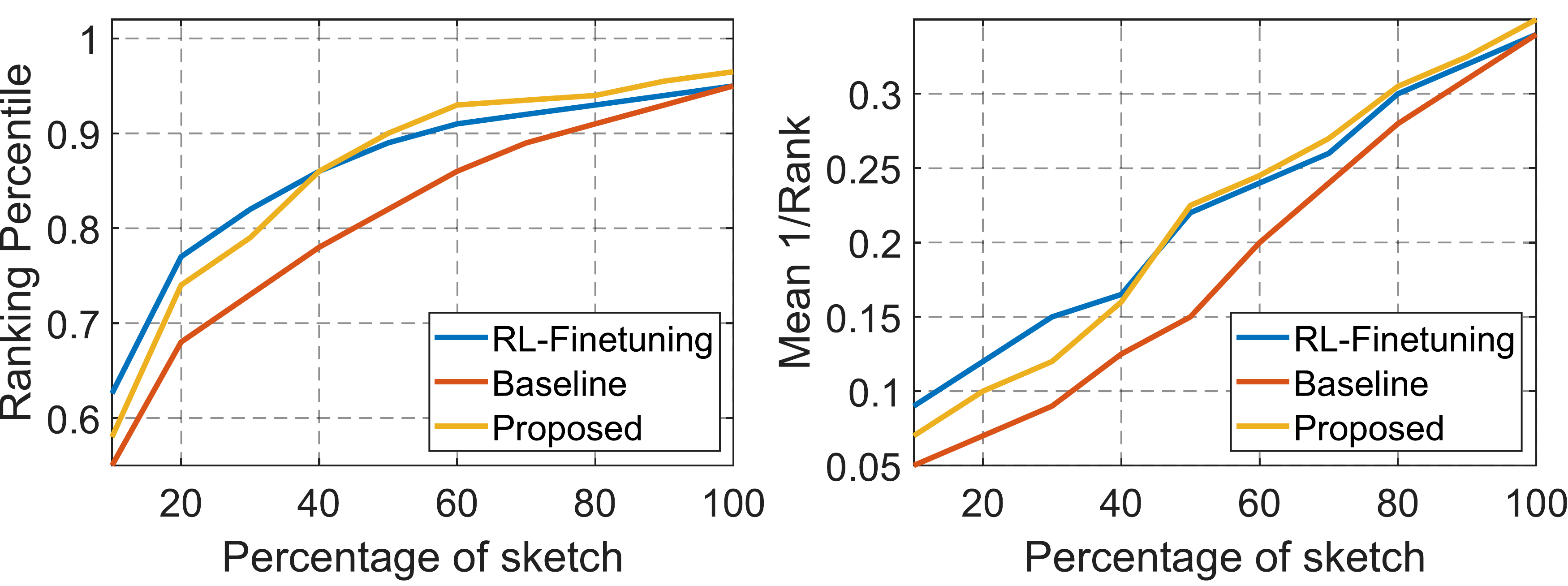}
\vspace{-0.75cm}
\caption{Comparative results under on-the-fly setup (Shoe-V2), visualised through percentage of sketch. Higher area under the plots indicates better early retrieval performance.}
\label{fig_onthefly}
\end{figure}

\keypoint{On-the-Fly Retrieval:} {Training a model with partial sketches generated by \emph{random stroke-dropping} gives rise to noisy gradient, and thus this naive baseline falls short compared to RL-based fine-tuning that consider the complete sketch drawing episode for training.} In lieu of RL-based fine-tuning \cite{bhunia2020sketch}, we train an on-the-fly retrieval model from meaningful (holds ability to retrieve) partial sketches augmented through our critic network that have a higher scalar score than $\frac{1}{20}$. While training a continuous RL pipeline  \cite{bhunia2020sketch} is unstable and time-consuming, we achieve a competitive on-the-fly \emph{r@A(r@B)} performance  of $85.78$($21.1$) with \emph{basic} triplet-loss based model trained with \emph{smartly} augmented partial sketches compared to $85.38$ ($21.24$) as claimed in \cite{bhunia2020sketch} on ShoeV2. From Fig.~\ref{fig_onthefly}, we can see that at very early few instances, RL-Based fine-tuning \cite{bhunia2020sketch} performs better, while ours achieve a significantly better performance as the drawing episode proceeds towards completion. While early sketch drawing episode is too coarse that hardly it can retrieve, through modelling the retrieval ability (with threshold of $\frac{1}{10}$) of partial sketches, we can reduce the number of time we need to feed the rasterized sketch by $42.2\%$ with very little drop in performance (r@A(r@B): $85.07$ ($20.98$)). Thus modelling partial sketches lead to significant computational edge under on-the-fly setting. 

\begin{figure}[!hbt]
\centering
\includegraphics[width=1\linewidth]{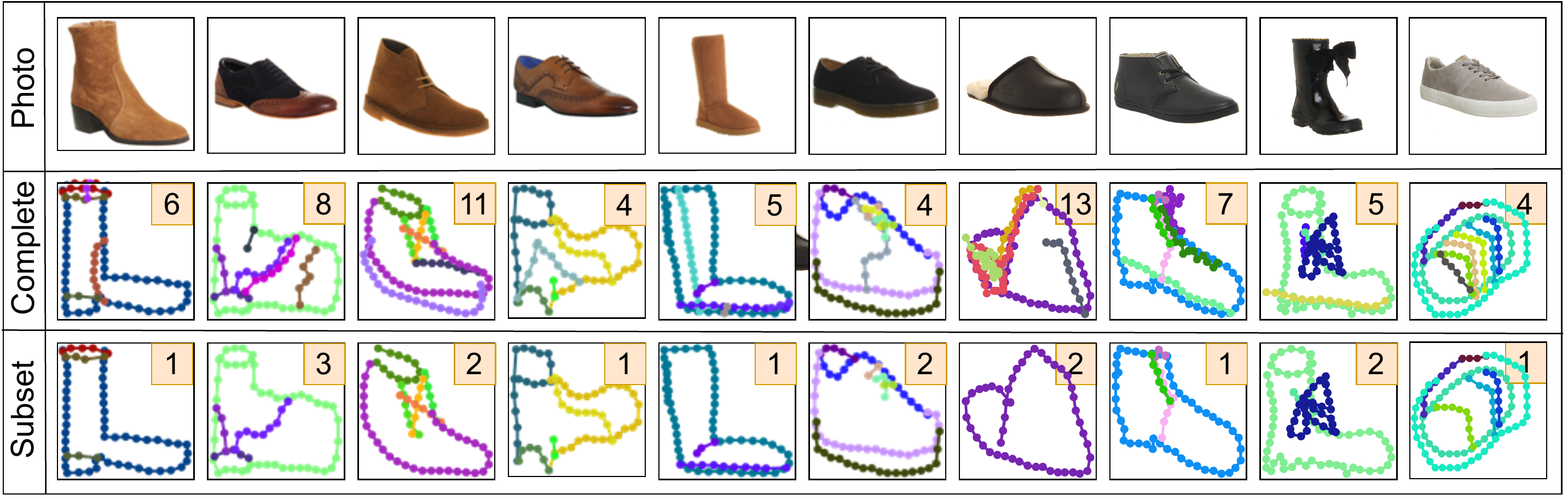}
\vspace{-0.75cm}
\caption{Examples showing selected subset performing better (rank in box) than complete sketch from ShoeV2.}
\label{fig6}
\end{figure}

\begin{figure}[!hbt]
\centering
\includegraphics[ width=1\linewidth]{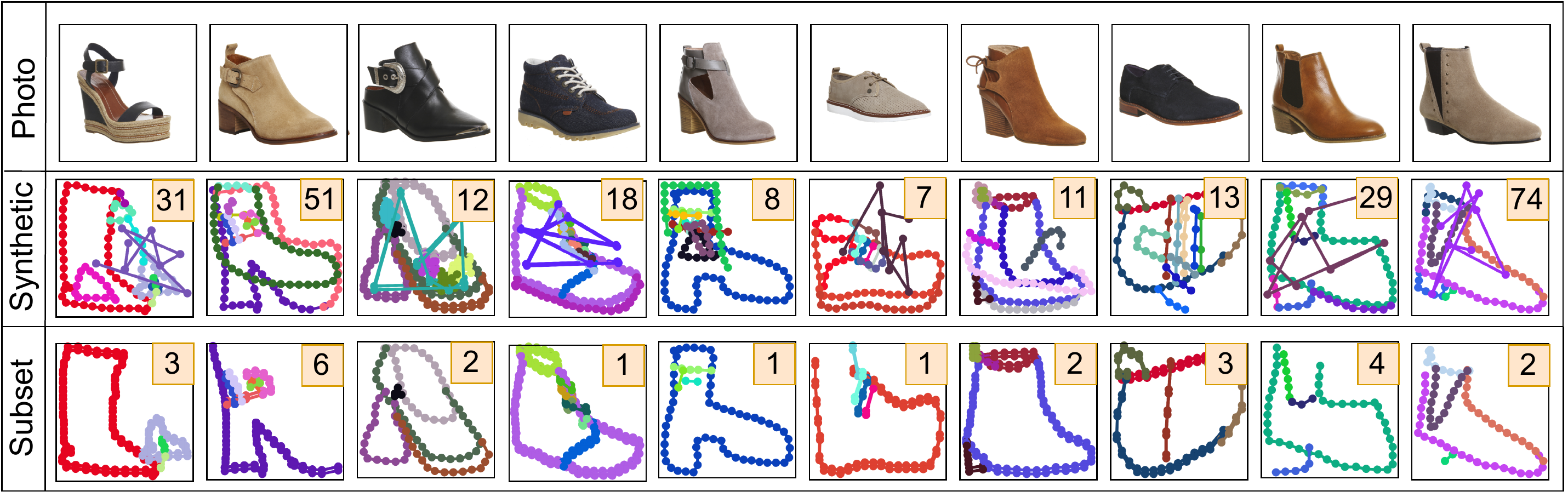}
\vspace{-0.75cm}
\caption{Examples showing ability to perform (rank in box) under \emph{synthetic} noisy sketch input on ShoeV2.}
\label{fig7}
\vspace{0.01cm}
\end{figure}

\keypoint{Resistance to Noisy Stroke:} The significance of stroke subset selector is quantitatively shown in Table~\ref{tab:my_label2234}.  While it validates our potential under inherent low-magnitude noise existed in the dataset (shown in Fig.~\ref{fig6}), we further aim to see how our method works on extreme noisy situation.  In particular, we augment the training sketches by synthetic noisy patches, and train our subset selector with a pre-trained retrieval model. During testing, we synthetically add noisy strokes \cite{liu2020unsupervised}, and pass it through stroke-subset selector (pre-processing module) before feeding it to the retrieval model. While excluding the selector, the top@1 (top@5) drops to $13.4\%$($44.9\%$) in presence of synthetic noises, our stroke subset selector can improve them to $37.2\%$($68.2\%$) by eliminating the synthetic noisy strokes (see Fig.~\ref{fig7}).

\vspace{-0.2cm}
 
\keypoint{Ablation on Design:} (i)  Instead of designing the stroke subset selector through hierarchical LSTM, another straight forward way is to use one layer bidirectional LSTM, where every coordinate point is being fed to each time step. However, the top@1(top@5) lags behind by $4.9\%$($6.7\%$) than ours, which verifies the necessity of hierarchical modelling of sketch vectors to consider the compositional relationship in our problem. Replacing LSTM by Transformer leads to no meaningful improvement in our case.  (ii) Being a pre-processing step, we compare the extra time required for selecting the optimal stroke set. In particular, it adds extra {$22.4\%$} multiply-add operations and $18.3\%$ extra CPU time compared standard baseline FG-SBIR. 
(iii) Compared to different RL methods \cite{schulman2017proximal},  we get best results with PPO actor-critic version with clipped surrogate objective that beats its actor-only alternative by $1.7\%$ top@1 accuracy(ShoeV2). Importantly, training with critic network leads to one important byproduct of modelling retrieval ability of partial sketches. (iv) Exploring different possible reward functions, we conclude that combining rewards from both ranking and feature embedding space through triplet loss gives most optimum performance than ranking only counterpart by extra $1.2\%$ top@1 accuracy (ShoeV2). Please refer to supplementary for more details.

\vspace{-0.1cm} 
\section{Conclusion}
\vspace{-0.2cm} 
\label{chapter:NT_FGSBIR:Conclusion}
In this chapter, we tackle the ``fear to sketch'' issue by proposing an intelligent stroke subset selector that automatically selects the most representative stroke subset from the entire query stroke set. Our stroke subset selector can detect and eliminate irrelevant (noisy) strokes, thus boosting performance of any off-the-shelf FG-SBIR framework. To this end, we designed an RL-based framework, which learns to form an optimal stroke subset by interacting with a pre-trained FG-SBIR model. We also show how the proposed selector can augment other sketch applications in a plug-and-play manner.



\aneeshan{Addressing the main concerns for a practical FG-SBIR paradigm, we now look towards the second main theme of this thesis --  learning powerful representations via self-supervised learning from unlabelled data, thus increasing its scope of application.}
This circles us back to one of the most general yet fundamental of challenges -- scarcity of sketch-data. Sketch datasets are limited by a noticeable margin compared to large-scale image datasets. To alleviate the data annotation bottleneck, many unsupervised/self-supervised methods propose to pre-train a good feature representation from large scale unlabelled data.  Although several self-supervised tasks are commonly used for such training, it is non-trivial in context of visual understanding of sketches, as they are  largely different from the pixel-perfect depiction of real photos. This pops the question, how to design a technique \textit{specific to sketches} that shall help create a good visual understanding of sketch in an unsupervised manner. Consequently, we are lead to explore the ways a sketch can be represented -- as a raster image, and as a temporal sequence of points. Exploiting this dual representation of sketches, we thus design a sketch-specific self-supervised task to facilitate unsupervised representation learning that aim to benefit various downstream sketch-based visual understanding tasks, in the following chapter.

\chapter{Self-Supervised Learning on Sketches}
    \label{chapter:Self}
    \minitoc
    \newpage
    
    \section{Overview}
    
\begin{figure}[!hbt] 
\begin{center}
  \includegraphics[width=0.7\linewidth]{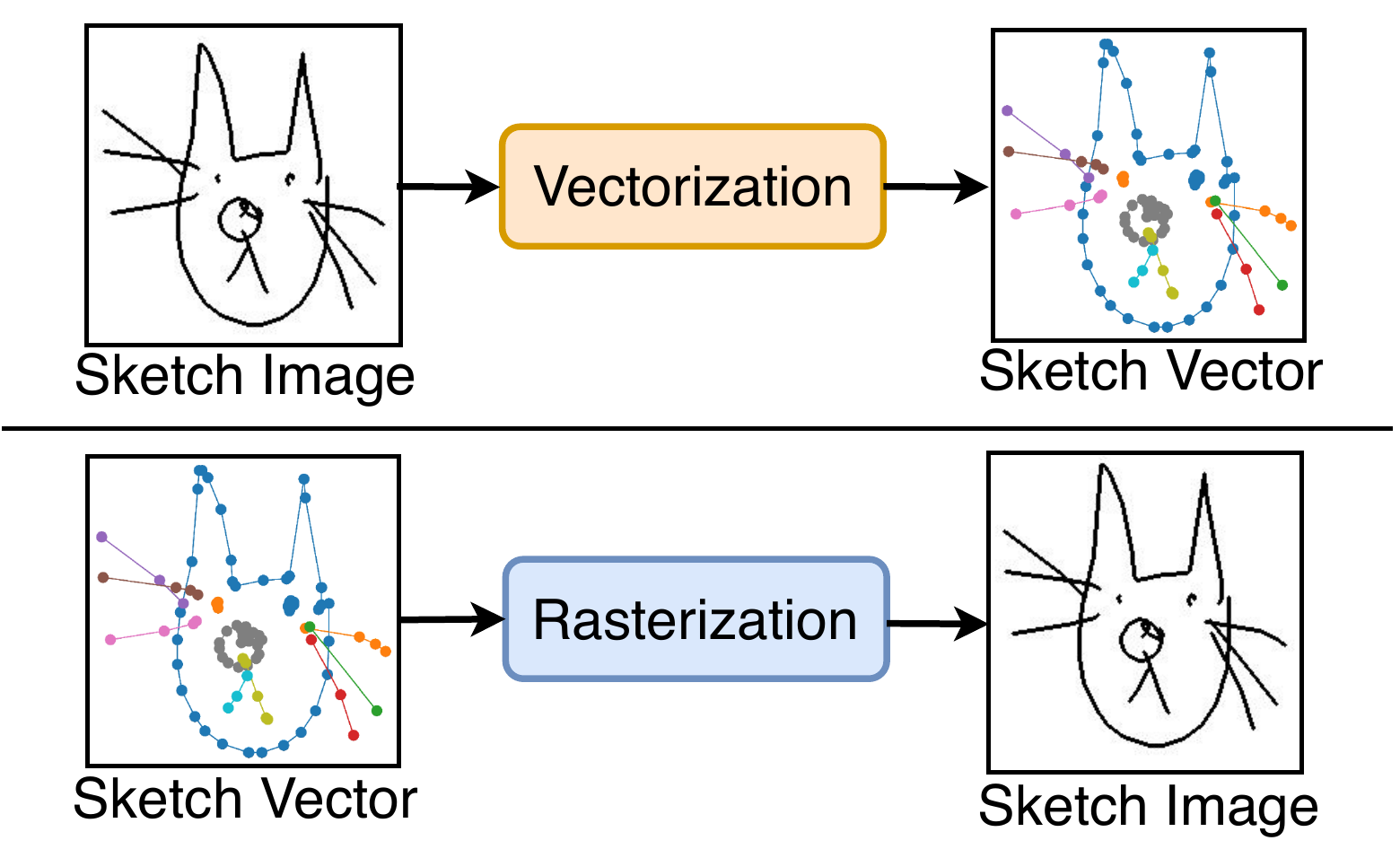}
\end{center}
\vspace{-0.8cm}
  \caption{Schematic of our proposed self-supervised method for sketches. Vectorization drives representation learning for sketch images; rasterization is the pre-text task for sketch vectors.}
\label{fig:Self_fig1.pdf}
\end{figure}

 
\aneeshan{Following our second theme, we begin our fourth chapter by addressing factors specific to sketches, that would benefit in making sketch applicable to a more generalised scenario of downstream vision tasks. } 
In this chapter, we propose a self-supervised method for a class of visual data that is distinctively different than photos: sketches \cite{yu2016sketchAnet, sangkloy2016sketchy} and handwriting \cite {poznanski2016cnn} images.  Although sketch and handwriting have been studied as two separate topics by different communities, there exists an underlying similarity in how they are captured and represented. More specifically, they are both recorded as the user's pen tip follows a trajectory on the canvas, and rendered as sparse black and white lines in image space. Both are abstract, in the sense that the same object or grapheme can be drawn in many possible ways \cite{song2018learning, ha2017neural}; while sketch in particular poses the challenge of variable levels of detail~\cite{sain2020cross} depicted.
Both sketch \cite{xu2018sketchmate} and handwriting \cite{graves2008novel} can be represented in rasterized pixel space, or as a temporal point sequence \cite{ha2017neural}. While each modality has its own benefits, we propose a novel self-supervised task that takes advantage of this dual image/vector space representation. In particular, we use cross modal translation between image and vector space as a self-supervised task to improve downstream performance using either representation (Figure~\ref{fig:Self_fig1.pdf}).
 
Most existing self-supervised methods are defined for single data modalities. Existing methods for images \cite{chen2020simple, he2020momentum, gidaris2018unsupervised} or videos \cite{korbar2018cooperative} are designed for pixel perfect renderings of scenes or objects, and as such are not suited for sparse black and white handwritten images. For example colorization \cite{zhang2016colorful} and super-resolution \cite{ledig2017photo} pre-texts, and augmentation strategies such as color distortion, brightness, and hue adjustment used by state of the art contrastive methods \cite{he2020momentum, chen2020simple, grill2020bootstrap} --  are not directly applicable to line drawings. For vector sequences, self-supervised methods typically addressed at speech such as Contrastive Predictive Coding (CPC) \cite{henaff2019data} could be used off-the-shelf but do not explicitly handle the stroke-by-stroke nature of handwriting. \cut{On the other hand}Conversely, BERT-like pre-training strategies have had some success with vector-modality sketches \cite{sketchbert2020} but cannot be applied to image-modality sketches. In contrast, our framework can be used to learn a powerful representation for both image and vector domain sketch analysis tasks.
Although a multi-view extension of contrastive learning for self-supervision~\cite{tian2019contrastiveCMC} has been attempted,  we show empirically that our cross-view rasterization/ vectorization synthesis approach provides a superior self-supervision strategy.

In summary, we design a novel self-supervised framework that exploits the dual raster/vector sequence nature of sketches and handwritten data through cross-modal translation (Figure~\ref{fig:Self_fig1.pdf}). Our cross-modal framework is simple and easy to implement from off-the-shelf components. Nevertheless it learns powerful representations for both raster and vector represented downstream sketch and handwriting analysis tasks. Empirically, our framework surpasses state of the art self-supervised methods and approaches and sometimes surpasses the fully supervised alternatives. 

\section{Methodology}
\label{chapter:sketch2vec:Methodology}

\subsection{Overview}
\keypoint{Overview:} Our objective is to design a self-supervised learning method that can be applied over both rasterized image and vector representation of any hand drawn data (e.g., sketch or handwriting); and furthermore it should exploit this complementary information for self-supervised learning. Towards this objective, we pose the feature learning task as a cross-modal translation between image and vector space using state-of-the-art encoder-decoder architectures. In other words, the training objective is to learn a latent space for the source modality from which the corresponding sample in the target modality is predictable. Once the cross-modal translation model is trained, we can remove the decoder and use the encoder as a feature extractor for source modality data. For instance, learning to translate from image space to vector space, we obtain an encoder that can embed raster encoding of hand drawn images into a meaningful latent representation, and vice-versa. 

Touch-screen devices and stylus-pens give us easy access to hand drawn data represented in both modalities simultaneously. Our training dataset consists of N  samples $\{I_i, V_i\}_{i=1}^{N}$, where  $I \in \mathcal{I}$ and $V \in \mathcal{V}$ are rasterized image and vector representation respectively. In particular, $I$ is a spatially extended image of size $\mathbb{R}^{H \times W \times3}$, and $V$ is a sequence of pen states $(v_1, v_2, \cdots, v_T)$, where $T$ is the length of the sequence. In order to learn feature representation on image space, we learn a \emph{vectorization} operation $\mathrm{\mathcal{I} \mapsto \mathcal{V}}$. Conversely, a \emph{rasterization} operation $\mathrm{\mathcal{V} \mapsto \mathcal{I}}$ is trained to provide a vector space representation. It is important to note that we do not use any category-label for sketch data or character/word annotation for handwritten data in our feature learning process. Thus, it can be trained in a class agnostic manner without any manual labels, satisfying the criteria of self-supervised learning. 

\vspace{0.5cm}
\begin{figure}[!hbt]
\begin{center}
\includegraphics[width=\linewidth]{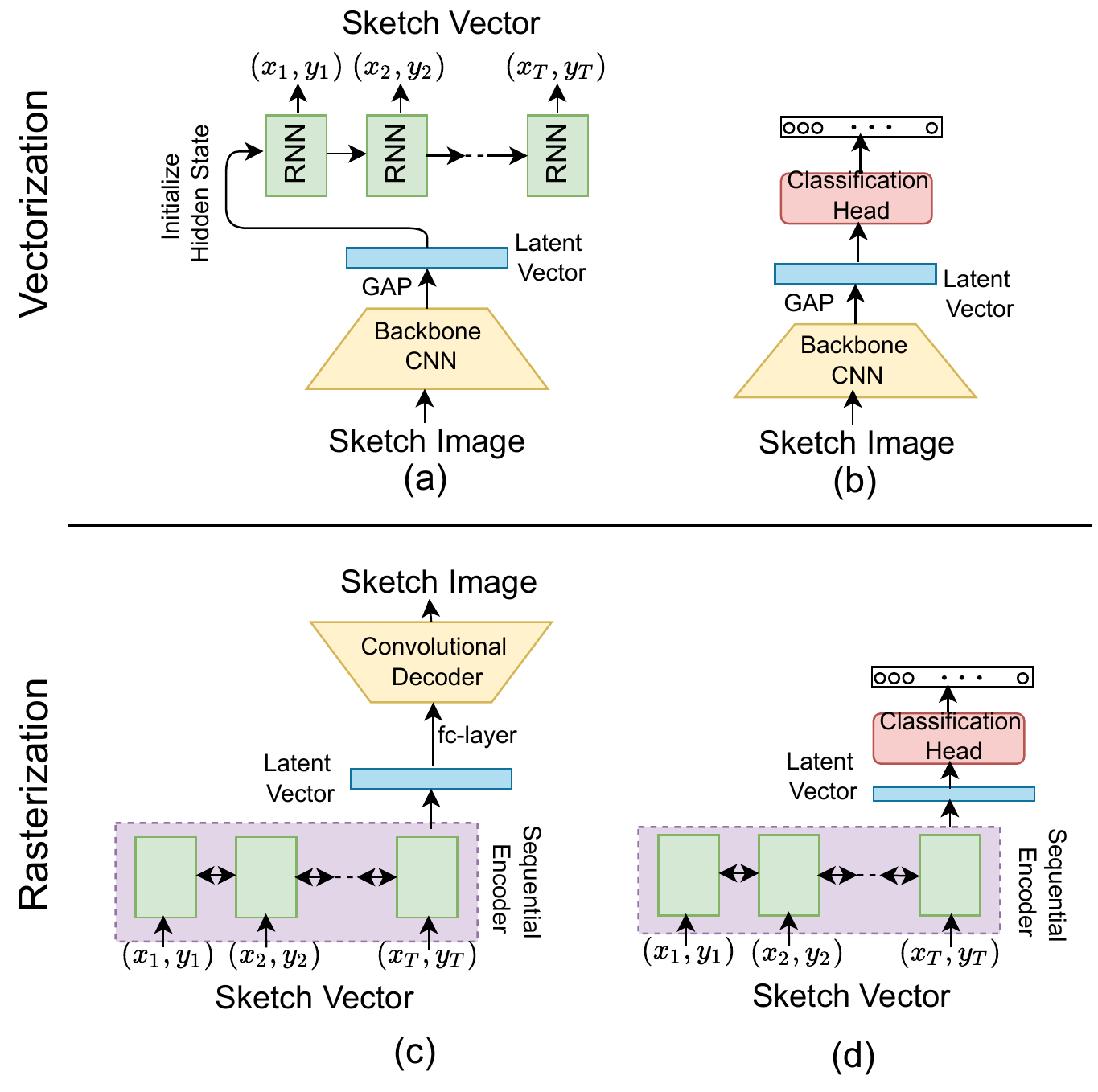}
\vspace{-1cm}
  \caption{Illustration of the architecture used for our self-supervised task for sketches and handwritten data (a,c), and how it can subsequently be adopted for downstream tasks (b,d). Vectorization involves translating sketch image to sketch vector (a), and the convolutional encoder used in the vectorization process acts as a feature extractor over sketch images for downstream tasks (b). On the other side, rasterization converts sketch vector to sketch image (c), and provides an encoding for vector-based recognition tasks downstream (d).}
\end{center}
\label{fig:Fig2}
\vspace{0.2cm}
\end{figure}

\subsection{Model Architecture}
For cross-modal translation, encoder $E(\cdot)$ embeds the data from source modality into a latent representation, and decoder $D(\cdot)$ reconstructs the target modality given the latent vector. $E(\cdot)$ and $D(\cdot)$ will be designed differently for the different source and target modalities. While rasterized image space is represented by a three-channel RGB image $\mathbb{R}^{H \times W \times 3}$, we use five-element vector $v_t=(x_t, y_t, q_t^1, q_t^2, q_t^3)$ to represent pen states in stroke-level modelling. In particular, $(x_t, y_t)$ is absolute coordinate value in a normalised $H \times W$ canvas, while the last  three  elements  represent  binary  one-hot vector \cite{ha2017neural} of  three  pen-state  situations:   pen touching  the  paper,  pen  being  lifted and end  of drawing. Thus, the size of vector representation is $V\in\mathbb{R}^{T\times5}$, where T is the sequence length. 

\subsection{Vectorization} For translating an image to its sequential point coordinate equivalent, image encoder ${E_I(\cdot)}$ can be any state-of-the-art convolutional neural network \cite{kolesnikov2019revisiting} such as ResNet. To predict the sequential point coordinates, decoder $D_V(\cdot)$ could be any sequential network, e.g. RNN. In particular, given an image $I$, let the extracted convolutional feature map be $F = E_I(I) \in \mathbb{R}^{h \times w \times d}$, where $h$, $w$ and $d$ signify height, width and number of channels respectively. Applying global max pooling (GAP) to $F$ and flattening, we obtain a vector $l_I$ of size $\mathbb{R}^d$, which will be used as the representation for input image $I$ once the encoder-decoder model is trained. Next, a linear-embedding layer is used to obtain the initial hidden state of the decoder RNN as follows: $h_0 = W_hl_I+ b_h$.  The hidden state $h_t$ of decoder RNN  is updated as follows: ${h_t = RNN(h_{t-1} ; [l_I, P_{t-1}] )}$, where  $P_{t-1}$ is the last predicted point and $[\cdot]$ stands for a concatenation operation. Thereafter, a fully-connected layer is used to predict five-element vector at each time step as: $P_t = W_yh_t+ b_y$, where $P_t = (x_t, y_t, q^1_t, q^2_t, q^3_t)$ is of size $\mathbb{R}^{2+3}$, whose first two logits represent absolute coordinate $(x,y)$ and the latter three  for pen's state position $(q^1, q^2, q^3)$. We use simple mean-square error and categorical cross-entropy losses to train the absolute coordinate and pen state prediction (softmax normalised) respectively. Thus, $(\hat{x}_t, \hat{y}_t, \hat{q}^1_t, \hat{q}^2_t, \hat{q}^3_t)$ being the ground-truth coordinate at $t$-th step, the training loss is: 
\vspace{-0.3 cm}
\begin{equation}
\begin{aligned}
  L_{I \rightarrow V} &= \frac{1}{T}\sum_{t=1}^{T} \left \| \hat{x}_t - {x}_t \right \|_2 +  \left \| \hat{y}_t - {y}_t\right \|_2  \\
    &- \frac{1}{T}\sum_{t=1}^{T}\sum_{i=1}^{3}\hat{q^i_t} \log \Big(\frac{\exp({q^i_t})}{\sum_{j=1}^{3}\exp({q^j_t})}\Big)
\end{aligned}
\end{equation}


\subsection{Rasterization} To translate a sequence of point coordinates $V$ to its equivalent image representation $I$, any sequential network such as RNN \cite{collomosse2019livesketch, graves2008novel} or Transformer \cite{sketchbert2020}, could be used as the encoder $E_V(\cdot)$, and we experiment with both. For RNN-like architectures \cite{ha2017neural}, we feed the five elements vector $v_t$ at every time step, and take the hidden state of final time step as the encoded latent representation. For Transformer like encoders, we take input via a trainable linear layer to convert each five element vector to the Transformer's model dimension. Additionally, we prepend a learnable embedding to the input sequence, similar to BERT's \emph{class token} \cite{devlin2018bert}, whose state at the output acts as the encoded latent representation.  Finally, the encoder latent representation $l_V\in\mathbb{R}^d$ is fed via a fully-connected layer to a standard convolutional decoder $D_I(\cdot)$. $D_I(\cdot)$ consists of series of fractionally-strided convolutional layers \cite{isola2017image} to up-sample the spatial size to $H \times W$ at the output. \cut{Given the corresponding rasterized image $\hat{I}$,  and the generated output from the convolutional decoder be  $I$, we use mean-square error as the training objective:} Given the vector $V$ and raster $I$ data pairs, we  use mean-square error as the training objective:

\vspace{-0.2cm}
\begin{equation}
L_{V \rightarrow I} = - \mathbb{E}_{(I, V) \sim \mathcal{(I,V)}}  \left \| I -  D_I(E_V(V))\right \|_2 
\end{equation}
\vspace{-0.3cm}

We remark that due to the well known regression to mean problem \cite{mathieu2015deep}, the generated images are indeed blurry. Adding an adversarial loss \cite{isola2017image} does not give any improvement in our representation learning task, and sometimes leads to worse results due to mode collapse issue in adversarial learning. However, synthesising realistic images is not our goal in this work. Rather, it is a pretext task for learning latent representations for vector sequence inputs. 

\vspace{-0.2cm}
\section{Application of Learned Representation}
We apply our self-supervised learning method on both sketch \cite{ha2017neural, eitz2012humans} and handwriting data \cite{IamDataset}, as both can be represented in image and vector space. 
\vspace{-0.2cm}
\subsection{Sketch Analysis:} We use sketch-recognition \cite{yu2016sketchAnet} and sketch-retrieval \cite{xu2018sketchmate, sketchbert2020} as downstream tasks to evaluate the quality of learned latent representation obtained by our self-supervised pre-training. For both classification and retrieval, we evaluate performance with both sketch image and sketch vector representations using vectorization and rasterization as pre-training task respectively. For classification, we simply apply a fully-connected layer with softmax on the extracted latent representation from pre-trained encoder. For retrieval, we could use the latent representation itself. However we find it helpful to project the latent feature through a fully connected layer into $256$ dimensional embedding space, and optimise the model through triplet loss \cite{dey2019doodle, bhunia2020sketch}. Along with triplet loss, that minimises intra-class distance while maximising inter-class distance, we also apply a classification loss through a linear layer \cite{collomosse2019livesketch} to further aid the retrieval learning framework.  

\subsection{Handwriting Recognition:} 
As handwriting recognition \cite{bhunia2019handwriting} is a (character) sequence task oriented at decoding a whole word from an image\cut{ (`sketching' the word during pre-training, and recognising the letters as the downstream task)}, we use a slightly modified vectorization encoder $E_I(\cdot)$ for offline/image recognition compared to the sketch tasks. Following \cite{shi2018aster}, the (word) image encoder extends a conventional ResNet architecture with a 2-layer BLSTM image feature encoder before producing a final state that provides the latent vector for the input image. This feature is then fed to a sequential decoder to `sketch' the word during cross-modal self-supervised pre-training, and to a recognition model to recognise the word in the downstream task. In contrast, the rasterization encoder $E_V(\cdot)$ is defined similarly as for the sketch tasks. For the downstream task, after encoding either vector and raster inputs, we follow \cite{oord2018representation} in using an attentional decoder \cite{shi2018aster, zhang2019sequence} to recognize the word by predicting the characters sequentially. This decoder module consists of a BLSTM layer followed by a GRU layer that predicts the characters.

\section{Experiments}
\subsection{Dataset} 
For sketches, we use the standard QuickDraw \cite{ha2017neural} and TU-Berlin \cite{eitz2012humans} datasets for evaluation as they contain both raster and vector image representations. QuickDraw contains 50 million sketches from 345 classes. We use the split from \cite{ha2017neural} where each class has 70K training samples, 2.5K validation, and 2.5K test samples. Meanwhile, TU-Berlin comprises of 250 object categories with 80 sketches in each category. We apply the Ramer-Douglas-Peucker (RDP) algorithm to simplify the sketches. For handwriting we use IAM offline and online  datasets \cite{IamDataset}: The offline set  contains 115,320 word images, while the online set contains point coordinate representation of $13,049$ lines of  handwriting.  We pre-process line-level online data to segment it into $70,648$ valid words, and use synthetic rasterization to create training data for vector and raster views.

\subsection{Implementation Details} 
We implemented our framework in PyTorch \cite{paszke2017automatic} and conducted experiments on a 11 GB Nvidia RTX 2080-Ti GPU. 
While a GRU decoder of hidden state size $512$ is used in all the vectorization process, we use convolutional decoder from \cite{isola2017image}  in the rasterization process. Following a recent self-supervised study analysis \cite{kolesnikov2019revisiting, grill2020bootstrap}, we use ResNet50 as the CNN encode images, unless otherwise mentioned. For vector sketch recognition, we use a  Transformer \cite{vaswani2017attention} encoder with 8 layers, hidden state size 768, MLP size 2048, and 12 heads.  For offline handwritten images, the encoder architecture is taken from \cite{shi2018aster} and comprises a ResNet like convolutional architecture followed by a 2 layers BLSTM. For online handwriting, we feed 5-element vectors at every time step of a 4-layers stacked BLSTM \cite{carbune2020fast} with hidden state size 512.  We use Adam optimiser with learning 0.0001 and batch size of 64 for all experiments.   

\subsection{Evaluation Protocol} 
For sketch recognition, Top-1 and Top-5 accuracy is used, and for category level sketch-retrieval, we employ Acc@top1 and mAP@top10 as the evaluation metric. For handwriting recognition, we use Word Recognition Accuracy following \cite{bhunia2019handwriting}.

\subsection{Competitor} 

We compare with existing self-supervised learning methods that involve pre-text task like context prediction \cite{doersch2015unsupervised},  Auto-Encoding \cite{kingma2013auto}, jigsaw solving \cite{noroozi2016unsupervised}, rotation prediction \cite{gidaris2018unsupervised}. Clustering based representation learning Deep Cluster  \cite{caron2018deep} is also validated on sketch datasets. Furthermore, we compare with three state-of-the-art contrastive learning based self-supervised learning methods, namely, SimCLR \cite{chen2020simple}, MoCo \cite{he2020momentum}, and BYOL \cite{grill2020bootstrap}. While these self-supervised learning methods are oriented at RGB photos rather than sketch or handwritten data, we also compare with Sketch-Bert \cite{sketchbert2020} as the only work employing self-supervised learning on vector sketches. We note that self-supervised methods designed for image data can not be used off-the-shelf for vector sketches. The only exception is Contrastive Predictive Coding (CPC) \cite{oord2018representation}  which has been used to handle both images and sequential data (e.g. speech signals). Finally, we compare with a state of the art multi-modal self-supervised method Contrastive Multi-view Coding (CMC) \cite{tian2019contrastiveCMC}, which performs contrastive learning of (mis)matching instances across modalities. Here, we use the same encoder for raster sketch and vector sketch like ours for a fair comparison. 

\setlength{\tabcolsep}{5pt}
\begin{table}[!hbt]
    \centering
    \scriptsize
    \caption{Linear model evaluation of fixed pre-trained features. ResNet50 for image space and Transformer for vector space inputs.}
    \vspace{-0.3cm}
    \begin{tabular}{r|cccccccc}
        \hline \hline
         & \multicolumn{8}{c}{Recognition} \\
        \cline{2-9}
         & \multicolumn{4}{c}{Image Space} & \multicolumn{4}{c}{Vector Space}  \\
        \cline{2-9}
         & \multicolumn{2}{c}{QuickDraw} & \multicolumn{2}{c}{TU-Berlin} & \multicolumn{2}{c}{QuickDraw} \\
         & Top-1 & Top-5 & Top-1 & Top-5 & Top-1 & Top-5 & Top-1 & Top-5 \\
        \hline
        Supervised & 76.1\% & 91.3\% & 78.6\% & 90.1\% & 73.5\% & 90.1\% & 62.9\% & 80.7\%  \\
 
        Random & 15.5\% & 26.2\% & 18.4\% & 29.3\% & 12.7\% & 23.6\% & 9.6\% & 19.4\%  \\
        \hdashline 
        Context \cite{doersch2015unsupervised} &  44.6\% & 69.2\% & 43.3\% & 67.5\% & - & - & - & -  \\
        Auto-Encoder \cite{kingma2013auto} &  26.4\% & 48.1\% & 22.6\% & 47.5\% & - & - & - & -   \\
        Jigsaw \cite{noroozi2016unsupervised} &  46.9\% & 71.5\% & 45.7\% & 69.8\% & - & - & - & -  \\
       Rotation \cite{gidaris2018unsupervised} &  53.5\% & 78.7\% & 51.2\% & 77.1\% & - & - & - & -  \\
     Deep Cluster \cite{caron2018deep} & 39.4\% & 62.7\% & 38.7\% & 60.2\% & - & - & - & -  \\
    MoCo \cite{he2020momentum} &  65.7\% & 85.1\% & 64.3\% & 82.8\% & - & - & - & -   \\
   SimCLR \cite{chen2020simple} &  65.5\% & 85.1\% & 64.3\% & 82.9\% & - & - & - & -  \\
    BYOL \cite{grill2020bootstrap} &  66.8\% & 85.8\% & 65.7\% & 83.7\% & - & - & - & -  \\
Sketch-BERT \cite{sketchbert2020} & - & - & - & - & 65.6\% & 85.3\% & 52.9\% & 78.1\%   \\
CMC \cite{tian2019contrastiveCMC}  & 63.6\% & 83.9\% & 61.7\% & 81.3\% & 61.2\% & 81.5\% & 51.4\% & 77.5\%  \\
CPC \cite{oord2018representation}  & 54.3\% & 79.0\% & 52.9\% & 77.9\% & 59.3\% & 81.3\% & 50.5\% & 76.6\%  \\
Ours-(L)  & 71.9\% & 89.7\% & 70.6\% & 85.9\% & 67.2\% & 86.5\% & 55.6\% & 79.4\%\\
        \hline
    \end{tabular}
    
\vspace{0.3cm}
   \begin{tabular}{r|cccccccc}
        \hline \hline & \multicolumn{8}{c}{Retrieval} \\
        \cline{2-9}
           & \multicolumn{4}{c}{Image Space} & \multicolumn{4}{c}{Vector Space} \\
        \cline{2-9}
           & \multicolumn{2}{c}{QuickDraw} & \multicolumn{2}{c}{TU-Berlin} & \multicolumn{2}{c}{QuickDraw} & \multicolumn{2}{c}{TU-Berlin} \\
          & A@1 & m@10 & A@1 & m@10 & A@1 & m@10 & A@1 & m@10 \\
        \hline
        Supervised &  62.3\% & 69.4\% & 69.1\% & 74.7\% & 58.5\% & 77.1\% & 50.2\% & 67.4\% \\
 
        Random   & 10.6\% & 21.3\% & 13.4\% & 26.7\% & 9.8\% & 21.5\% & 9.2\% & 17.6\% \\
        \hdashline 
        Context \cite{doersch2015unsupervised}  & 30.7\% & 34.9\% & 28.4\% & 32.7\% & - & -& - & - \\
        Auto-Encoder \cite{kingma2013auto}  & 16.4\% & 24.4\% & 15.3\% & 20.4\% & - & -& - & - \\
        Jigsaw \cite{noroozi2016unsupervised}  & 31.6\% & 38.9\% & 30.6\% & 35.4\% & - & -& - & - \\
       Rotation \cite{gidaris2018unsupervised}  & 37.5\% & 45.1\% & 36.4\% & 41.8\% & - & -& - & - \\
     Deep Cluster \cite{caron2018deep} & 29.2\% & 36.8\% & 27.3\% & 31.9\% & - & -& - & - \\
    MoCo \cite{he2020momentum}  & 42.5\% & 46.8\% & 42.5\% & 46.9\% & - & -& - & - \\
   SimCLR \cite{chen2020simple}  & 43.3\% & 50.7\% & 41.5\% & 46.7\% & - & -& - & - \\
    BYOL \cite{grill2020bootstrap}  & 45.4\% & 52.5\% & 43.8\% & 49.1\% & - & -& - & - \\
Sketch-BERT \cite{sketchbert2020}  & 48.9\% & 68.1\% & 40.7\% & 58.8\% \\
CMC \cite{tian2019contrastiveCMC}   & 40.6\% & 45.8\% & 38.5\% & 43.3\% & 45.2\% & 66.7\% & 40.3\% & 58.2\% \\
CPC \cite{oord2018representation}    & 37.9\% & 43.1\% & 36.4\% & 40.9\% & 43.1\% & 63.6\% & 39.3\% & 57.9\% \\
Ours-(L)   & 52.3\% & 59.5\% & 47.7\% & 59.1\% & 49.5\% & 68.9\% & 42.1\% & 59.6\% \\
        \hline
    \end{tabular}    
    \label{tab:my_label1}
    \vspace{0.5cm}
\end{table}

\setlength{\tabcolsep}{5pt}
\begin{table}[!hbt]
    \centering
    \scriptsize
    \caption{Semi-supervised fine-tuning using $1\%$ and $10\%$ labelled training data on QuickDraw.}
    \vspace{-0.3cm}
    \begin{tabular}{r|cccccccc}
        \hline \hline
         & \multicolumn{8}{c}{Recognition} \\
        \cline{2-9}
         & \multicolumn{4}{c}{Image Space} & \multicolumn{4}{c}{Vector Space}  \\
        \cline{2-9}
           & \multicolumn{2}{c}{QuickDraw} & \multicolumn{2}{c}{TU-Berlin} & \multicolumn{2}{c}{QuickDraw} & \multicolumn{2}{c}{TU-Berlin} \\
         & Top-1 & Top-5 & Top-1 & Top-5 & Top-1 & Top-5 & Top-1 & Top-5 \\
        \hline
        Supervised & 25.1\% & 47.3\% & 55.4\% & 79.0\% & 17.3\% & 37.5\% & 43.9\% & 65.9\% \\
 
        Random &  33.9\% & 55.8\% & 56.8\% & 80.5\% & - & - & - & -  \\
        \hdashline 
        Context \cite{doersch2015unsupervised} &  44.6\% & 69.2\% & 43.3\% & 67.5\% & - & - & - & -  \\
        Auto-Encoder \cite{kingma2013auto} &  21.5\% & 40.7\% & 45.1\% & 70.6\% & - & - & - & - \\
        Jigsaw \cite{noroozi2016unsupervised} &  36.5\% & 57.4\% & 57.4\% & 80.3\% & - & - & - & - \\
       Rotation \cite{gidaris2018unsupervised}  &  38.8\% & 59.1\% & 59.6\% & 80.7\% & - & - & - & - \\
     Deep Cluster \cite{caron2018deep} & 32.2\% & 54.5\% & 54.7\% & 79.2\% & - & - & - & - \\
    MoCo \cite{he2020momentum} &  46.0\% & 70.5\% & 62.2\% & 83.7\% & - & - & - & - \\
   SimCLR \cite{chen2020simple} &  46.1\% & 70.5\% & 62.1\% & 83.6\% & - & - & - & - \\
    BYOL \cite{grill2020bootstrap} &  47.3\% & 72.0\% & 62.7\% & 84.1\% & - & - & - & - \\
Sketch-BERT \cite{sketchbert2020} & - & - & - & - & 45.1\% & 69.8\% & 62.4\% & 81.7\%  \\
CMC \cite{tian2019contrastiveCMC}  & 44.6\% & 68.2\% & 61.7\% & 82.7\% & 44.6\% & 68.4\% & 61.7\% & 81.6\% \\
CPC \cite{oord2018representation}  & 40.6\% & 65.7\% & 60.7\% & 81.9\% & 43.5\% & 67.7\% & 61.6\% & 81.7\%  \\
Ours-(L)  & 51.2\% & 76.4\% & 65.6\% & 85.2\% & 46.8\% & 70.9\% & 63.2\% & 83.9\% \\
        \hline
    \end{tabular}
    
\vspace{0.3cm}
   \begin{tabular}{r|cccccccc}
        \hline \hline & \multicolumn{8}{c}{Retrieval} \\
        \cline{2-9}
           & \multicolumn{4}{c}{Image Space} & \multicolumn{4}{c}{Vector Space} \\
        \cline{2-9}
           & \multicolumn{2}{c}{QuickDraw} & \multicolumn{2}{c}{TU-Berlin} & \multicolumn{2}{c}{QuickDraw} & \multicolumn{2}{c}{TU-Berlin} \\
        & A@1 & m@10 & A@1 & m@10 & A@1 & m@10 & A@1 & m@10 \\
        \hline
        Supervised &  13.4\% & 34.3\% & 43.9\% & 63.7\% & 9.1\% & 29.0\% & 41.0\% & 60.8\% \\
 
        Random   & 24.4\% & 30.5\% & 42.6\% & 48.4\% & - & -& - & - \\
        \hdashline 
        Context \cite{doersch2015unsupervised}  & 30.7\% & 34.9\% & 28.4\% & 32.7\% & - & -& - & - \\
        Auto-Encoder \cite{kingma2013auto}  & 15.2\% & 21.4\% & 32.7\% & 37.6\% & - & -& - & -  \\
        Jigsaw \cite{noroozi2016unsupervised}  & 27.7\% & 35.2\% & 44.7\% & 51.2\% & - & -& - & - \\
       Rotation \cite{gidaris2018unsupervised}  & 28.4\% & 35.2\% & 44.7\% & 51.8\% & - & -& - & - \\
     Deep Cluster \cite{caron2018deep} & 24.4\% & 31.2\% & 43.6\% & 47.7\% & - & -& - & - \\
    MoCo \cite{he2020momentum}  & 35.9\% & 43.1\% & 52.7\% & 57.4\% & - & -& - & - \\
   SimCLR \cite{chen2020simple}  & 35.1\% & 42.7\% & 52.3\% & 57.4\% & - & -& - & -  \\
    BYOL \cite{grill2020bootstrap}  & 36.5\% & 43.0\% & 52.8\% & 59.8\% & - & -& - & -  \\
Sketch-BERT \cite{sketchbert2020}  & - & - & - & -& 36.5\% & 60.0\% & 52.9\% & 72.9\% \\
CMC \cite{tian2019contrastiveCMC}  & 34.7\% & 41.9\% & 51.1\% & 57.4\% & 35.4\% &  58.1\% & 52.6\% & 72.8\%  \\
CPC \cite{oord2018representation} & 33.4\% & 40.1\% & 50.5\% & 57.7\% & 34.1\% & 56.6\% & 52.3\% & 72.8\%\\
Ours-(L)   & 38.6\% & 45.6\% & 60.4\% & 81.4\% & 37.1\% & 61.5\% & 53.2\% & 74.3\% \\
        \hline
    \end{tabular}    
    \label{tab:my_label2}
\vspace{0.2cm}
\end{table}

\setlength{\tabcolsep}{6pt}
\begin{table}[!htbp]
    \centering
    \footnotesize
    \caption{Accuracy on QuickDraw dataset with linear classifier trained on representation from various depth within the network.}
    \begin{tabular}{cccccc}
        \hline \hline
        Method & Block1 & Block2 & Block3 & Block4 & Pre-logits\\
        \hline
         Supervised & 7.0\% & 14.9\% & 35.6\% & 72.5\% & 76.1\% \\
         \hdashline
         Jigsaw \cite{noroozi2016unsupervised} & 4.2\% & 8.1\% & 26.8\% & 39.9\% & 46.9\% \\
         Rotation  \cite{gidaris2018unsupervised} & 5.2\% & 11.2\% & 27.5\% & 45.4\% & 53.5\% \\
         Deep Cluster \cite{caron2018deep}  & 4.1\% & 8.6\% & 19.6\% & 33.4\% & 39.4\% \\
       CMC \cite{tian2019contrastiveCMC}  & 6.1\% & 11.9\% & 29.7\% & 56.7\% & 63.6\% \\
         MoCo \cite{he2020momentum}  & 7.7\% & 13.5\% & 31.6\% & 60.3\% & 65.7\%  \\
         SimCLR \cite{chen2020simple} & 6.7\% & 12.6\% & 32.0\% & 59.5\% & 65.5\% \\
         BYOL  \cite{grill2020bootstrap} & 9.0\% & 15.1\% & 32.2\% & 61.2\% & 66.8\% \\
         Ours & 10.1 & 15.2\% & 34.4\% & 67.5\% & 71.9\% \\
        \hline
    \end{tabular}
    \label{tab:my_label3}
\end{table}

\subsection{Results on Sketch Representation Learning}\label{sec:perf} 
\keypoint{Sketch Recognition:}  Following the traditional protocol of evaluation for self-supervised learning \cite{chen2020simple, grill2020bootstrap}, we first evaluate our representations by training a linear classifier on the top of frozen representation. We report the recognition accuracy in  Table(left) \ref{tab:my_label1}. On QuickDraw, Top-1 accuracy of $71.9\%$ and $67.2\%$ is obtained for sketch images and sketch vectors respectively, approaching the supervised counterparts of $76.1\%$ and $73.5\%$. For TU-Berlin accuracies of $70.6\%$ and $55.6\%$ also approach the supervised figures $78.6\%$ and $62.9\%$. The gap with supervised method is larger for TU-Berlin dataset because of having less data compared to QuickDraw dataset. Interestingly, the performance over image level data is comparatively better than using sketch vectors.  

We next evaluate the semi-supervised setup, where we fine-tune the whole network using smaller subset of training data, $1\%$ and $10\%$. Our self-supervised methods helps to learn good initialization  such that in this low data regime, ours is significantly better than its supervised counter part as shown in Table \ref{tab:my_label2}. Finally, we evaluate the learned features from various depths of our convolutional encoder for sketch raster image classification in Table \ref{tab:my_label3}. Overall, our close competitors are contrastive learning based family of self-supervised methods, e.g. SimCLR, BYOL, MoCo. We attribute the superiority of our method over other self-supervised methods, on the sketch dataset, to the task-design that exploits the intrinsic dual representation of sketch data.

\keypoint{Sketch Retrieval:} For sketch retrieval, first we use the extracted latent feature from pre-trained self-supervised network for triplet metric learning of an additional linear embedding layer. From the retrieval performance in Table~\ref{tab:my_label1} (right) we see a relative performance between the methods that is similar to the previous sketch classification experiments. However, the retrieval performance using the fixed self-supervised latent feature is $9-10\%$ below the supervised version.  In the semi-supervised experiment, we fine-tune the complete model including  linear layer and the pre-trained feature extractor using $1\%$ and $10\%$ of the training data respectively.  Table \ref{tab:my_label2} shows that our self-supervised method has a clear edge over supervised counter part in this low data regime. Qualitative cross-modal generated and retrieved results are shown in Figure~\ref{fig:Graph1_Full} \& \ref{fig:Graph2_Full}, respectively.

\setlength{\tabcolsep}{6pt}
\begin{table}[!htbp]
 \scriptsize
    \centering
    \caption{Handwriting recognition using feature extracted from fixed pre-trained encoder.}
    \vspace{-0.2cm}
    \begin{tabular}{ccccc}
        \hline\hline
         & \multicolumn{2}{c}{Offline} & \multicolumn{2}{c}{Online} \\
        \cline{2-5}
         & Lexicon & No Lexicon &  Lexicon & No Lexicon  \\
        \hline
        Supervised \cite{shi2018aster} & 87.1\% & 81.5\% & 88.4\% & 82.8\% \\
        Random & 10.4\% & 6.3\% & 7.4\% & 4.9\% \\
            \hdashline
        CPC \cite{oord2018representation} & 72.2\% & 63.7\% & 71.5\% & 62.8\% \\
        Ours & 75.4\% & 68.6\% & 73.1\% & 66.9\% \\
        \hline
    \end{tabular}
     \label{tab:my_label4}
\end{table}

\setlength{\tabcolsep}{6pt}
\begin{table}[!htbp]
 \scriptsize
    \centering
    \caption{Handwriting recognition under semi-supervised setup.}
     \vspace{-0.2cm}
    \begin{tabular}{ccccc}
        \hline\hline
         & \multicolumn{2}{c}{Offline} & \multicolumn{2}{c}{Online} \\
        \cline{2-5}
         & 1\% Training & 10\% Training & 1\% Training & 10\% Training \\
        \hline
        Supervised \cite{shi2018aster} & 19.7\% & 40.6\% & 20.5\% & 42.4\% \\
        \hdashline
        CPC \cite{oord2018representation} & 29.1\% & 55.4\% & 27.8\% & 54.2\% \\
        Ours & 38.5\% & 59.2\% & 36.8\% & 56.7\% \\
        \hline
    \end{tabular}
    \label{tab:my_label5}
    \vspace{0.5cm}
\end{table}

\begin{figure*}[t]
        \centering
		\includegraphics[height=2.7cm, width=0.24\linewidth]{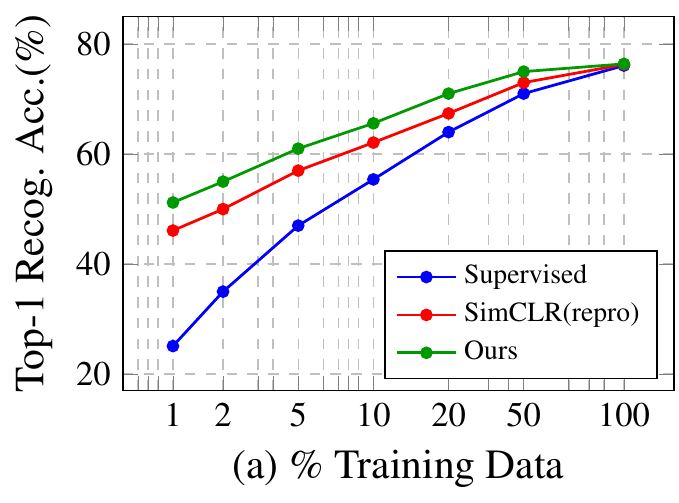}
 		\includegraphics[height=2.7cm, width=0.24\linewidth]{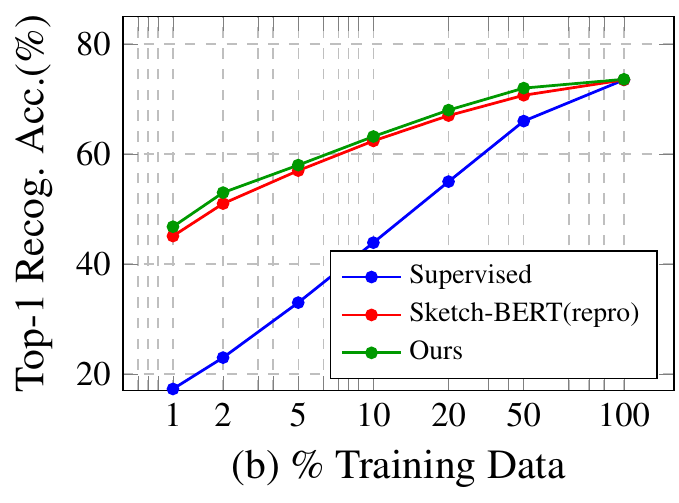}
		\includegraphics[height=2.7cm, width=0.24\linewidth]{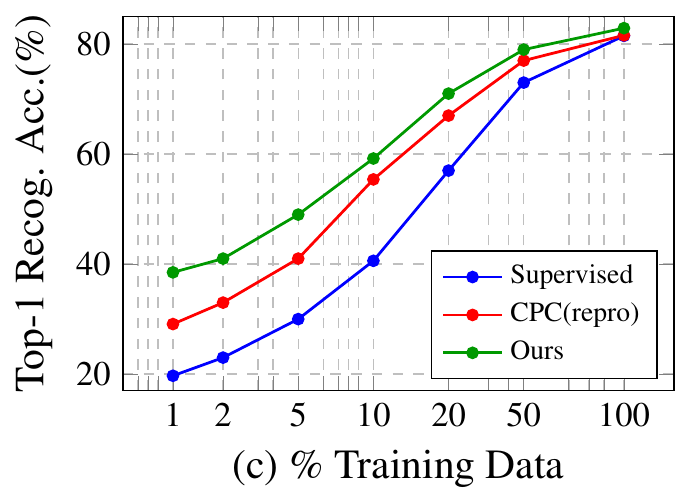}
		\includegraphics[height=2.7cm, width=0.24\linewidth]{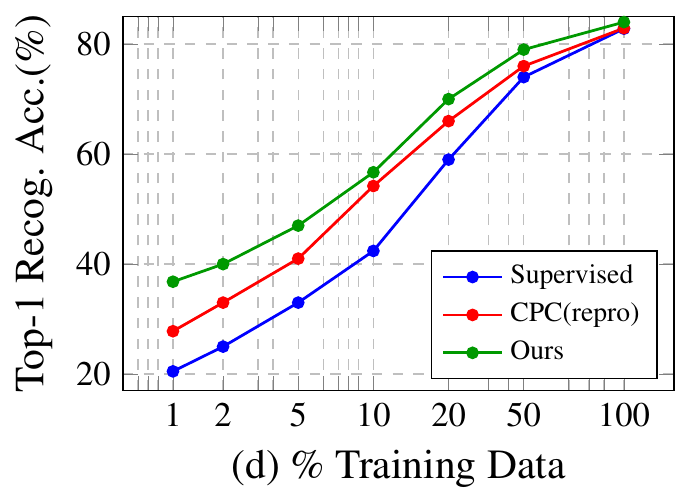} \\  
		\includegraphics[height=2.7cm, width=0.24\linewidth]{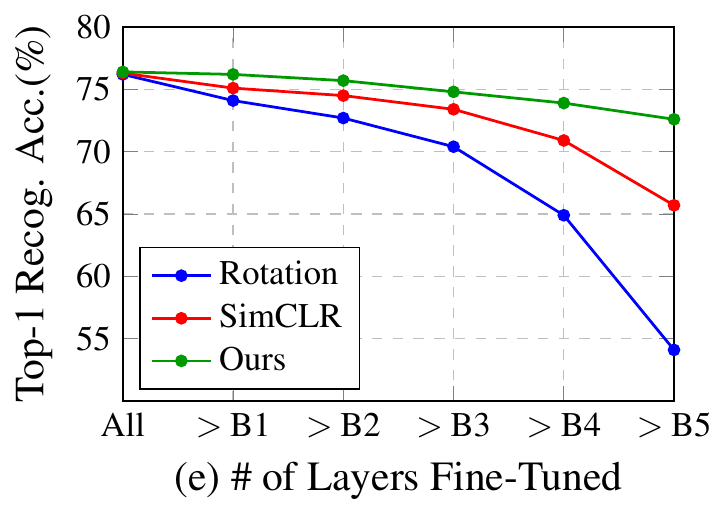}
		\includegraphics[height=2.7cm, width=0.24\linewidth]{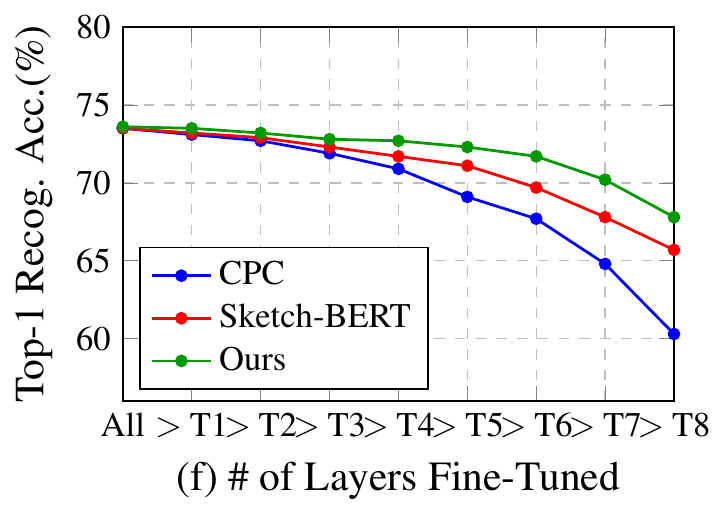}
		\includegraphics[height=2.7cm, width=0.24\linewidth]{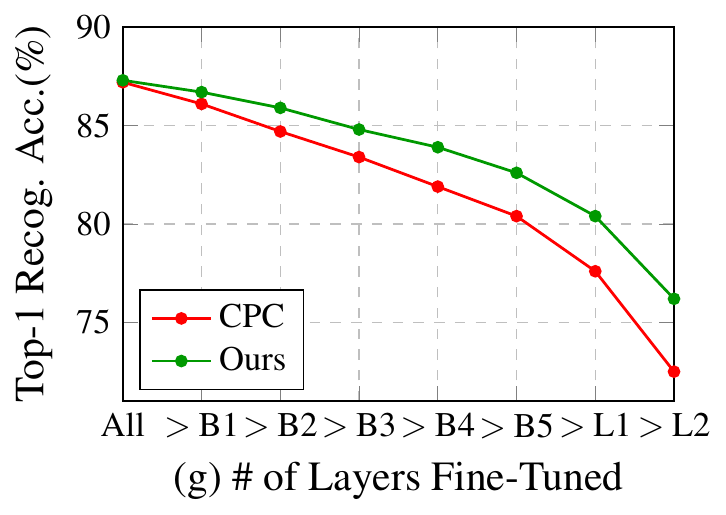}
		\includegraphics[height=2.7cm, width=0.23\linewidth]{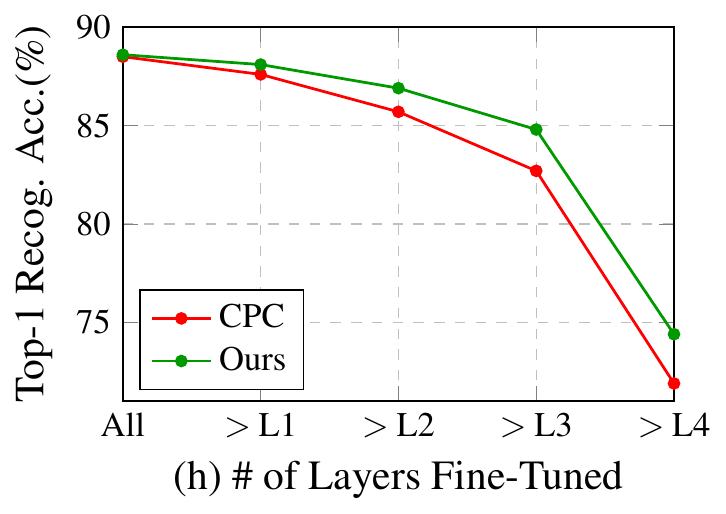}
	\vspace{-0.05in}
	\caption{Performance at varying training data size for (a) sketch image classification (b) sketch vector classification on Quick-Draw, and (c) offline handwritten image recognition (d) online handwriting recognition, respectively. In the same order, comparative performance is shown through fine-tuning different number of layers: (e) sketch image encoder uses ResNet-50 having 5-convolutional blocks. (f) 8-layers stacked Transformer is used for sketch vector encoder. (g) ResNet-like convolutional encoder (having 5 blocks) followed by 2 layers BLSTM employs offline word image encoder (h) 4 layers stacked BLSTM is used for encoder online word images.`$\mathrm{>X}$' represents all layers above $X$ are fine-tuned. 
	\label{plots}
\vspace{0.2cm}
 }
	\label{fig:data_layers}
\vspace{0.3cm}
\end{figure*}

\vspace{-0.8cm}
\subsection{Results on Handwriting Recognition}\label{htr}
To the best of our knowledge, there has been no work applying self-supervised learning to handwritten data. We compare our self-supervised  method with CPC which can handle sequential data. In Table~\ref{tab:my_label4}, we use the extracted frozen sequential feature from each encoder to train an attentional decoder based text recognition network. We see that our Sketch2Vec surpasses CPC, but both methods do not match supervised performance. In a semi-supervised setup (Table \ref{tab:my_label5}), we add an attentional decoder and fine-tune the whole pipeline using $1\%$ and $10\%$ training data respectively. In this case, the self-supervised methods achieve a significant margin over the supervised alternative. Furthermore, we observe that initialising the network (both offline and online) with weights pre-trained on our self-supervised setup, followed by training (supervised) the entire pipeline, yields higher results than initialising with random weights, by a margin of $1.4\%$ and $1.2\%$, under the same experimental setup. This concludes that our smart pre-training strategy is a better option, instead of training handwriting recognition network from scratch.

\setlength{\tabcolsep}{4pt}
\begin{table}[!htbp]
    \centering
    \footnotesize
    \caption{Ablative study (Top-1 accuracy) on architectural design using QuickDraw. (V)ectorization and (R)asterization indicate representation learning on image and vector space, respectively.}
    \vspace{-0.2cm}
    \begin{tabular}{lcc}
        \hline \hline
         Ablation Experiment & Image Space & Vector Space \\
        \hline
         (a) Absolute coordinate in the decoding (V): & 71.9\% & -- \\
         (b) Offset coordinate in the decoding (V) & 69.5\% & -- \\
         (c) Absolute coordinate in the encoding (R): & -- & 67.2\% \\
         (d) Offset coordinate in the encoding (R) & -- & 67.1\% \\
         (e) LSTM decoder (V) : & 70.7\% & -- \\
         (f) GRU decoder (V) : & 71.9\% & -- \\
         (g) Transformer decoder (V) : & 68.6\% & -- \\
         (h) LSTM encoder (R) : & -- & 66.7\% \\
         (i) GRU encoder (R) : & -- & 66.1\% \\
         (j) Transformer encoder (R) : & -- & 67.2\% \\
         (k) Two-way Translation (V+R) : & 70.3\% & 66.1\% \\
         (l) Attentional Decoder (V) : & 68.0\% & -- \\
        \hline
    \end{tabular}
    \label{tab:architechture}
    \vspace{0.5cm}
\end{table}

\begin{figure}[!hbt]
	\begin{center}
		\includegraphics[width=1\linewidth]{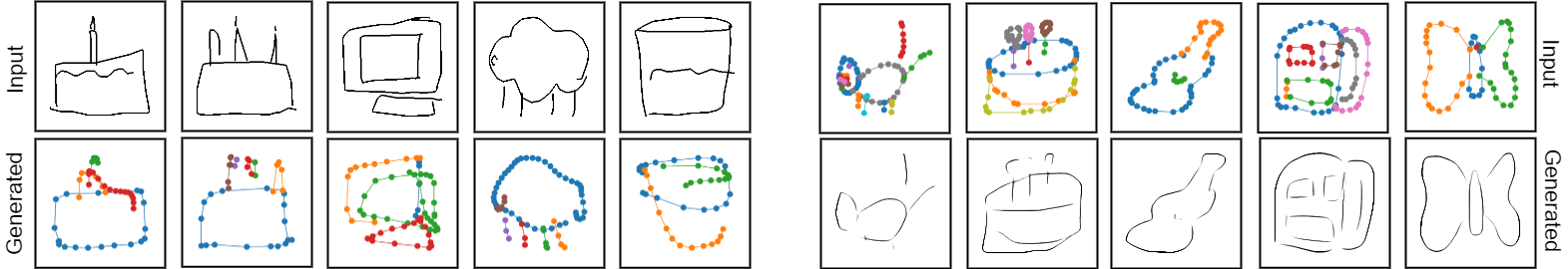}
	\end{center}
	\vspace{-0.2in}
	\caption{Qualitative results showing generated cross-modal translation. (a) Vectorization: raster sketch image to vector sketch, (b) Rasterization: vector sketch to raster sketch image.}
	\label{fig:Graph1_Full}
    \vspace{0.5cm}
\end{figure}

\subsection{Ablative Study}

\noindent \textbf{Data Volume and Layer Dependence:} 
Performance under varying amounts of training data is shown in Figure \ref{fig:data_layers} for both sketch classification and handwriting recognition. We also evaluate performance as a function of number of trained/frozen layers during fine-tuning. We can see that Sketch2Vec performs favorably to state of the art alternatives SimCLR, SketchBERT, and CPC -- especially in the low data, or few tuneable layers regimes. 

\noindent \textbf{Architectural Insights:} We perform a thorough ablative study to provide insights on our architecture design choices using the sketch recognition task in Table~\ref{tab:architechture}. (i) In the sketch image to sketch vector translation process, using  \emph{absolute coordinate}  is found to give better representative feature for sketch images over using offset coordinate \cite{ha2017neural} values. However, for representation learning over vector sketches, we did not notice any significant difference in performance, provided the absolute coordinates are normalised.  (ii) We found absolute coordinate with regression loss gives better performance than using offset with log-likelihood loss as used in \cite{song2018learning}. (iii) We use deterministic cross-modal encoder-decoder architecture since VAE-based design \cite{song2018learning} reduces the performance. (iv) We also compare with different sequential decoders in the vectorization process, e.g. LSTM, GRU, and Transformer. Empirically, GRU is found to work better than others. (v) For sketch classification on vector space, we also compare with  LSTM, GRU, and Transformer encoder architecture respectively, with Transformer giving optimum results. (vi) Another intuitive alternative could be to use two-way cross-modal translation using additional source-to-source and target-to-target decoder, however, we experience performance drop. (vii) We also add an attentional block for sequential decoding in vectorization process that leads to a drop in performance by $3.9\%$. We conjecture that adding attention gives a shortcut connection to the convolutional feature map, and the sequential task becomes comparatively easy, which is why the self-supervised pre-training fails to learn global semantic representation for classification.

\begin{figure}[!htbp]
	\begin{center}
		\includegraphics[ width=1\linewidth]{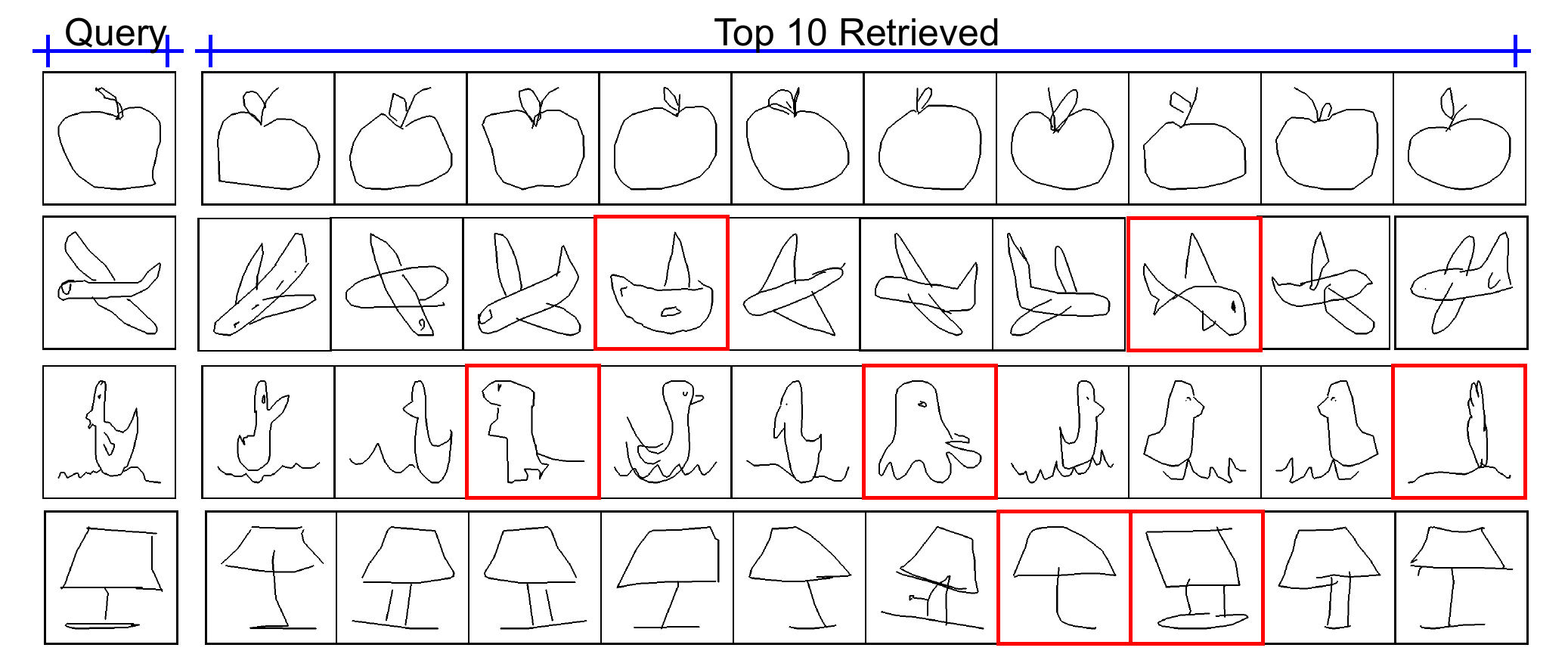}
		\includegraphics[ width=1\linewidth]{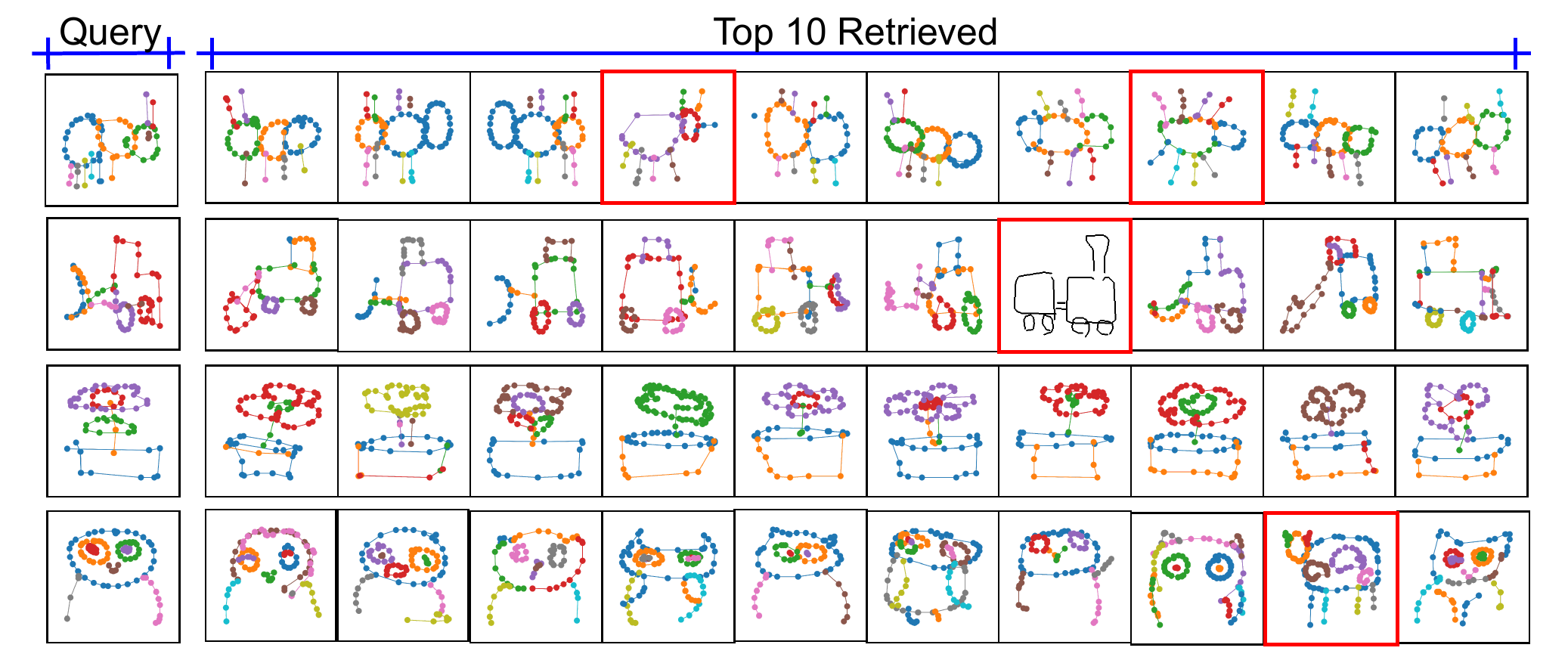}
	\end{center}
	\vspace{-0.2in}
	\caption{Qualitative retrieved results on (a)  raster sketch images (via vectorization task) (b) vector sketches (via rasterization task) using  pre-trained latent feature. Red denotes false positive cases.}
	\label{fig:Graph2_Full}
	\vspace{.5cm}
\end{figure}

\begin{figure}[!htbp]
	\begin{center}
		\includegraphics[height=4.5cm, width=1\linewidth]{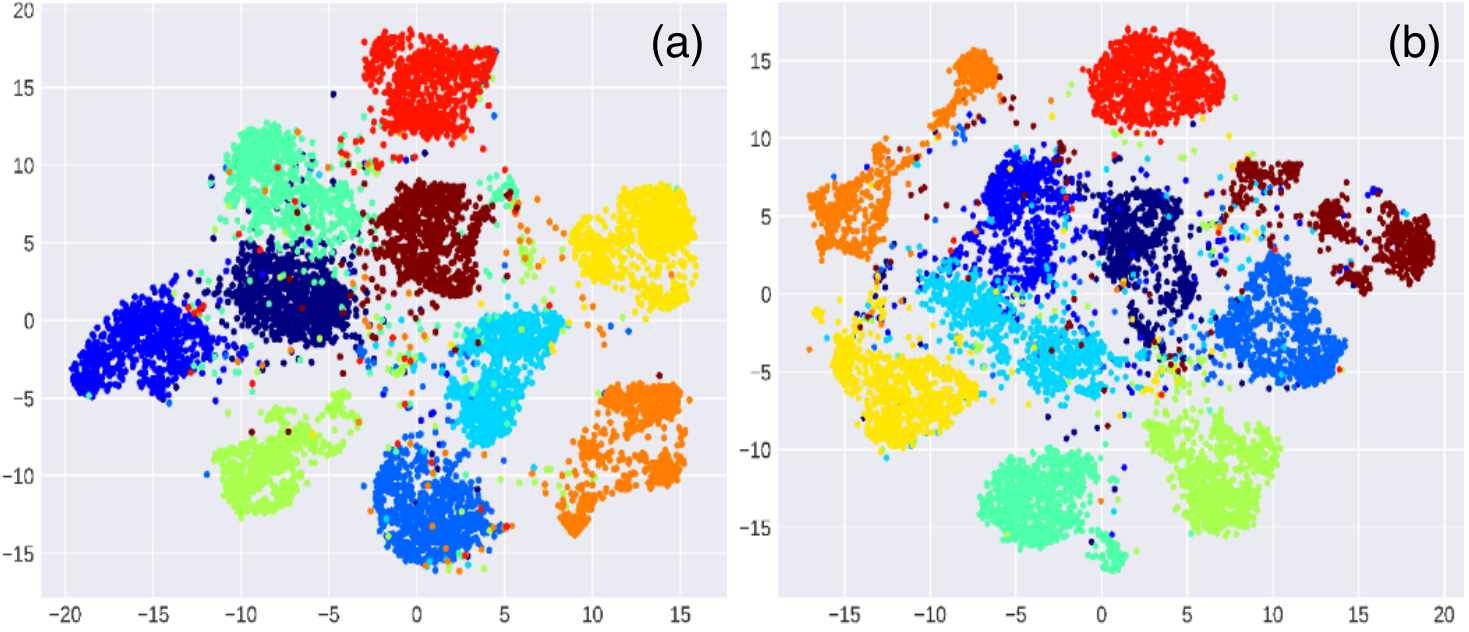}
	\end{center}
	\vspace{-0.15in}
	\caption{T-SNE Plots on features extracted by our self-supervised method (a) vectorization (sketch images) (b) rasterization (sketch vectors) for 10 QuickDraw classes.}
	\label{fig:Graph3_Full}
\end{figure}

\noindent \textbf{Cross-category Generalisation:} One major objective of unsupervised representation learning is to learn feature representation that can generalise to other categories as well. Thus, we split 345 QuickDraw classes into two random disjoint set \cite{dey2019doodle} of $265$ and $80$ for self-supervised training and evaluation, respectively. Model trained using our self-supervised task, is further evaluated on unseen classes (Table \ref{tab:cross-category}) using a linear classifier on extracted frozen feature. We obtain a top-1 accuracy of $65.1\%$ and  $58.4\%$ on sketch images and vectors, respectively, compared to  $71.9\%$ and  $67.2\%$ while using all classes in the self-supervised pre-training. Under same setting, SimCLR is limited to $53.6\%$ for sketch image classification. This confirms a significant extent of generalizable feature learning through our self-supervised task over sketch data.

\setlength{\tabcolsep}{10pt}
\begin{table}[!ht]
    \centering
    \footnotesize
    \caption{Cross-category recognition accuracy on QuickDraw.}
     \vspace{-0.2cm}
    \begin{tabular}{c|cc|cc}
        \hline \hline
         & \multicolumn{2}{c|}{Image Space} & \multicolumn{2}{c}{Vector Space} \\
        \cline{2-5}
         & Top-1 & Top-5 & Top-1 & Top-5 \\
        \hline
        MoCo \cite{he2020momentum} & 53.4\% & 77.6\% & -- & -- \\
        SimCLR \cite{chen2020simple} & 53.6\% & 77.6\% & -- & -- \\
        CPC \cite{oord2018representation} & 46.8\% & 71.3\% & 48.1\% & 73.3\% \\
        Ours & 65.1\% & 85.6\% & 58.4\% & 81.2\% \\
        \hline
    \end{tabular}
    \label{tab:cross-category}
    \vspace{0.2cm}
\end{table}

\noindent \textbf{Cross-dataset Generalisation:} We further use model trained on QuickDraw dataset to extract feature over TU-Berlin dataset, followed by linear evaluation. Compared within dataset training accuracy of $70.6\%$ ($55.6\%$), we obtain a cross-dataset accuracy (Table \ref{tab:cross-dataset}) of $58.9\%$ ($36.9\%$) on TU-Berlin sketch-images (sketch-vectors) without much significant drop in accuracy, thus signifying the potential of our self-supervised method for sketch data.  

\setlength{\tabcolsep}{10pt}
\begin{table}[!ht]
    \centering
    \footnotesize
    \caption{Cross-dataset (QuickDraw $\mapsto$ Tu-Berlin) recognition accuracy: Model pre-trained on QuickDraw is used to extract fixed latent feature on TU-Berlin, followed by linear model evaluation.}
    \begin{tabular}{c|cc|cc}
        \hline \hline
         & \multicolumn{2}{c|}{Image Space} & \multicolumn{2}{c}{Vector Space} \\
        \cline{2-5}
         & Top-1 & Top-5 & Top-1 & Top-5 \\
        \hline
        MoCo \cite{he2020momentum} & 47.5\% & 62.1\% & -- & -- \\
        SimCLR \cite{chen2020simple} & 47.2\% & 62.0\% & -- & -- \\
        CPC \cite{oord2018representation} & 41.4\% & 60.8\% & 27.7\% & 50.9\% \\
        Ours & 58.9\% & 80.5\% & 36.9\% & 61.7\% \\
        \hline
    \end{tabular}
    \vspace{0.2cm}
    \label{tab:cross-dataset}
\end{table}

\noindent \textbf{Cross-Task Generalisation:} Both sketch and handwriting, are hand-drawn data having similarity in terms of how they are recorded, and represented in image and vector space. Thus, we explore whether a model trained on handwritten data using self-supervised task can generalise over sketches, and vice versa. The pooling stride is adjusted so that, using sketch convolutional encoder, we can get sequential feature, and handwriting convolutional encoder can give feature vector representation on sketch images using global pooling. Following this protocol, we obtain (Table \ref{tab:cross-task}) a reasonable cross-task top-1 accuracy of $37.6\%$ and $33.7\%$ on sketch images and vectors on QuickDraw. Conversely, we get no-lexicon WRA of $28.4\%$ and $26.3\%$ on handwritten offline word images and online word vectors, respectively.

\setlength{\tabcolsep}{2pt}
\begin{table}[!ht]
    \centering
    \footnotesize
    \caption{Cross-task (Sketch $\leftrightarrow$ Handwriting) generalisation results on extracted fixed latent feature. Lexicon: (L), No-Lexicon: (NL) }
    \vspace{-0.1cm}
    \begin{tabular}{c|cc|cc|cc|cc}
        \hline \hline
         & \multicolumn{4}{c|}{Sketch (QuickDraw)} & \multicolumn{4}{c}{Handwriting (IAM)} \\
        \cline{2-9}
         & \multicolumn{2}{c|}{Image} & \multicolumn{2}{c|}{Vector} & \multicolumn{2}{c|}{Image} & \multicolumn{2}{c}{Vector} \\
         & Top-1 & Top-5 & Top-1 & Top-5 & L & NL & L & NL \\
        \hline
        Random & 14.6\% & 25.7\% & 11.8\% & 22.9\% & 9.8\% & 6.1\% & 7.1\% & 4.5\% \\
        \hdashline
        CPC \cite{oord2018representation} & 19.7\% & 37.8\% & 17.6\% & 36.9\% & 19.5\% & 12.5\% & 15.7\% & 9.7\% \\
        Ours & 37.6\% & 58.4\% & 33.7\% & 55.8\% & 33.8\% & 28.4\% & 31.6\% & 26.3\% \\
        \hline
    \end{tabular}
    \label{tab:cross-task}
    \vspace{0.5cm}
\end{table}

\noindent \textbf{Further Analysis} (i) We have also compared with other backbone CNN network, e.g. AlexNet, where we obtain Top-1 accuracy of $63.3\%$ compared to $71.9\%$ on using ResNet50. This confirms suitability of our design across different backbone architecture. (ii) In our implementation, we perform only basic horizontal flipping and random cropping for augmentation. We also experiment with multiple augmentation strategies mentioned in \cite{chen2020simple}, but notice no significant changes. (iii) Furthermore, following the recent works \cite{xu2018sketchmate, collomosse2019livesketch} that jointly exploits raster sketch-image and temporal sketch-vector for sketch representation, we simply concatenate extracted latent feature of sketch-image and sketch-vector, and evaluate through linear classifier. This joint feature improves the top-1 accuracy to $72.8\%$ compared to $71.9\%$ which uses raster image only.

\section{Conclusion}
We have introduced a self-supervision method based on
cross-modal rasterization/vectorization that is effective in
representation learning for sketch and handwritten data.
Uniquely our setup provides powerful representations for
both vector and raster format inputs downstream. Results on sketch recognition, sketch retrieval, and handwriting recognition show that our pre-trained representation approaches the performance of supervised deep learning in the
full data regime, and surpasses it in the low data regime.
Thus Sketch2Vec provides a powerful tool to scale and accelerate deep-learning-based freehand writing analysis going forward.

{Now that we have addressed unsupervised learning of a good sketch representation, practicality for sketch-based visual understanding demands, that sketch \textit{as a modality} be used to solve \aneeshan{more general computer vision tasks. For instance sketches in some form should be able to
ease general training for tasks,} where pixel prefect images themselves are unavailable for training. Say a situation comes where there are privacy concerns or copyright infringements over accessing images, it would be interesting if a pre-trained model can extend its knowledge to a few more classes by training upon just a few sketches doodled by users. In our next chapter therefore, we develop a novel problem formulation which shows that sketch could be used as a potential support modality for few-shot class incremental learning without violating any data-privacy concern.}
\chapter{Sketch-Based Incremental Learning}
    \label{chapter:SBIL}
    \minitoc
    \newpage
    
    \section{Overview}
    


\begin{figure}[!hbt]
\centering
\includegraphics[width=1\linewidth]{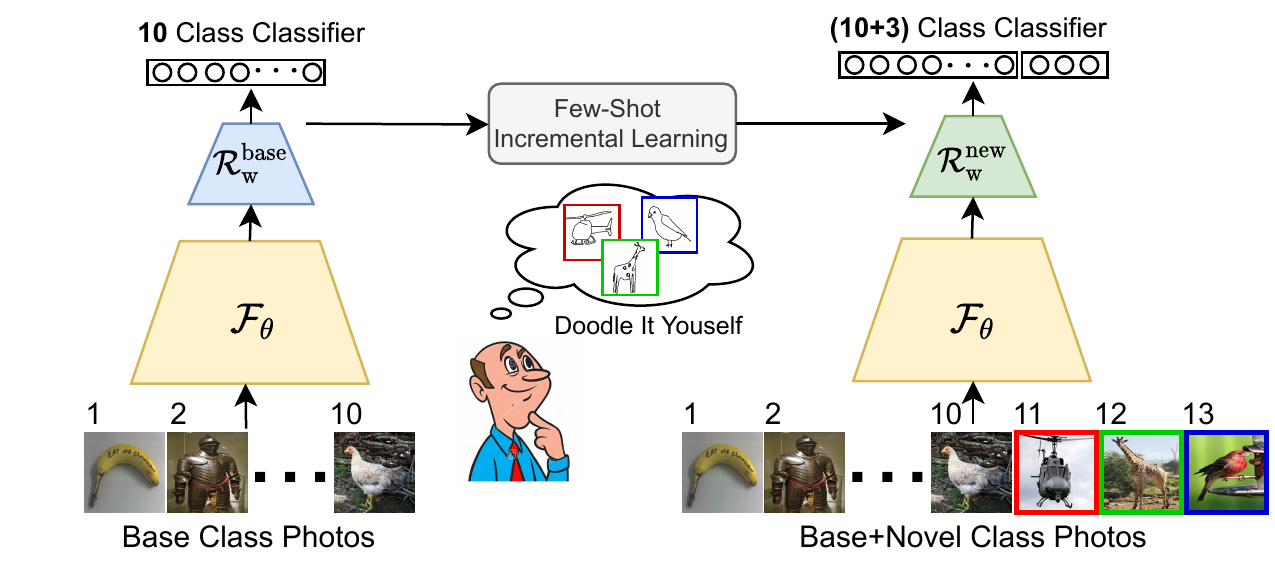}
\vspace{-0.8cm}
\caption{Illustration of our DIY-FSCIL framework. For instance, given \emph{sketch} exemplars (1-shot here) from $3$ novel classes as support-set, a $10$-class classifier gets updated to $(10+3)$-class classifier that can classify photos from both base and novel classes.} 
\label{chapter:SBIL:Fig1}
\vspace{-0.2cm}
\end{figure}

\aneeshan{Keeping parity with our second main theme of the thesis, we explore if it is possible to extend sketches as a support modality for different computer vision downstream tasks.}
In this chapter, we therefore explore the role of human sketches as a support modality for Few-Shot Class Incremental Learning (FSCIL). This results in a flexible FSCIL system that learns new classes just by observing a few sketches \emph{doodled} by users themselves. Fig.~\ref{chapter:SBIL:Fig1} schematically illustrates our ``Doodle It Yourself (DIY)" FSCIL scenario -- ``DIY-FSCIL". {This importantly addresses the aforementioned problems in that (i) learning is no longer fixed to just photos but flexibly cross-modal with other data forms (just as humans do),} and (ii) it works without asking the users to source photos which might have practical constraints attached (e.g., copyright, hazardous environments). There is of course also the added benefit of injecting creativity to the classifier by sketching something off the user's imagination \cite{ge2020creative}, e.g., a ``flying cow"?

\vspace{-0.2cm}

The advocate for sketches is largely motivated by the great line of work examining human-centric characteristics of sketches in many parallel applications -- notably image retrieval \cite{dey2019doodle}, where the fine-grained nature of sketches  is used to successfully conduct instance-level retrieval \cite{bhunia2021more, bhunia2020sketch, sain2021stylemeup, dutta2019semantically, collomosse2019livesketch}.
Sketches in context of FSCIL is closely reminiscent of its usage in fine-grained retrieval. While in retrieval they utilise the detailed nature of sketches to conduct sketch-photo matching, we use a few sketches collectively as faithful visual representatives (support) of novel classes for incremental learning.
    
Nonetheless, using sketches as class support in the FSCIL setting is non-trivial. \emph{Sketch}, despite being visually representative, is just a coarse contour-like depiction of the visual world, that sit in an entirely different domain from photo \cite{li2019episodic}. Thus, off-the-shelf models naively pre-trained on photos commonly fail to generalise well on sketches \cite{chen2019closer}. Moreover, due to its highly abstract nature, the same object may be sketched in various ways under unique user-styles \cite{song2018learning, sain2021stylemeup}, and with varied levels of detail \cite{sain2020cross}. We are also distinctly different to the parallel problem of SBIR -- SBIR typically gets exposed to \textit{paired sketch-photo} data at training to learn a cross-modal embedding. We on the other hand need to work with \textit{sketches only} at training (i.e., no photo information whatsoever), yet still aim to generate classification layer weights to classify \textit{photos} from novel classes. 

Three key design considerations for this cross-domain sketch-based FSCIL are: (i) how to make the model work cross-domain, (ii) how to preserve old class information, and (iii) how to leverage information from old classes to learn new ones. For the first issue, we design a gradient consensus based strategy that updates the model towards mutual agreement in the gradient space between sketch and photo domain, thus achieving a domain invariant feature extractor.  
For the second, we model an additional knowledge distillation loss to retain the acquired knowledge from old classes while incrementing the classifier to novel classes. Lastly, we devise a graph neural network to generate more discriminative decision boundaries for the incremented classifier via message passing between old and novel classes.

To sum up, our contributions are: (a) We extend the general line of incremental learning research even further towards practicality. (b) {We achieve that by introducing sketches as class support for FSCIL, allowing the system to learn from modalities other than just photos, and addressing issues around ethics and privacy while allowing user creativity.} (c) We introduce the first cross-modal framework to tackle this novel DIY-FSCIL problem.

\section{Problem Formulation}

\subsection{Dataset}
In few-shot class-incremental learning, we are provided with $K_{b}$ \emph{base} classes and $K_{n}$ \emph{novel} classes respectively. From the set of base classes, we have \emph{sufficient} access to labelled samples from \emph{photo} $\mathcal{D}_{base}^P=\{(p_i, {{y_i^p}} )\}_{i=1}^{N_b^s}$   and \emph{sketch}  $\mathcal{D}_{base}^S = \{(s_i, {{y_i^s}} )\}_{i=1}^{N_b^s}$ domains, where $y_i \in \mathcal{C}_{base} = \{C_1^b, C_2^b, \cdots, C_{K_b}^b\}$. On the other hand, for novel classes, we have \emph{minimal} access to labelled samples from \emph{only sketch} domain $\mathcal{D}_{novel}^S = \{(s_j, y_j)\}_{j=1}^{N_n^s}$ where number of samples for each novel category is limited, and $y_j \in \mathcal{C}_{novel} = \{C_1^n, C_2^n, \cdots, C_{K_n}^n\}$. Here, base and novel classes are completely disjoint, such that $\mathcal{C}_{base} \cap\mathcal{C}_{novel} = \Phi$.

\subsection{Model}
We have a neural network classifier, consisting of a feature extractor $\mathcal{F}_{\theta}$ followed by a linear classifier $\mathcal{R}_{w}$, such that $y = \mathcal{R}_{w}(\mathcal{F}_{\theta}(x))$. $\mathcal{F}_{\theta}$ is employed using a convolutional neural network followed by global-average pooling, and given an input image $x \in \mathbb{R}^{h \times w \times 3}$, we get a feature representation as $f_d = \mathcal{F}_{\theta}(x) \in \mathbb{R}^d$. Following \cite{gidaris2018dynamic}, for better generalisation $\mathcal{R}_{w}$ is devised as a \emph{cosine similarity}  function (unlike dot product based typical linear classifier), consisting a learnable $W$ matrix whose size is of $\mathbb{R}^{|C|\times d}$, where $|C|$ is the number of classes. Thus, $\mathcal{R}_{w}: \mathbb{R}^d \rightarrow \mathbb{R}^{|C|}$ outputs a probability distribution over classes as $p(\bar{y}) = {\texttt{softmax}( \hat{W}\cdot \frac{f_d}{{\left \| f_d \right \|_2}})}$. ${\hat{W}}$ is obtained by $l_2$ normalising every $d$ dimensional row-vector ${w}_k \in {W}$ that depicts weight-vector for $k^{th}$ class, \emph{i.e.} $\hat{w}_k =\frac{w_k}{{\left \| w_k \right \|_2}}$.

\subsection{Learning Objective}
The neural network classifier $\{\mathcal{F}_{\theta}, \mathcal{R}_{w}\}$ is trained from the abundant labelled samples of $K_b$ base classes, and let the initial base classifier be $\mathcal{R}_{w}^{base}: \mathbb{R}^{d} \rightarrow \mathbb{R}^{K_{b}}$ whose weight matrix is $W_{base} \in \mathbb{R}^{K_{b} \times d}$. During inference under FSCIL \cite{tao2020few}, we do \emph{not have} any access to labelled data of base classes, and given only $k$ (small number) \emph{sketch} samples for each of $\mathcal{C}_{n}$ novel categories, we intend to update the classifier  $\mathcal{R}_{w}^{base}$ to $\mathcal{R}_{w}^{new}$ which can recognise \emph{photos} from both $\mathcal{C}_{base} \cup \mathcal{C}_{novel} $ classes. To do so, we need to compute a new weight matrix ${W_{new}} \in \mathbb{R}^{(K_{b}+K_{n}) \times d}$ with respect to $\mathcal{R}_{w}^{new}: \mathbb{R}^{d} \rightarrow \mathbb{R}^{(K_{b}+K_{n})}$ that can perform $(K_{b}+K_{n})${-way} class classification. 


Therefore, our objective is to figure out a new ${W_{new}}$ matrix for classifier $\mathcal{R}_{w}^{new}$ using the previous base classes' knowledge $W_{base}$  and a few hand-drawn \emph{sketch exemplars} from novel classes such that (i) the knowledge of base classes is not forgotten (preserved), as well as (ii) it quickly adapts to novel classes using few samples, (iii) thus, enabling it to perform well on real photos minimising the domain gap \cite{li2018learning} with sketch samples from novel classes as support. 
Overall, our framework consists of \emph{three} modules {(i)} a backbone \emph{feature extractor} $\mathcal{F}_\theta$, {(ii)} a \emph{classifier} $\mathcal{R}_w$ {(iii)} a \emph{weight generator} $G_{\psi}$ that will take previous base classifier weights $W_{base}$ and sketch exemplars (support set) from novel classes as input, to generate a new weight matrix $W_{new}$ for updated classifier $\mathcal{R}_{w}^{base} \rightarrow 
\mathcal{R}_{w}^{new}$ in order to classify real photos from \emph{both} base and novel classes.

\section{Proposed Solution}

Our framework follows a \emph{two}-stage training. In the first stage, we train the model on base classes using standard cross-entropy loss, while in the second stage, we learn the \emph{weight generator} via few-shot pseudo-incremental learning. \cut{We assume that we have sufficient access to labelled data from base classes.} Once trained in the first stage, we freeze the weights of $\mathcal{F}_{\theta}$ in the next stage to (i) avoid over-fitting during the few-shot update and (ii) to alleviate catastrophic forgetting \cite{kirkpatrick2017overcoming} of the base classes. 

\vspace{0.25cm}
\begin{figure}[!hbt]
\includegraphics[width=1\linewidth]{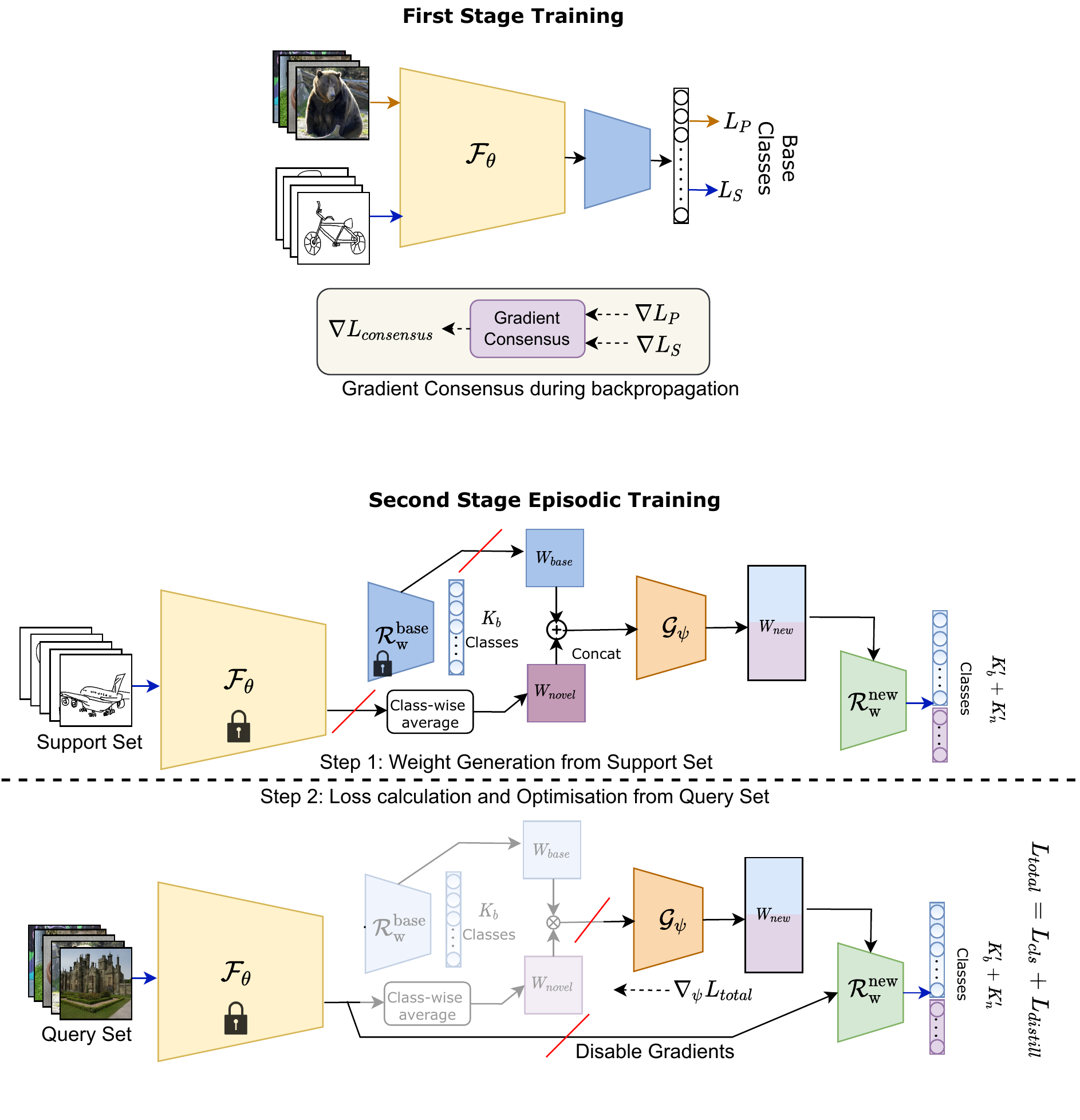}
\caption{(a) Primarily, we aim to learn a domain-agnostic backbone feature extractor ($\mathcal{F}_{\theta}$) through gradient consensus. (b) In the second stage, we learn a weight generator ($G_{\psi}$) via episodic pseudo incremental learning involving two steps. Firstly, to obtain an updated [base+novel] classifier, a sketch support set is utilised to produce weight vectors for novel classes as well as to refine weight vectors for base classes. Secondly, for loss computation, the resulting weight vectors are evaluated against real photos from both [base+novel] classes.}
\label{fig_arch}
\vspace{0.5cm}
\end{figure} 
 
\subsection{Cross Modal Pre-Training for Base Classes}

Unlike existing few-shot incremental learning, we need to handle the domain gap \cite{li2019episodic} between photos and sketches, so that the knowledge of incremental classes acquired through sketch exemplars can classify novel class images in real photo domain.
As we have sufficient access to labelled training data from both photo and sketch domains for base classes, a very straightforward way to handle the domain gap is to train by combining labelled photos and sketches (spatially extended) with equal probability in every mini-batch -- so that the model generalises equally well on both photos and sketches. Given an input $x$, let the model's output be $\bar{y} = \mathcal{R}_{w}^{base}(\mathcal{F}_{\theta}(x))$ where labelled data $(x, y)$ come from either photo $(p,{{y^p}}) \sim \mathcal{D}_{base}^P$ or sketch $(s,{{y^s}}) \sim \mathcal{D}_{base}^S$ domain with $x \in \{p, s\}$ and $y$ ({{$y^s$ or $y^p$}}) being corresponding one-hot encoded class label. The cross-entropy loss  $\mathcal{H}(\cdot,\cdot)$ can be calculated as $\mathcal{L} = \mathcal{H}(\bar{y},y) = -\sum_{i=1}^{K_{b}} y_i \log p(\bar{y}_i)$. Against a batch having $b$ photos and $b$ sketches, we can calculate the individual loss across photos and sketches as $\mathcal{L}_{P}$ and $\mathcal{L}_{S}$, respectively. Thereafter, we update the model by taking gradient $\nabla  \mathcal{L}_{total}$ over total loss which is given as follows:  
 
\begin{equation}
    \mathcal{L}_{total} = \frac{1}{b} \hspace{-0.5cm} \sum_{(p,y)  \sim \mathcal{D}_{base}^P} \hspace{-0.5cm} \mathcal{L}_{P}(p,y) +  \frac{1}{b} \hspace{-0.5cm} \sum_{(s,y)  \sim \mathcal{D}_{base}^S} \hspace{-0.5cm} \mathcal{L}_{S}(s,y)
\end{equation}

However, naively training with two significantly different domains (photo vs sketch) gives rise to \emph{conflicting gradients} within each batch, as information specific to the one domain might be irrelevant to the other, thereby suppressing the generalisation capability of the model. In other words, the information carried by $\nabla  \mathcal{L}_{P}$ and $\nabla  \mathcal{L}_{S}$ might not mutually agree, and adding them naively would lead to inhibiting \cite{yu2020gradient} the overall training signal.

\subsection{Gradient Consensus}  

Inspired from multi-task learning \cite{yu2020gradient} and domain generalisation \cite{mansilla2021domain} literature, we aim to update the model in the direction where there is an agreement in the gradient space between two domains in order to learn a domain invariant representation. In particular,  gradient vectors having the same sign will be retained, while those having conflicting signs will be set to \emph{zero}, as shown in Eq. \ref{gradsurgery}. Here, the $\mathrm{sig}(\cdot)$ is a sign operator, and $\nabla L_{P}^n$ and $\nabla L_{S}^n$ denote the $n$-th component of the gradient associated to photo and sketch domain respectively. The gradient consensus function $\delta(\cdot, \cdot)$ checks element-wise if the signs of the gradient components match, and returns 1 if all components have the same sign for a given $n$; otherwise 0.

\begin{equation}\label{gradsurgery}
 \delta (\nabla L_{P}^n, \nabla L_{S}^n) =\left\{\begin{array}{lll}
                1, &  \mathrm{sig}(\nabla L_{P}^n)= \mathrm{sig}(\nabla L_{S}^n) \\
                0, &  \textnormal{otherwise}
            \end{array}\right.
\end{equation} 
 
\begin{equation}\label{consensus_grad}
\nabla L_{consensus}^{n} =\left\{\begin{array}{lll}
                \nabla L_{P}^{n} + \nabla L_{S}^{n}, & \textnormal{if} &  \delta^n =1 \\
                0, &  \textnormal{if} &  \delta^n=0
            \end{array}\right.
\end{equation}

Here, we set the sign-conflicting gradient vectors to zero and sum the gradient vectors only when there is a sign-agreement. This gradient agreement strategy helps to reduce the harmful cross-domain gradient interference while updating the model parameters using $\nabla L_{consensus}^{n}$. Thus, enabling us to adjust the model parameters in a direction that helps to improve generalisation across both sketch and photo.

\subsection{Few-Shot Classifier Weight Generation}
In order to classify photo from novel classes, we need to design a mechanism that can generate \emph{additional} weight vectors for the novel classes. 
As we assume that only a few supporting hand-drawn sketch exemplars will be provided corresponding to every novel class, we design the weight generator $G_{\psi}$ under a few-shot paradigm \cite{wang2020generalizing}. $G_{\psi}$ produces weight vectors for novel classes and also re-generates (refines) weight vectors for base classes in order to get a better overall decision boundary in the presence of novel classes. Here, the two major objectives are (i) learn the knowledge of novel classes from \emph{fewer sketch exemplars}, while \emph{classifying photos} of novel classes through cross-modal generalisation (ii) \emph{not to degrade} the performance of base classes while learning the novel ones.

We employ sketch exemplars as a \emph{support set} to generate the \emph{new} weight matrix following the episodic training \cite{li2019episodic} of few-shot learning. To determine the loss for updating the weight generating module, the quality of the generated weight matrix is assessed against a \emph{query set} of photo samples. In particular, there are \emph{two} steps \cite{snell2017prototypical} while training the weight generation module. (i) {\emph{Weight generation} using support set:} sketch exemplars as support set are used together with $W_{base}$ to generate the new weight matrix $W_{new}$ (comprising both base and novel classes) (ii) {\emph{Loss calculation} on query set:}  $W_{new}$ is used to classify query set \emph{photos} in order to calculate loss, which is then utilised to optimise the weight generation module using gradient descent.

\subsection{Baseline Weight Generation}

$G_{\psi}$ takes two things as input (i) $W_{base}$ from $\mathcal{R}_{w}^{base}$ representing the knowledge of previous base classes (ii) class-wise representative features of novel classes from sketch exemplars. We assume to have access to $k$ sketch samples for each of the $K_{n}$ novel classes -- the support set. A straightforward way to get class-wise representative vectors is to average feature representations of sketches for each individual class. In particular, for $j^{th}$ novel class, the representative vector can be calculated as:
 
\begin{equation}
    w_{j}^{novel} = \frac{1}{k}\sum_{i=1}^{k} \mathcal{F}_{\theta}(s_i)
\end{equation}

Thereafter, by applying $l_2$ norm on each $ w_{j}^{novel}$, we can naively form the weight vectors of novel classes as $W_{novel} = \{w_{1}^{novel}, w_{2}^{novel}, \cdots, w_{K_{n}}^{novel}\} \in \mathbb{R}^{K_n \times d}$. The easiest way for incremental learning would be to use naive concatenation to get new weight matrix as $[W_{base}; W_{novel}] \in \mathbb{R}^{(K_{b} + K_{n}) \times d}$. However, it has \emph{two} major limitations (i) $W_{novel}$ remains unaware about the knowledge of bases classes (ii) $W_{base}$ which was discriminative across the base classes might lose its representation-potential when we add additional weight vectors of novel classes without modelling a mutual agreement strategy for learning discriminative decision boundaries across all $K_{b} + K_{n}$ classes. Thus, to attain an optimal decision boundary for all classes under incremental setup, an \emph{information passing} mechanism is critical for $W_{new}$ generation.

\subsection{Message Passing}

For \emph{information-propagation} among weight vectors of $K_{b}+K_{n}$ classes, we use Graph Attention Network (GAT) \cite{velivckovic2017graph}. GAT is a good choice for information-propagation owing to its permutation-invariance to sequence of weight vectors as the novel classes may appear in any order. As the weights are shared across different nodes, it can also handle incoming variable number of novel classes effortlessly. The input to GAT is given as $W_I = \{w_{1}^{base}, \cdots, w_{K_{b}}^{base}, w_{1}^{novel}, \cdots, w_{K_{n}}^{novel}\}$ having $K_{total} = K_{b} + K_{n}$ weight vectors, where each $w_i \in \mathbb{R}^d$ denotes an input to a specific node to GAT. First it computes relation co-efficient between every pair of nodes by inner product operation as $e_{i,j} =  \left \langle V_aw_i, V_bw_j \right \rangle$, with two learnable linear embedding weights $V_a$ and $V_b$. $e_{i,j}$ is  normalised by softmax function to get the attention weights with respect to node $i$ as: $a_{ij} = \frac{\exp(e_{ij})}{\sum_{k=1}^{K_{total}} \exp(e_{ik})}$. The update rule for $i^{th}$ node gathering information from all other nodes becomes  
 
\begin{equation}
    w_{i}^{update} = w_{i} + \big(\sum_{j=1}^{K_{total}}a_{i,j} V_c w_{j}\big)
\end{equation}

\noindent where, $V_c$ is a learnable linear transformation. We repeatedly update the weight vectors at every node in the graph, and finally we obtain the generated weight vectors for both base and novel classes as $W_{new}$.
In brief, $W_{new} = G_\psi(W_I): \mathbb{R}^{(K_b+K_n) \times d} \rightarrow \mathbb{R}^{(K_b+K_n) \times d}$, 
where $W_I = [W_{base}; W_{novel}] \in \mathbb{R}^{(K_b+K_n) \times d}$.
We thus generate the weight vectors for both base and novel classes during incremental learning.

\subsection{Episodic Pseudo Incremental Training}

Keeping the feature extractor $\mathcal{F}_\theta$ fixed, we train the few-shot weight generator $G_\psi$ taking inspiration from \cut{the recent developments of} few-shot learning literature \cite{snell2017prototypical, rezende2016one, santoro2016meta}. 
As the training dataset is limited, we episodically construct pseudo incremental task based \emph{only} on the base classes to mimic the real testing scenario. 

In particular, following the first stage of training, we get classifier weight matrix of base classes as $W_{base} \in \mathbb{R}^{K_{b} \times d}$.
In order to create each episode, we \emph{synthetically drop} $K_{n}'$ weight vectors from $W_{base}$, and we treat those corresponding classes as pseudo novel classes whose weights now need to be generated. That means, at a particular episode, the pseudo base class matrix becomes $W_{base}' \in \mathbb{R}^{K_{b}' \times d}$ where $K_{b}' = K_{b} - K_{n}'$. Thereafter, corresponding to those \emph{dropped} base classes which now become pseudo novel classes, we use $k$ sketch samples for each of the pseudo novel classes as the \emph{support set} to first generate representative class-wise weight vectors $W_{novel}' \in \mathbb{R}^{K_{n}' \times d}$, which is again fed to GAT together with $W_{base}'$ for relationship modelling to generate pseudo $W_{new}'$. In every episode, while support set ($\mathcal{S}$) is used to generate the classifier weights, another query set ($\mathcal{Q}$) involving real photos from both pseudo base and novel classes are fed through pre-trained backbone followed by classifier with newly generated weight matrix $W_{new}'$ to compute loss for optimisation. Please refer to Fig.~\ref{fig_arch}.

In contrast to earlier FSCIL works \cite{tao2020few, dong2021few}, our episodic training is cross-modal in nature, where the support and query set consists of sketch and photo respectively. As training is done over base classes with pseudo-novel classes, we found mixing both sketch and photo in the support set with \emph{gradient consensus} generalises better on real photos. However, sketches act as the only exemplars during real inference.

\subsection{Loss Functions}
Contrary to fully supervised classification from abundant training data, few-shot learning \cite{snell2017prototypical} is more challenging as only a few samples are available to generate the new weight matrix. Given this rationale, we aim to design the pseudo incremental learning by dropping weight-vectors from $W_{base}$, which is learned from base classes via standard supervised classification. We aim to see if the fully supervised knowledge learned in $W_{base}$ could provide training signal \cite{hinton2015distilling} to learn the $G_{\psi}$. To do so, we additionally define a \emph{distillation loss}  along with standard \emph{classification loss} calculated over the query set, which acts a consistency regularisation. This ensures that weight vectors predicted by the weight generator remain close to what has been learned through supervised classification from the first stage. In particular, following few-shot weight generation we get an incrementally learned classifier $\mathcal{R}_{w}^{new}$ with generated weight matrix $W_{new}$. On the other hand we already have $\mathcal{R}_{w}^{base}$, learned from first stage pre-training. Given a photo $p$ from query set ($\mathcal{Q}$), we treat the \emph{soft} prediction using $\mathcal{R}_{w}^{base}$ as a ground-truth to calculate the distillation loss. Thus, the total loss becomes $\mathcal{L}_{total} = \mathcal{L}_{cls} + \mathcal{L}_{distil}$ which is used to train $G_\psi$. \cut{Given}If $\mathcal{H}(\cdot, \cdot)$ be the cross-entropy loss, $\mathcal{L}_{cls}$ and $\mathcal{L}_{distil}$ are defined as:

\begin{equation}
 \mathcal{L}_{cls}  =  \frac{1}{ \left | \mathcal{Q} \right |}  \hspace{-0.1cm}   \sum_{(p,y)  \sim \mathcal{Q}} \hspace{-0.25cm} \mathcal{H}(\mathcal{R}_w^{new}({\mathcal{F}_{\theta}(p)}), y)
\end{equation}
 
\begin{equation}
\mathcal{L}_{distil} = \frac{1}{ \left | \mathcal{Q} \right |} \hspace{-0.2cm}  \sum_{(p,y)  \sim \mathcal{Q}} \hspace{-0.2cm} \mathcal{H}(\mathcal{R}_w^{new}({\mathcal{F}_{\theta}(p)}), \mathcal{R}_w^{base}({\mathcal{F}_{\theta}(p)}))
\end{equation}
    
\section{Experiments}
\subsection{Dataset} 
We evaluate our DIY-FSCIL framework on the popular Sketchy dataset \cite{sangkloy2016sketchy} which is a large collection of photo-sketch pairs. As paired photo-sketch is \emph{not essential} for our framework, we use the extended version of Sketchy with $60,502$ additional photos that Liu \etal \cite{liu2017deep} later introduced for category-level SBIR. In particular, Sketchy-extended \cite{liu2017deep} holds 125 categories with $75,471$ sketches and $73,002$ images in total. Existing zero-shot SBIR \cite{dey2019doodle, dutta2019semantically} works split the dataset into $104/21$ disjoint classes for training/testing(unseen). We keep the same $21$ classes for testing (novel classes), while for hyperparameter tuning, out of 104 classes, we consider $64$ for training and the rest $40$ classes for validation. In summary, we call them $T_{train}$ (64 classes), $T_{val}$ (40 classes), and $T_{test}$ (21 classes) respectively. The train set ($T_{train}$) is often referred to as \emph{base dataset} and is further split into three subsets $(T_{train}^{train}:T_{train}^{val}:T_{train}^{test})=(60\%:20\%:20\%)$. The subset $T_{train}^{test}$ is used to evaluate the overall performance on the base classes during incremental setup. The steps outlined above are followed for both sketches and photos. For every model evaluation, we follow the same settings, including the categories' division and incremental training samples.

\subsection{Implementation Details} 
We have implemented the DIY-FSCIL framework using PyTorch \cite{paszke2017automatic} and conducted the experiments using one {11-GB NVIDIA RTX 2080-Ti GPU}. We employ the standard ResNet18 model  as the backbone feature extractor ($\mathcal{F}_{\theta}$).  The features of the input image are derived from the final pooling layer of the $\mathcal{F}_{\theta}$ with a dimension of $d=512$. We use a one-layer GAT to design our weight generator $G_{\psi}$. In the initial stage, the feature extractor ($\mathcal{F}_{\theta}$) is trained on the training set $T_{train}^{train}$ for 100 epochs.
In the second stage $\mathcal{F}_{\theta}$ is kept frozen while training the weight generation module involving GAT for next 60 epochs. We use SGD optimiser with learning rate 0.01 and batch size of 8 for all experiments. In order to reduce the error caused by the random sampling of the incremental classes and its samples, we report the average results obtained by \emph{five} different seeds. 

\subsection{Evaluation Protocol} 
Following the incremental step $\mathcal{R}_{w}^{base} \rightarrow \mathcal{R}_{w}^{new}$, we evaluate the performance of staged operations $\mathcal{F}_{\theta} \circ \mathcal{R}_{w}^{new}$ under three circumstances -- (a) upon only novel classes, (b) upon only base classes, and (c) upon both base and novel classes. While for only novel classes the class label space consists of $y_j \in \mathcal{C}_{novel} = \{C_1^n \cup  C_2^n  \cdots  C_{K_n}^n\}$, the same for only base classes becomes $y_j \in \mathcal{C}_{base} = \{C_1^b \cup  C_2^b \cdots C_{K_b}^b\}$. Furthermore, for evaluation under base and novel classes, the label space spans across $y_j \in \mathcal{C}_{both} = \{\mathcal{C}_{base} \cup \mathcal{C}_{novel}\}$. These three evaluating situations answer -- (a) how well the model adapts to novel classes from few (1 or 5) sketch examples, (b) how well the model is able to preserve the accuracy (mitigating catastrophic forgetting) of the base classes for which the training data is inaccessible during the incremental step, (c) how well the model performs overall for both base and novel classes. Following the two-stage training using $T_{train}^{train}$, i.e., pre-training on base-classes followed by learning few-shot weight generator, we obtain $\mathcal{F}_{\theta}$ and $G_{\psi}$, which are used for inference under incremental setup.

\keypoint{Evaluation on novel classes (Acc@{novel}):} \cut{For this setting without base classes, we use $\mathcal{C}_{novel}$.} Test set ($T_{test}$) is used to create few shot tasks similar to episodic training. These few shot tasks are formed by sampling $K_{novel}=5$ categories. Then, we sample one (1-shot) or five (5-shot) exemplars per category (sketches) and $15$ query samples per category (photos). Here, the query samples will be from the same novel categories but we make sure that they do not overlap with the exemplars under a particular episode. {$G_{\psi}$ uses exemplar embeddings obtained via $\mathcal{F}_{\theta}$, along with base weights, to generate incremented classifier's weight $W_{new}$, which is then evaluated on the query set.} Apart from helping to understand the model's capability to learn novel classes in a few-shot setting excluding the base classes, this metric also helps assessing the \emph{model's generalisation capability} on \emph{cross-domain} data. Following the existing FSCIL literature \cite{gidaris2018dynamic}, we create $600$ few-shot tasks  and report the average results from them.

\keypoint{Evaluation on base classes (Acc@{base}):} To verify the potential of mitigating the catastrophic forgetting issue, we evaluate the recognition performance on base categories using the $T_{train}^{test}$ subset on the incremented classifier $\mathcal{R}^{new}_w$. Here, we create few shot tasks by randomly sampling $K_{b}=5$ categories from the base classes without replacement, followed by evaluating with $15$ query photos for each category.

\keypoint{Evaluation on all classes (Acc@{both}):} Here the label space spans across all the classes ($y_j \in \mathcal{C}_{both}$). In each episode, we sample from all $K_{b}$ base and $K_{n}$ novel classes. Then, we sample one (1-shot) or five (5-shot) exemplars per novel category (sketches), and 15 query samples for each of the base and novel categories (photos) to evaluate the performance. This metric helps us determine how the base classes' knowledge affects novel classes and vice-versa.

\subsection{Competitor} 

As there exists no prior work dealing with sketch-based FSCIL, we implement the following set of baselines and their adaptions in order to assess the contribution of our proposed framework. $\bullet$ \textbf{B1:} We use a combination of old-base and new-novel classes to \emph{retrain} the complete model. Besides requiring a lot of computational power, this suffers from a severe class imbalance problem between sufficiently available base classes and few exemplars from novel classes.  Nevertheless, this can not be realised in a real scenario. $\bullet$ \textbf{B2:} We only fine-tune the model using the novel classes. It acts as a naive baseline, and is limited due to the issue of catastrophic forgetting.  $\bullet$ \textbf{B3:}  We freeze backbone feature extractor $F_\theta$, and use the class-wise average feature of sketch exemplars as the representative weight-vectors of novel classes along with the pre-trained base-classifier. In other words, we remove the GAT module from our proposed framework. $\bullet$ \textbf{B4:} We further examine the performance of our framework by training both the $\mathcal{F}_{\theta}$ along with the $\mathcal{G}_\psi$. This is used to analyse the importance of freezing the feature extractor $\mathcal{F}_{\theta}$. $\bullet$ \textbf{B5:} During testing, we utilise real images as the support set. As images are more detailed than sketches, this model serves as our upper boundary. However, it fails to address our main concern of violating the data privacy norm. For a fair comparison, we utilise the same settings for all the models as our framework. Although existing FSCIL methods are not specifically designed \cite{gidaris2018dynamic, tao2020few, snell2017prototypical} to deal with cross-modal sketch exemplars, we naively adopt those under our sketch-based FSCIL setup.

\setlength{\tabcolsep}{3pt}
\begin{table}[!ht]
    \centering
\caption{{Average classification accuracy of DIY-FSCIL framework using our self-designed baselines and adopted SOTA FSCIL \cite{gidaris2018dynamic} (not specifically designed for cross-modal). For every experiment, we create $600$ episodes each with $5$ random classes from both novel and base categories separately.  Each episode contains a total $15\times5$ (15 samples from each of the 5 classes) and $15\times5$ query photos from from both novel and base categories respectively. $B5^*$ is an upper bound.}}    
    \vspace{0.05cm}
    \footnotesize
    
    \begin{tabular}{cccccccc}
        \hline
        \multicolumn{3}{c}{\multirow{2}{*}{Methods}}
        &   \multicolumn{3}{c}{\textbf{5-Shot Learning}}\\
        \cline{4-6} 
         & & & Acc@both & Acc@base &Acc@novel\\
        \hline
        \multirow{5}{*}{Baselines} 
         & \multicolumn{2}{c}{$B1$} & 36.29 \% & 73.94\% & 38.92\%\\
         & \multicolumn{2}{c}{$B2$} & 25.86\% & 32.85\% & 70.58\%\\
        
        & \multicolumn{2}{c}{$B3$} & 58.92\% & 73.81\%  & 72.34\%\\
        
        & \multicolumn{2}{c}{$B4$} &  54.50\% & 71.68\% & 71.81\%\\
        
        & \multicolumn{2}{c}{$B5^*$} & 71.52\% & 75.72\% & 85.46\%\\
        \hdashline
         \multirow{3}{*}{SOTA FSCIL} 
         & \multicolumn{2}{c}{\cite{gidaris2018dynamic}} &  50.45\% & 74.35\% & 65.81\%\\
        & \multicolumn{2}{c}{\cite{snell2017prototypical}} & 45.25\% & 74.10\% & 63.46\%\\
        & \multicolumn{2}{c}{\cite{tao2020few}} & 51.54\% & 73.21\% & 66.82\%\\
  \hdashline
         Ours 
        & \multicolumn{2}{c}{\textbf{DIY-FSCIL}} & 60.54\% & 74.38\% & 75.84\% \\
       \hline
    \end{tabular}
 \vspace{2cm}
    \begin{tabular}{cccccccc}
        \hline
        \multicolumn{3}{c}{\multirow{2}{*}{Methods}}
        &   \multicolumn{3}{c}{\textbf{1-Shot Learning}}\\
        \cline{4-6} 
         & & & Acc@both & Acc@base &Acc@novel\\
        \hline
        \multirow{5}{*}{Baselines} 
         & \multicolumn{2}{c}{$B1$} & 31.52\% & 73.98\% & 34.68\%\\
         & \multicolumn{2}{c}{$B2$}  &  28.81\% & 40.91\% & 50.24\%\\
        
        & \multicolumn{2}{c}{$B3$}  & 53.35\% & 73.75\%  & 59.93\%\\
        
        & \multicolumn{2}{c}{$B4$}  & 51.41\%  & 71.68\% & 51.44\%\\
        
        & \multicolumn{2}{c}{$B5^*$}  &  63.47\% & 75.83\% & 73.90\%\\
        \hdashline
         \multirow{3}{*}{SOTA FSCIL} 
         & \multicolumn{2}{c}{\cite{gidaris2018dynamic}}  &  44.71\% & 73.98\% & 64.21\% \\
        & \multicolumn{2}{c}{\cite{snell2017prototypical}}  & 41.97\% & 74.60\% & 61.85\%\\
        & \multicolumn{2}{c}{\cite{tao2020few}}  & 45.81\% & 73.58\% & 63.95\%\\
  \hdashline
         Ours 
        & \multicolumn{2}{c}{\textbf{DIY-FSCIL}}  & 54.97 \% & 74.06\% & 64.10\%\\
       \hline
    \end{tabular}
    \label{tab:maintable2}
 \vspace{-1.5cm}
\end{table}

\setlength{\tabcolsep}{4.5pt}
\begin{table}[t]
    \centering
    \caption{Ablative study: GAT (Graph Attention Network), GC (Gradient Consensus), KD (Knowledge Distillation Loss), CMT (Cross-Modal Training)}
     \vspace{-0.2cm}
    \scriptsize
    \begin{tabular}{cccccccc}
        \hline
        \multirow{2}{*}{GAT} & \multirow{2}{*}{GC} & \multirow{2}{*}{KD} & \multirow{2}{*}{CMT} & & \multicolumn{3}{c}{Metrics} \\
        \cline{6-8} 
         & & & & & Acc@both & Acc@base & Acc@novel  \\
        \hline
        \multirow{2}{*}{\bluecheck} & \multirow{2}{*}{\bluecheck}  & \multirow{2}{*}{\bluecheck}  & \multirow{2}{*}{\bluecheck}  &$5-$shot & 60.54\% & 74.38\%  & 75.84\% \\
        & & & & $1-$shot & 54.97\% & 74.06\%  & 64.10\%\\
        \hline
        
        \multirow{2}{*}{\textcolor{red}{\xmark}} & \multirow{2}{*}{\bluecheck}  & \multirow{2}{*}{\bluecheck}  & \multirow{2}{*}{\bluecheck}  &$5-$shot &  58.92\% & 73.81\%  & 72.34\%\\
        & & & & $1-$shot & 53.35\% & 73.75\%  & 59.93\%\\
        \hdashline
        
        \multirow{2}{*}{\textcolor{red}{\xmark}} & \multirow{2}{*}{\textcolor{red}{\xmark}}  & \multirow{2}{*}{\bluecheck}  & \multirow{2}{*}{\bluecheck} &$5-$shot &  58.47\% & 73.96\%  & 71.67\%\\
        & & & & $1-$shot & 53.22\% & 73.67\%  & 59.46\%\\
        \hdashline
        
        \multirow{2}{*}{\textcolor{red}{\xmark}} & \multirow{2}{*}{\textcolor{red}{\xmark}}  & \multirow{2}{*}{\textcolor{red}{\xmark}}  & \multirow{2}{*}{\bluecheck}  &$5-$shot &   57.47\% & 70.96\%  & 69.67\%\\
        & & & & $1-$shot & 51.22\% & 71.67\%  & 57.46\%\\
        \hdashline
        
        \multirow{2}{*}{\textcolor{red}{\xmark}} & \multirow{2}{*}{\textcolor{red}{\xmark}}  & \multirow{2}{*}{\textcolor{red}{\xmark}}  & \multirow{2}{*}{\textcolor{red}{\xmark}}  &$5-$shot &  35.19\% & 62.98\%  & 40.52\% \\
        & & & & $1-$shot & 27.67\% & 61.72\%  & 32.83\%\\
        
        \hline
    \end{tabular}
    \label{tab:ablation}
    \vspace{+0.4cm}
\end{table}

\setlength{\tabcolsep}{6pt}
\begin{table}[!htbp]
    \centering
    \caption{Performance with varying n-way/k-shot evaluation}
    \vspace{-0.3cm}
    \footnotesize
    \begin{tabular}{cccccc}
        \hline
         &   &   & \multicolumn{3}{c}{Metrics}\\
        \cline{4-6}
         & & & Acc@both & Acc@base & Acc@novel \\
        \hline
        \multirow{5}{*}{$5-$ way} 
         & \multicolumn{2}{c}{$1-$shot} & 54.97\% & 74.06\% & 64.10\% \\
        & \multicolumn{2}{c}{$5-$shot} & 60.54\% & 74.38\% & 75.84\%  \\
        & \multicolumn{2}{c}{$10-$shot} & 61.61\% & 74.14\%  & 76.95\%\\
         & \multicolumn{2}{c}{$15-$shot} & 62.08\% & 73.95\%  & 77.48\%\\
          & \multicolumn{2}{c}{$20-$shot} & 62.35\% & 74.83\%  & 78.35\%\\
        \hdashline
        \multirow{5}{*}{$10-$ way} 
         & \multicolumn{2}{c}{$1-$shot} & 43.62\% & 73.24\% & 47.31\% \\
        & \multicolumn{2}{c}{$5-$shot} & 51.82\% & 73.37\% & 59.97\%  \\
        & \multicolumn{2}{c}{$10-$shot} & 53.75\% & 73.54\%  & 61.21\%\\
         & \multicolumn{2}{c}{$15-$shot} & 55.46\% & 73.38\% & 62.74\% \\
          & \multicolumn{2}{c}{$20-$shot} & 57.58\% & 73.23\% & 64.37\%  \\
        
      \hline
    \end{tabular}
    \label{tab:exemplar_results}
    \vspace{0.5cm}
\end{table}

\subsection{Performance Analysis} 
In Table \ref{tab:maintable2}, we report the comparative results using the standard \emph{one}-shot and \emph{five}-shot sketch-based FSCIL setting on Sketchy dataset. We make the following observations: (i) Despite using abundant memory and computational resources \textbf{B1} performs poorly on the novel classes, due to the absence of any mechanism to handle few shot classes (i.e., severe class imbalance). This suggests that few shot paradigm is essential to perform reasonably well on novel classes. (ii)  \textbf{B2} adapts fine-tuning on the novel classes without heavy computational overhead. However, doing so declines the model's performance on the base classes due to catastrophic interference. (iii) While \textbf{B3} outperforms baselines \textbf{B1} and \textbf{B2}, it fails to model mutual agreement between base and novel classes for learning discriminative decision boundaries under incremental setup, revealing the importance of our weight refining strategy through GAT module. (iv) Low performance of \textbf{B4} signifies the necessity of freezing the weights of $\mathcal{F}_{\theta}$ during the second stage of training in order to reduce the catastrophic forgetting problem, and also to generalise notably better on \emph{unseen} categories. (v) \textbf{B5} (upper bound) achieves the best numbers, as the support set comes directly from photos, unlike ours where we have a critical challenge due to the domain gap between sketch exemplars and query photos. (vi) Moreover, the performance of SOTA FSCIL methods is limited by a margin of $9.09\%$ under DIY-FSCIL setup. 

To summarise, our framework helps in solving the challenging DIY-FSCIL problem by both alleviating the catastrophic forgetting of the old classes and enhancing the learning of the new classes under a cross-modal sketch-based few shot setting. Moreover, the proposed framework effectively enables the users to build their own novel classes with the support of their imaginative drawings.

\subsection{Further Analysis and Insights} 

\noindent \textbf{Ablation Study:} To justify the contribution of individual design components we conduct an ablation study as reported in Table \ref{tab:ablation}. (i) \emph{GAT:} To access the importance of weight refinement, we remove the GAT module and adapt the framework accordingly. Consequently Acc@novel significantly drops to $62.34\%$ with a decrease of $3.5\%$ for $5$-shot case, and is more pronounced for $1$-shot context, where we perceive a larger drop of $4.17\%$. This observation further strengthens our initial assumption that GAT models an effective mutual agreement strategy to learn discriminative decision boundaries across all $K_{b} + K_{n}$ classes. (ii) \emph{Gradient Consensus (GC):} The use of GC improves the model's performance substantially, and this is particularly apparent in the initial stages. During the initial stage training, GC improves the model accuracy by $2.72\%$ via effective handling of the harmful cross-domain gradient interference while updating the model parameters.  (iii) \emph{Knowledge Distillation (KD):} KD-based regularisation seeks to provide stability and enforces learning of weight generation module. Getting rid of it reduces the Acc@both by a significant $3.75\%$($3.07\%)$ for $1$($5$)-shot setting, thus illustrating its necessity. (iv) \emph{Cross Modal Training (CMT):} While we use only sketch exemplars as the support set in real inference, during episodic training we mix both sketch and photos along with gradient consensus strategy to bridge the domain gap in weight generation process. Removing this cross-modal training drops Acc@both by $27.3\%$($25.35\%$) for $1$($5$)-shot setting. In summary, all of the components work in unison to produce the best overall performance.

\setlength{\tabcolsep}{9pt}
\begin{table}[!htbp]
    \centering
    \caption{Comparative study between \emph{sketch} vs \emph{text} for support set}
    \vspace{-0.3cm}
    \footnotesize
    \begin{tabular}{cccc}
        \hline
         &   \multicolumn{3}{c}{One-shot learning}\\
        \cline{2-4}
         & Acc@both & Acc@base & Acc@novel \\
        \hline
         Text (Word2Vec) & 22.85\% & 73.98\%  & 26.15\%\\
         Text (GloVe) & 22.80\% & 74.04\%  & 26.85\%\\
         Sketch (Ours) & 54.97\% & 74.06\%  & 64.10\%\\
       \hline
    \end{tabular}
    \label{tab:text_results}
\vspace{.1cm}
\end{table}

\vspace{.1cm}
\begin{figure}[!hbt]
	\begin{center}
		\includegraphics[width=1\linewidth]{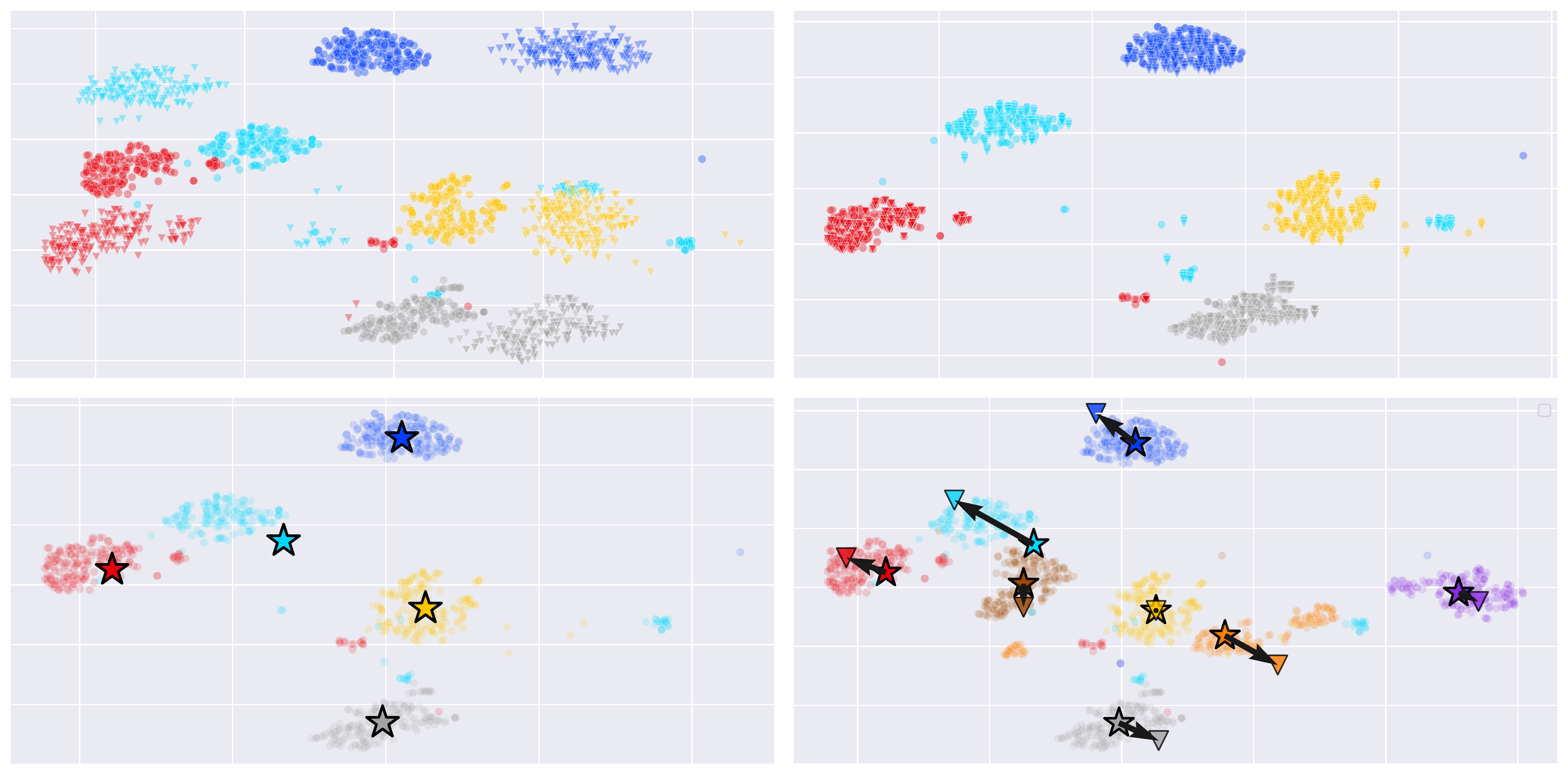} \\
		\vspace{-2.50in}
		\text{\hspace{-2.3in} \small{(a)} \hspace{2.2in} \small{(b)}}\\[1.05in]
		\text{\hspace{-2.3in} \small{(c)} \hspace{2.2in} \small{(d)}}\\
		\vspace{1.05in}
	\end{center}
	\vspace{-0.3in}
	\caption{t-SNE Plots: (a) Photo ($\circ$) and Sketch ($\nabla$) on the common embedding space of $\mathcal{F}_{\theta}$ using naive baseline. (b) Photo ($\circ$) and Sketch ($\nabla$) on the shared embedding space of $\mathcal{F}_{\theta}$ obtained \emph{using} our framework. This can be inferred that our method aligns the both modalities better by minimising the domain gap. 
(c) Base Classes (d) DIY-FSCIL. Here, deep-colour points are class prototypes, light-colour ones show the distribution of real data, star represents class representations during 1st stage, delta (bold) represents the refined vectors during incremental stage, and black arrow indicates the change in weights. Our DIY-FSCIL pushes the classifier weights away from the uncertain areas, resulting in better decision boundaries. Zoom in for better view.}
	\label{fig:tsne_plots}
	\vspace{.2cm}
\end{figure}

\noindent \textbf{Effect of the number of sketch-exemplars:} Next, to investigate how the number of classes and samples affect the overall model performance, we evaluate the framework by varying the number of shots from $\{1, 5, 10, 15, 20\}$ and the number of ways only from $\{5, 10\}$. We depict the corresponding results in Table \ref{tab:exemplar_results}. We infer that larger way hurts the performance because of the ambiguity created by the new classes, while the model's performance increases when training with more number of samples. This portrays the potency of our proposed framework for other CIL variants.
\vspace{0.05cm}

\noindent \textbf{Comparison with text as support-set:} \cut{Here, we inspect the results by using the sketches as class support in comparison to text. We make use of the popular word embedding models, such as Word2Vec \cite{mikolov2013efficient}, and GloVe \cite{pennington2014glove} to generate class label representations.} 
In order to compare our approach with text-based support set, we use the word embeddings from Word2Vec \cite{mikolov2013efficient} and GloVe \cite{pennington2014glove} to generate class representations.
We delineated the results in Table \ref{tab:text_results}. The fine-grained nature of sketches helped surpass text results by a wide margin, showing its efficacy as the class support and a possible substitute to photos. \cut{The sketches' fine-grained nature enabled them outperform text results by a large margin, demonstrating its usefulness as a class support and a viable photo alternative.}

\noindent \textbf{Visualisation of GAT refined features:}
With t-SNE \cite{van2008visualizing}, we visualise class representation weight vectors and classifier weights in a low-dimension space. We exhibit the results for the two configurations -- (i) with GAT, and (ii) without GAT. For this study, five classes are chosen randomly as the base classes, and five additional classes are added as incremental classes. As evident from Fig. \ref{fig:tsne_plots}, during incremental setup, the GAT module refines weights efficiently to push the classifier weights away from the uncertain areas, resulting in better decision boundary.

\section{Conclusion}
\label{chapter:SBIL:Conclusion}
In this chapter, we have introduced a novel framework for few shot class incremental learning without violating the data privacy and ethical norms. This method also empowers the users to construct novel categories just by providing a few imaginative sketches doodled by themselves. The proposed framework unifies Knowledge Distillation, Gradient Consensus, and Graph Attention Networks to handle this newly proposed DIY-FSCIL paradigm. The effectiveness of the framework is validated by various experiments on the Sketchy dataset \cite{sangkloy2016sketchy}. Our framework is also extendable to other IL methods beyond the CIL used in this study. 

\chapter{Conclusions and Future Work}
    \label{chapter:conclusionAndFutureWork}
    \minitoc
    \newpage
    
\section{Conclusion}
\label{section:ConclusionsConslusion}
This thesis focuses on the practical applicability of sketch for visual understanding tasks. In this spirit, we explore the problem from not only the niche aspect of Fine-grained retrieval alone, but \aneeshan{also from a second theme of generalising widespread use and applicability of sketches for a variety of computer vision tasks. It includes} a broader perspective of self-supervised learning for sketches and sketch-based incremental learning. For the FG-SBIR task alone, we have extended its literature from three angles.  Firstly, we have introduced an \textit{on-the-fly} paradigm for retrieval, where users can retrieve as soon as they start drawing, thus reducing the time through early retrieval. Secondly, semi-supervised learning has been employed for cross-modal instance-level retrieval problems like FG-SBIR to perform better under the \textit{low-data regimes} besides improving the overall performance. Thirdly, we dig deeper by analysing the bottleneck caused by noisy/irrelevant strokes in sketches for FG-SBIR, and faced by many users who are not confident with sketching. Accordingly, we have proposed a noise-tolerant FG-SBIR system through a stroke-subset selector that only selects the meaningful strokes, contributing towards faithful retrieval and eliminating the noisy/irrelevant ones.
Fortifying visual understanding from the perspective of self-supervised learning, we design a sketch-specific pre-text task on hand-drawn data that aim to benefit any sketch-based visual understanding problem via better representation learning. This would essentially reduce the bottleneck of limited training dataset size for various sketch-based visual understanding tasks. Finally, showcasing the practical importance of using sketch as a support modality, we focus on incremental learning. Here, we introduce an entirely novel sketch-based visual understanding task where we show how sketch could be a helpful support modality for few-shot class-incremental learning, thus opening the door for further research. Next, we briefly summarise the contributions made through every chapter of this thesis, from high-level research motivation to deep-learning architecture-specific design choices. 

\subsection{On-the-fly FG-SBIR} 
We for the first time have introduced an on-the-fly retrieval framework for sketch-based image retrieval literature. We design on-the-fly FG-SBIR architecture with reinforcement learning based training.  A conventional solution using triplet loss \cite{yu2016sketch, song2017fine} treats each instant of the same sketch as an independent example. In other words, its feature learning does not account for the temporal sequence nature of the input. Our RL-based framework optimises the reward non-myopically over the whole sequence, such that after backpropagation, representation learning is performed with knowledge of the whole sequence. To ensure that RL works well, we carefully design both global and local rewards which further improve the performance. 
    
\subsection{Semi-supervised FG-SBIR}    
We here aim to remove the bottleneck that dictates all prior FG-SBIR research, by introducing a semi-supervised framework that enables the training of FG-SBIR models using large scale unlabelled data. The key idea of our work is to synergise photo-to-sketch generation and FG-SBIR, i.e., to explore the positive information exchange between generation and retrieval. In particular, we introduce a novel discriminator guided instance weighting scheme, along with consistency loss to train the retrieval model with synthetic photo-sketch pairs. In other words, our framework can be used with any retrieval model in a plug-and-play manner.

\subsection{Noise-Tolerant FG-SBIR}  

Here we identify the bottleneck due to noisy strokes in the problem of fine-grained SBIR and propose a stroke subset selector to assign importance to each stroke. The selector is trained using reinforcement learning with pre-trained FG-SBIR models and can serve as a pre-processing module, bringing significant performance gain. Furthermore, besides being a pre-processing selector, our stroke-subset selector also has multiple downstream applications, such as quantifying the ability to retrieve for a  partial sketch and data augmentation by using rasterized versions of different subsets of strokes as extra training data to the actual FG-SBIR models. 

\subsection{Self-Supervised Learning on Sketches}

Ours is the first self-supervised pre-text task for sketch and handwriting, that is applicable to both raster and vector format data in downstream tasks. Furthermore, we are the first to demonstrate the utility of having dual image/vector space representation. The significant contribution lies in advocating for a self-supervised task that uses the dual raster/vector representation nature unique to sketch and handwriting data for representation learning. Developing novel pre-text tasks for different data modalities is the mainstay of self-supervised learning research, and our work aims to advance that goal. 

\subsection{Sketch-based Incremental Learning}   
{We first address the problem of few-shot class incremental learning (FSCIL) within the broader paradigm of few-shot image classification. Our work primarily addresses two issues in this area: (1) if a model can successfully learn from a modality different from usual photos (i.e. sketches), and (2) the unavailability of photos for training due to ethical/privacy issues. To tackle these challenges, we present a framework that infuses (i) gradient consensus for domain invariant learning, (ii) knowledge distillation for preserving old class information, and (iii) graph attention networks for message passing between old and novel classes. Our work seeks to study the role of human-generated sketches as a support modality in a novel FSCIL pipeline. In the domain of few-shot learning, we aim to address several broader theoretical aspects in regard to the use of sketches for FSCIL (i.e. ensuring high cross-modal learning performance/ how to leverage old class information to learn newer ones whilst preserving the former and preventing model “forgetting”).}

\section{Future research directions}
\label{section:future-research}
Apart from the directions explored in this thesis, there could be various other potential future research pathways worth exploring. Furthermore, we have categorised them into two sub-sections. First, we describe them for specific sketch-community, and later we address them from a broader computer vision perspective.

\subsection{Specific Sketch Community}  

\hspace{0.4cm} (i) As sketching is a dynamic process, not only fine-grained retrieval, but our on-the-fly design is also relevant for other sketch-based visual understanding tasks as well. For instance, following the first introduction of on-the-fly fine-grained SBIR, there have been a few follow-up works \cite{liu2022bi, dai2022multi, wang2021deep} that try to improve FG-SBIR performance under on-the-fly setup further. Furthermore, on-the-fly design could also be explored specifically for the sketch-to-image generation/editing, which is yet another fore-front of sketch-based visual understanding tasks. 
In other words, using on-the-fly sketch-to-image generation \cite{chen2018sketchygan} or editing \cite{yang2020deep}, the user can interactively generate photo-realistic images, while successively adding fine-grained details on the instantly generated images from partial sketches.  
Just as we use a reinforcement learning-based design to deal with partial sketches for retrieval, a similar sketch-specific design in context of generative modelling is needed for this.
\cut{to connect partial sketches with the larger output space of photo domain for photo-realistic image generation from such partial sketches.} 
While there is no information passing between successive instances of the sketch drawing in our inference process, some sequential dependency could be critical for on-the-fly sketch-based image generation/editing \cite{nealen2005detailediting}.  

\vspace{0.2cm}
(ii) We have already shown photo-to-sketch generation as a conjugate task for fine-grained SBIR under semi-supervised setup here. As a future endeavour in the same spirit, photo-to-sketch generation \cite{li2019photo} could be connected  with free-hand sketch-to-image generation, to improve image generation quality from abstract sketches \cite{qi2017simpleface}. 

\vspace{0.2cm}
iii) The hierarchical sketch-vector encoder introduced in Noise-Tolerant SBIR work, could be used as a simple yet powerful alternative for sketch-embedding network in vector space, for different sketch-based visual understanding tasks. Additionally, we feel that removing inconsistent strokes drawn by the users to generate plausible RGB images \cite{yang2020deep} from free-hand sketches is quite relevant an idea to explore further. This in turn would reduce the unwanted deformation that usually appears due to the abstract and subjective nature of sketch.
 
\vspace{0.2cm}
iv) Our self-supervised pre-text task in this thesis connects the feature learning process for sketch and handwriting. Therefore, how sketch and handwriting could help each other for unsupervised feature learning on sparse chirographic (e.g. sketch/handwriting) image data is an interesting direction to explore further. Beyond this, vectorization and rasterization  process could also be connected in a common latent space \cite{romback2020n2n}, and some mutual information based modelling \cite{oord2018representation} could further improve the feature learning on sketch. Moreover, vectorization as a pre-text task whose supervision is free-of-cost have been recently adopted for zero-shot SBIR \cite{Sketch3T} under test-time adaptation pipeline. Similarly, the dual representation of sketch is a generic concept and could be employed for different sketch-based visual understanding problems as an auxiliary task for feature learning in future. 

\vspace{0.2cm}

v) While we have shown that sketch could be a potential support modality for few-shot class incremental learning \cite{he2020incrementalonline}, sketch due to its fine-grained capability could be further studied for few-shot dynamic hierarchical classification. For example, in an N-class classification problem, we can further make the N-th class more fine-grained as $N_A$ , $N_B$ and $N_{rest}$ where few-shot sketch exemplars for $N_A$ and $N_B$ could be provided via sketches. This new problem setup could open a new research direction in the intersection of sketch-based few-shot learning \cite{bhunia2022doodle} and fine-grained image classification \cite{chang2021your}.

\subsection{Broader Vision Community}  

\hspace{0.4cm} (i) While we introduce the concept of on-the-fly system with early retrieval objective in context of sketch-based image retrieval, such design holds the potential to influence standard text-based image retrieval \cite{wang2019camp} design as well. For example, a reinforcement learning based design could benefit the text-based retrieval keep human in the loop of the process \cite{ruiyi2019textRL}.

\vspace{0.2cm}
ii) Our semi-supervised fine-grained SBIR is one of the first works that introduce semi-supervised learning for cross-modal instance-level retrieval. Therefore, not only FG-SBIR, such design could be useful for other instance-level matching problems like image captioning \cite{xu2021towards}, text/tag \cite{gattupalli2019weakly} based image retrieval, etc. Moreover, the discriminator guided instance-specific re-weighting to quantify the certainty of pseudo-labelled data \cite{petrovai2022pseudo} is a generic design choice that could be helpful for cross-modal semi-supervised literature in general. 

\vspace{0.2cm}

iii) Our hierarchical stroke-subset selector could be easily extended for vision problems where an efficient selection strategy is the major objective, e.g. video \cite{rochan2019video} or text \cite{liu2019text} summarisation. Furthermore, this paradigm of training a  pre-processing module  for a given pre-trained model, involving any intermediate non-differentiable operation, is a pivotal design choice in computer vision and our attempt at solving it could reinforce a plethora of other vision task in general as well.

\vspace{0.2cm}

iv) Beyond hand-drawn data, our Sketch2Vec self-supervised task could be extended for vector graphics data \cite{reddy2021im2vec}, where images are represented as collections of parameterised shape primitives \cite{stephan2022sketchprimitives} rather than a regular raster of pixel values. Moreover, this is relevant in context of multi-modal \cite{taleb2021multimodal} self-supervised learning where the data could be represented in multiple-modality due to its nature of representation. 

\vspace{0.2cm}

v) Our sketch based few-shot class incremental method in general could be used as an evaluation paradigm to check the generalizability of few-shot frameworks \cite{cheraghian2021semantic, tao2020few,  li2020prototypical}. In recent times, there has been an increased interest in domain-generalisation \cite{li2019episodic} research. Our works connect three dimensions of computer vising literature into a single pipeline -- few-shot learning \cite{snell2017prototypical}, domain generalisation \cite{li2019episodic}, and incremental learning \cite{cheraghian2021semantic, cheraghian2021semantic}, together. Using our findings and the nature of our solution therefore, various downstream tasks can be attempted in either of these directions.


\thispagestyle{empty}

\bibliographystyle{plain}
\bibliography{Ref}
\end{document}